\documentclass[a4paper,fleqn]{cas-sc}
\usepackage[square,numbers,compress]{natbib}
\usepackage{csquotes}
\usepackage{xcolor}
\usepackage{wasysym}
\usepackage[colorinlistoftodos,prependcaption]{todonotes}
\usepackage[inline]{enumitem}
\usepackage{xspace}
\usepackage{hyperref}
\usepackage{acronym}

\usepackage{multicol}

\usepackage{subcaption}
\usepackage{float}
\usepackage{longtable}
\usepackage[usenames,dvipsnames]{pstricks}
\usepackage{pstricks-add}
\usepackage{amssymb}
\usepackage{pifont}

\def\tsc#1{\csdef{#1}{\textsc{\lowercase{#1}}\xspace}}
\tsc{WGM}
\tsc{QE}
\tsc{EP}
\tsc{PMS}
\tsc{BEC}
\tsc{DE}

\usepackage{algorithm}
\usepackage{algpseudocode}

\begin{document}
\let\WriteBookmarks\relax
\def\floatpagepagefraction{1}
\def\textpagefraction{.001}

\shorttitle{RL for Acceleration and Optimization on FPGA/IC }

\shortauthors{Dr. H Kheddar}

\title [mode = title]{Reinforcement Learning {for Hardware Acceleration and System-Level Optimization} on FPGA and Integrated-Circuit Platforms{: A Survey} }                      

\vskip2mm

\author[1]{Hamza Kheddar}
[orcid=0000-0002-9532-2453]
\ead{kheddar.hamza@univ-medea.dz}
\credit{Conceptualization; Methodology; Resources; Investigation; Writing original draft; Writing, review, and editing}

\address[1]{LSEA Laboratory, Department of Electrical Engineering, University of Medea, 26000, Algeria}

\tnotetext[1]{The first is the corresponding author.}

\begin{abstract}
Reinforcement learning (RL) has emerged as a powerful paradigm for enabling adaptive, autonomous, and self-optimizing behavior in modern computing systems, particularly as applications increasingly migrate toward embedded, edge, and heterogeneous hardware platforms. Field-programmable gate arrays (FPGAs) and integrated circuits (ICs) offer compelling advantages for RL deployment, including low latency, energy efficiency, and fine-grained architectural customization, but they also introduce unique design challenges that differ substantially from conventional CPU- and GPU- based implementations. This review presents a comprehensive survey of RL implementations on FPGA and IC platforms, with a particular emphasis on architectural design, performance measurement, and hardware–algorithm co-design. A broad body of more than one hundred studies is systematically analyzed and organized into major RL families, ranging from model-based Markov decision processes and classical value-based methods to deep RL, Actor-critic (AC), policy-gradient (PG), and multi-agent approaches. The review further categorizes existing works according to abstraction levels, hardware realization strategies, and application domains, highlighting how different RL algorithms map onto reconfigurable and fixed hardware substrates. In contrast to prior surveys that focus narrowly on specific algorithms or platforms, this work provides a unified perspective spanning both FPGA and IC implementations, encompassing acceleration techniques, reconfigurable hardware approaches, energy-aware control, and system-level optimization. Detailed comparisons are provided in terms of resource utilization, latency, throughput, power consumption, and achieved performance gains. Additionally, this review identifies recurring architectural patterns, discusses practical limitations, and outlines open research challenges and future directions for RL-driven intelligent hardware systems. By consolidating design insights, evaluation metrics, and application trends, this review aims to serve as a comprehensive reference for researchers and practitioners working at the intersection of RL and hardware platforms.
\end{abstract}

\begin{keywords}
Reinforcement learning  \sep
FPGA Acceleration \sep
AI Hardware \sep
Information processing \sep
Neural Network Acceleration \sep
Hardware-Aware Learning 
\end{keywords}

\maketitle

\begin{table*}[]
\centering
{\footnotesize \section*{Acronyms and Abbreviations}}
\begin{multicols}{3}
\footnotesize
\begin{acronym}[LUTRAM] 
\acro{A3C}{asynchronous advantage actor–critic}
\acro{AC}{actor-critic}
\acro{AI}{artificial intelligence}
\acro{ALU}{arithmetic logic unit}
\acro{ANN}{artificial neural network}
\acro{ARM}{adaptable receptive module}
\acro{ASIC}{application-specific integrated circuit}
\acro{BPN}{binarized p-network}
\acro{BRAM}{block RAM}
\acro{CLB}{configurable logic block}
\acro{CNN}{convolutional neural network}
\acro{CPU}{central processing unit}
\acro{CUT}{circuit under test}
\acro{DDPG}{deep deterministic policy gradient}
\acro{DDQN}{double deep Q-network}
\acro{DDR}{double data rate}
\acro{DL}{deep learning}
\acro{DNN}{deep neural network}
\acro{DP}{dynamic programming}
\acro{DPR}{dynamic partial reconfiguration}
\acro{DQN}{deep Q-network}
\acro{DRL}{deep reinforcement learning}
\acro{DSA}{dynamic spectrum access}
\acro{DSC}{domain-specific
compilation}
\acro{DSP}{digital signal processing}
\acro{DVFS}{dynamic voltage and frequency scaling}
\acro{EDA}{electronic design automation}
\acro{EV}{electric vehicle}
\acro{FF}{flip-flop}
\acro{FIFO}{first-in first-out}
\acro{FIPS}{full iterations processed per second}
\acro{FPGA}{field-programmable gate array}
\acro{FSM}{finite state machine}
\acro{GNN}{graph neural network}
\acro{GPU}{graphical processing unit}
\acro{HDC}{hyperdimensional computing}
\acro{HLS}{high-level synthesis}
\acro{HT}{hardware trojan}
\acro{IC}{integrated circuit}
\acro{IoT}{internet of thing}
\acro{LLM}{large language model}
\acro{LUT}{lookup table}
\acro{LUTRAM}{LUT-based distributed memories}
\acro{MAC}{multiply accumulator}
\acro{MARL}{multi-agent reinforcement learning}
\acro{MC}{Monte--Carlo}
\acro{MDP}{Markov decision process}
\acro{ML}{machine learning}
\acro{MLP}{multilayer perceptron}
\acro{MMCM}{mixed-mode clock manager}
\acro{NoC}{network-on-chip}
\acro{PID}{proportional-integral-derivative}
\acro{PL}{programmable logic}
\acro{PLL}{phase-locked loop}
\acro{PPO}{proximal policy optimization}
\acro{PS}{processing system}
\acro{PG}{policy gradient}
\acro{QL}{Q-learning}
\acro{QoR}{quality of results}
\acro{RAM}{random access memory}
\acro{RL}{reinforcement learning}
\acro{RNN}{recurrent neural network}
\acro{RTL}{register-transfer level}
\acro{SAC}{soft ator–critic}
\acro{SARSA}{state–action–reward–state–action}
\acro{SCA}{side-channel analysis}
\acro{SCOAP}{sandia controllability/observability analysis program}
\acro{SoC}{system-on-chip}
\acro{SRAM}{static random-access memory}
\acro{STDP}{spike-timing–dependent plasticity}
\acro{SLD}{system-level design}
\acro{TD}{temporal-difference}
\acro{TD3}{twin delayed deep deterministic policy gradient}
\acro{UAV}{unmanned aerial vehicle}
\acro{VHDL}{very high-speed integrated circuit hardware description language}
\acro{SNN}{spiking neural network}
\end{acronym}

\end{multicols}
\end{table*}


\section{Introduction}
\label{sec1}

Recent years have witnessed a growing shift toward intelligent, autonomous decision-making embedded directly within computing systems, with increasing emphasis on assessing the real-world applicability and feasibility of algorithms that demonstrate strong performance under simulation-based evaluations. As workloads become increasingly dynamic and heterogeneous—particularly in networking and edge computing, cyber--security, robotics \cite{basaklar2023dtrl}, natural language processing, intelligent sensing, and embedded environments—traditional static control and optimization techniques, such as proportional--integral--derivative  control and fixed-rule heuristic approaches \cite{pastrana2025integrating} or meta-heuristic-based optimization methods \cite{sharma2022systematic}, often struggle to cope with variability in data rates, power budgets, timing constraints, and environmental conditions. These approaches are typically designed around fixed operating assumptions and lack the adaptability required to respond effectively to non-stationary workloads or unforeseen runtime disturbances.

\Ac{RL}, as a core paradigm within \ac{ML}, has gained prominence in this context because it provides a principled framework for sequential decision-making under uncertainty. By continuously interacting with an environment and updating its policy based on observed rewards, an \ac{RL} agent can learn to adapt its behavior online, even when system dynamics are partially unknown or change over time. This capability makes \ac{RL} particularly attractive for modern computing platforms that must balance competing objectives such as performance, energy efficiency, latency, and reliability under tight operational constraints. However, translating \ac{RL} from software-centric implementations to practical hardware platforms introduces significant challenges. Many \ac{RL} algorithms rely on iterative updates, stochastic exploration, and memory-intensive computations that are difficult to reconcile with the strict latency, power, and determinism requirements of embedded systems. Implementations based purely on general-purpose \acp{CPU} or \acp{GPU} often suffer from high energy consumption, unpredictable execution timing, and limited scalability, which restrict their suitability for real-time or resource-constrained applications.

\Ac{FPGA}s and \acp{IC} provide an alternative hardware substrate that fundamentally reshapes how \ac{ML}, particularly \ac{RL}, can be realized in practice. \acp{FPGA} provide fine-grained parallelism, customizable datapaths, and reconfigurability, allowing \ac{RL} algorithms to be mapped onto hardware architectures tailored to specific workloads and constraints. \ac{IC}-based realizations, including \acp{ASIC} and \acp{SoC} platforms, further extend these advantages by enabling deterministic execution, compact form factors, and ultra-low power operation. At the same time, these platforms impose stricter design constraints, limited flexibility after deployment, and increased development complexity compared with software-based solutions. As a result, the intersection of \ac{RL} with hardware design has evolved into a multidisciplinary research area encompassing algorithm--hardware co-design \cite{jiang2020hardware}, hardware acceleration \cite{liu2025automated,gerlinghoff2021e3ne,inayat2024fpga}, architectural specialization \cite{chowdhury2024retrieval}, numerical precision optimization \cite{azzouzi2026efficient}, and system-level control, including digital twin--based frameworks \cite{oulefki2025innovative}.  In this emerging landscape, \ac{RL} plays a dual role: it can be treated as a workload to be accelerated on hardware, and it can also serve as a control mechanism for the hardware itself, governing tasks such as dynamic voltage and frequency scaling, resource allocation, scheduling, partial reconfiguration, and energy management. This duality blurs the traditional boundary between learning algorithms and hardware platforms, creating new opportunities as well as new design challenges. Despite rapid progress, the existing literature on RL implementations for \ac{FPGA} and \ac{IC} platforms remains fragmented.

Motivated by the increasingly practical shift of \ac{RL} from a purely algorithmic concept to a hardware-constrained workload, this review consolidates and structures the fast-growing body of \ac{RL} implementations on \acp{FPGA} and \acp{IC} platforms. 
In particular, this survey emphasizes real-time pattern recognition and perception-driven decision-making systems, where hardware constraints critically affect learning efficiency, latency, energy consumption, and deployment feasibility. 
Instead of revisiting algorithmic fundamentals, it adopts a hardware-centric lens to clarify how design and implementation choices shape real-world realizations of \ac{RL}. 
The review organizes the literature into coherent categories and distills recurring design patterns, trade-offs, and limitations, thereby clarifying the current state-of-the-art and highlighting open challenges that guide future research at the intersection of \ac{RL}, pattern recognition, and intelligent hardware.

\subsection{Related work and our contribution}

Several surveys have examined the relationship between \ac{RL} and reconfigurable or heterogeneous hardware, yet significant gaps remain. The survey in \cite{rothmann2022survey} provides an extensive overview of domain-specific \ac{RL} architectures, with a primary focus on \ac{FPGA}-based accelerators. However, it does not integrate \ac{IC} implementations, unified abstraction layers, nor does it cover the complete spectrum of \ac{RL} algorithmic families such as \ac{SARSA}, \ac{MC}, \ac{AC}, and \ac{PPO}. Similarly, the work in \cite{goswami2023application} concentrates on \ac{ML} applications within \ac{FPGA} \ac{EDA} tools. While valuable, its treatment of \ac{RL} is brief, lacks architectural depth, and does not relate \ac{RL} methods to accelerator design or runtime optimization. The review in \cite{zhang2023machine} centers on \ac{ML} techniques for control and monitoring but does not address \ac{RL}-driven \ac{FPGA} or \ac{IC} architectures, nor broader comparisons across \ac{RL} categories. Additionally, the heterogeneous-chip scheduling survey in \cite{miao2025review} identifies \ac{RL} primarily as a future research direction without analyzing existing \ac{RL} implementations, hardware mapping strategies, or design-flow implications. Consequently, none of these studies provide a unified \ac{RL}-centric perspective that spans both \ac{FPGA} and \ac{IC} platforms while simultaneously comparing all major \ac{RL} types, synthesizing research areas and applications, and identifying cross-cutting architectural and methodological limitations. A detailed comparison of our coverage with prior surveys is provided in Table~\ref{tab:01}. 

\begin{table*}[ht!]
\centering
\scriptsize
\caption{Comparison of the proposed review with existing  FPGA-related surveys and reviews. A  mark (\CIRCLE{}) signifies that a specific area has been covered, whereas a  mark (\Circle{}) denotes that it has not been addressed. The symbol (\LEFTcircle{}) indicates that key aspects of the field remain largely unaddressed.}
\label{tab:01}
\begin{tabular}[ht!]{m{0.4cm}m{0.7cm}m{0.4cm}m{0.7cm}m{0.2cm}m{0.5cm}m{0.5cm}m{0.3cm}m{0.2cm}m{0.5cm}m{0.3cm}m{0.5cm}m{0.3cm}m{0.3cm}cccccc}
\hline
Ref. & Paper & NoPs & Related &\multirow{2}{*}{IC}  & \multirow{2}{*}{FPGA} &  \multirow{2}{*}{BGD} &
\multicolumn{8}{c}{RL types} & Research  & RL     & Case & Limitations &  \\ \cline{8-15}
& type & RL & work& &  & & MDP& QL & SARSA &  DQN  & DDPG & AC  &  MC & PPO &  areas    & Apps. & study & 
 Perspectives  \\ \hline

\cite{rothmann2022survey} & Survey& 14 & \Circle{} &  \Circle{} & \CIRCLE{} & \LEFTcircle{} & \Circle{} &  \CIRCLE{} & \Circle{} &  \CIRCLE{} & \CIRCLE{} & \Circle{} & \Circle{} & \CIRCLE{} & \Circle{} & \Circle{} & \Circle{} & \LEFTcircle{}\\ \hline
  
\cite{goswami2023application} & Survey & 9 & \CIRCLE{} & \Circle{} & \CIRCLE{} & \CIRCLE{} & \Circle{} &  \Circle{} & \Circle{} &  \Circle{} & \Circle{} & \Circle{} & \Circle{} & \Circle{} & \Circle{} & \LEFTcircle{} & \Circle{} & \LEFTcircle{} \\ \hline

\cite{zhang2023machine} & Review & 1 & \Circle{} & \LEFTcircle{} & \CIRCLE{} & \LEFTcircle{} &  \Circle{} & \Circle{} &  \Circle{} & \Circle{} & \Circle{} & \Circle{} & \Circle{} & \Circle{} & \Circle{} & \LEFTcircle{} & \Circle{} & \LEFTcircle{} \\ \hline

\cite{miao2025review} & Review & 1 & \Circle{} & \CIRCLE{}  & \CIRCLE{}  & \Circle{} &  \Circle{} & \Circle{} &  \Circle{} & \Circle{} & \Circle{} & \Circle{} & \Circle{} & \Circle{} & \Circle{} & \Circle{} & \Circle{} & \LEFTcircle{}\\ \hline

\textbf{Our}& Review & 120 & \CIRCLE{} & \CIRCLE{} & \CIRCLE{} & \CIRCLE{} &  \CIRCLE{} & \CIRCLE{} &  \CIRCLE{} & \CIRCLE{} & \CIRCLE{} & \CIRCLE{} & \CIRCLE{} & \CIRCLE{} & \CIRCLE{} & \CIRCLE{} & \CIRCLE{} & \CIRCLE{}\\ \hline
\end{tabular}
\begin{flushleft}
\scriptsize{Abbreviations: Background (BGD);  Number of papers (NoPs)   }     
\end{flushleft}
\end{table*}

In contrast, the present review fills these gaps by systematically analyzing more than 120 \ac{FPGA}/\ac{IC}-related \ac{RL} works and providing a unified and hardware-centric synthesis. The main contributions are summarized as follows:

\begin{itemize}[itemsep=1pt,leftmargin=15pt]
    \item \textbf{Unified \ac{FPGA}/\ac{IC} perspective:} consolidates \ac{RL}-based realizations across reconfigurable accelerators and dedicated \ac{IC} engines within a single \ac{RL}-centric view.
    \item \textbf{Comprehensive \ac{RL} coverage:} covers all major \ac{RL} families (including \ac{SARSA}, \ac{MC}, \ac{AC} methods, and modern deep \ac{RL} such as \ac{PPO}) and relates algorithmic classes to hardware realization patterns.
    \item \textbf{Abstraction levels and toolflows:} structures the literature by hardware abstraction levels and design flows, clarifying how implementation choices affect deployability and reproducibility.
    \item \textbf{Metrics-driven comparison:} synthesizes reported evaluation metrics (latency/throughput, resource utilization, and energy efficiency) and highlights best-reported performance and improvements under application-driven constraints.
    \item \textbf{Research areas and application synthesis:} maps the literature into major research fields and provides an evidence-driven synthesis of the most investigated application domains of \ac{RL}-enabled intelligent hardware.
    \item \textbf{Table-based visual benchmarking:} summarizes most reviewed works in unified tables capturing the target application, best performance/improvement, key limitations, and implementation notes, enabling quick visual comparison.
    \item \textbf{Cross-cutting limitations and open challenges:} distills recurring architectural and methodological bottlenecks and outlines actionable directions for future \ac{RL}-based intelligent hardware systems.
\end{itemize}

Finally, while several \ac{RL}-based \ac{FPGA}/\ac{IC} implementations incorporate \acp{CPU} or \acp{GPU}, these processors typically serve auxiliary functions---such as orchestration, environment simulation, or data management---whereas the dominant learning and inference workload is executed on \acp{FPGA}, \acp{ASIC}, and custom hardware engines, with general-purpose processors contributing minimally to the computational pipeline.

\subsection{Review methodology}

To ensure methodological rigor, transparency, and reproducibility, this review adopts a systematic literature search and selection strategy centered on peer-reviewed studies addressing the implementation of \ac{RL} on \ac{FPGA} and \ac{IC} platforms. The primary search was conducted using Scopus as the main indexing engine, which provides broad coverage of major scientific publishers and digital libraries, including IEEE Xplore, Elsevier ScienceDirect, ACM Digital Library, Web of Science, Wiley, Springer, and IET Digital Library, among others. To consistently retrieve works at the intersection of \ac{RL} algorithms and hardware realization, a structured Boolean query was formulated and applied across databases using title, abstract, and keyword fields. The search logic was designed to capture both classical and deep \ac{RL} methods, while explicitly constraining results to hardware-oriented implementations rather than purely algorithmic or simulation-based studies. The generic query structure can be expressed as:

\texttt{Selected papers =
FROM (“Title” OR “Abstract” OR “Keywords”)
SELECT (References WHERE
(RL OR DRL OR “Reinforcement Learning” OR “Deep Reinforcement Learning”) AND (FPGA OR “Field-Programmable Gate Array” OR ASIC OR IC OR SoC OR “Integrated Circuit”)) FILTER (Language = “English” AND Year BETWEEN 2015 AND 2026 AND DocumentType = (“Journal Article” OR “Conference Paper”))}                                 

The logical operators AND and OR ensure precise targeting of studies that explicitly combine RL with \ac{FPGA} or \ac{IC} hardware platforms. The selected time window reflects the emergence of practical \ac{RL} hardware realizations, from early tabular and \ac{MDP}-based accelerators to modern deep and \ac{AC} architectures.

{To balance completeness with technical focus, the retrieved studies were further screened using relevance and quality criteria. Seminal \ac{RL} hardware-acceleration works were retained when they directly supported the \ac{FPGA}/\ac{IC}-centered taxonomy, architectural discussion, performance comparison, or application-level analysis of this review. However, earlier works already extensively covered in prior surveys were not repeated exhaustively unless they established a necessary technical baseline or motivated a later research direction. This avoids redundant historical coverage while preserving the foundational context required for the proposed survey.}

{Redundant studies, papers with insufficient hardware-implementation details, works lacking meaningful evaluation results, and studies only weakly related to \ac{RL}-based \ac{FPGA}/\ac{IC} platforms were excluded. Priority was given to recent studies that provide clear implementation details and contribute to emerging directions such as energy-efficient \ac{RL} acceleration, quantized \ac{RL} hardware, \ac{FPGA}/\ac{IC}-based runtime optimization, \ac{RL}-assisted synthesis, \ac{MARL} hardware, and application-specific deployment.}

\subsection{Paper structure}

\begin{figure}[!htbp]
    \centering
    \includegraphics[width=0.8\linewidth]{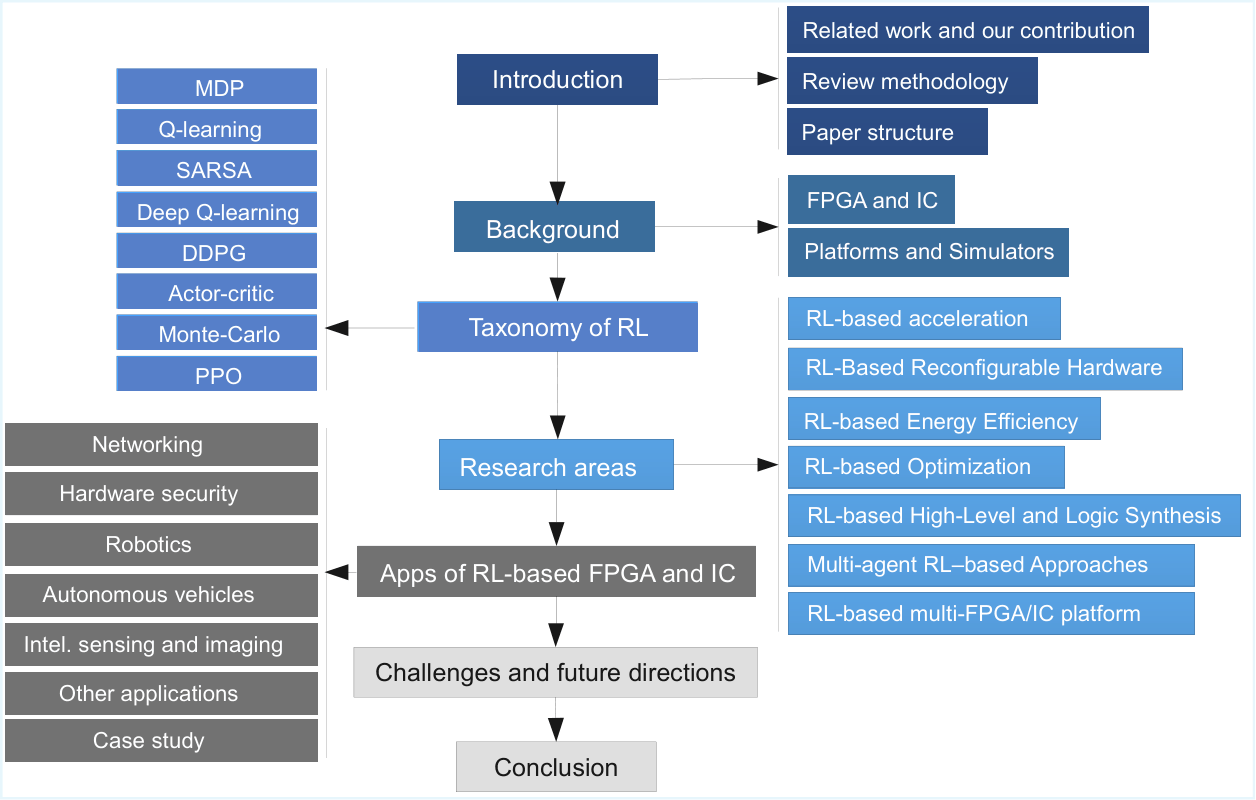}
    \caption{Survey roadmap outlining the organization of \ac{RL}-based \ac{FPGA} and \ac{IC} platforms.}
    \label{fig:rm}
\end{figure}

The remainder of this review is organized as follows. Section \ref{sec2} provides essential background on \acp{FPGA} and \acp{IC} platforms, clarifying implementation abstraction levels and typical toolflows/simulators used in practice. Section \ref{sec3} presents a taxonomy of \ac{RL}-based \ac{FPGA}/\ac{IC} works (considering \ac{RL} in a broad sense), organizing the literature by major \ac{RL} families and highlighting implications for hardware realization. Section \ref{sec4} synthesizes the major research areas that define \ac{RL}-based \ac{FPGA}/\ac{IC} design and summarizes the dominant directions and recurring design patterns reported in the literature. Section \ref{sec5} reviews the most investigated application domains of \ac{RL}-enabled \ac{FPGA}/\ac{IC} systems (e.g., networking, information security, intelligent sensing/imaging, and other domains), emphasizing reported improvements and practical constraints. Section \ref{sec6} consolidates the key challenges and limitations that still hinder scalability and deployment maturity as well as outlines future research directions to advance more robust, scalable, and reusable intelligent hardware systems. Finally, Section \ref{sec7} concludes the review. A roadmap of the overall review is provided in Figure \ref{fig:rm}.

\section{Background}
\label{sec2}

\subsection{FPGA and IC}

Modern \acp{FPGA} are heterogeneous fabrics composed of several tightly coupled resources rather than only a sea of logic cells. The core of the device is formed by a 2D array of \acp{CLB}. Each \ac{CLB} contains one or more \emph{slices}, and a slice integrates \acp{LUT}, \acp{FF}, optional \ac{LUTRAM}, and dedicated carry logic. \acp{LUT} implement arbitrary combinational functions, \acp{FF} provide sequential storage, \ac{LUTRAM} enables small on-chip memories close to the logic, and fast carry chains support efficient arithmetic operations such as adders and counters.

\begin{figure}[!htbp]
    \centering
    \includegraphics[width=0.85\linewidth]{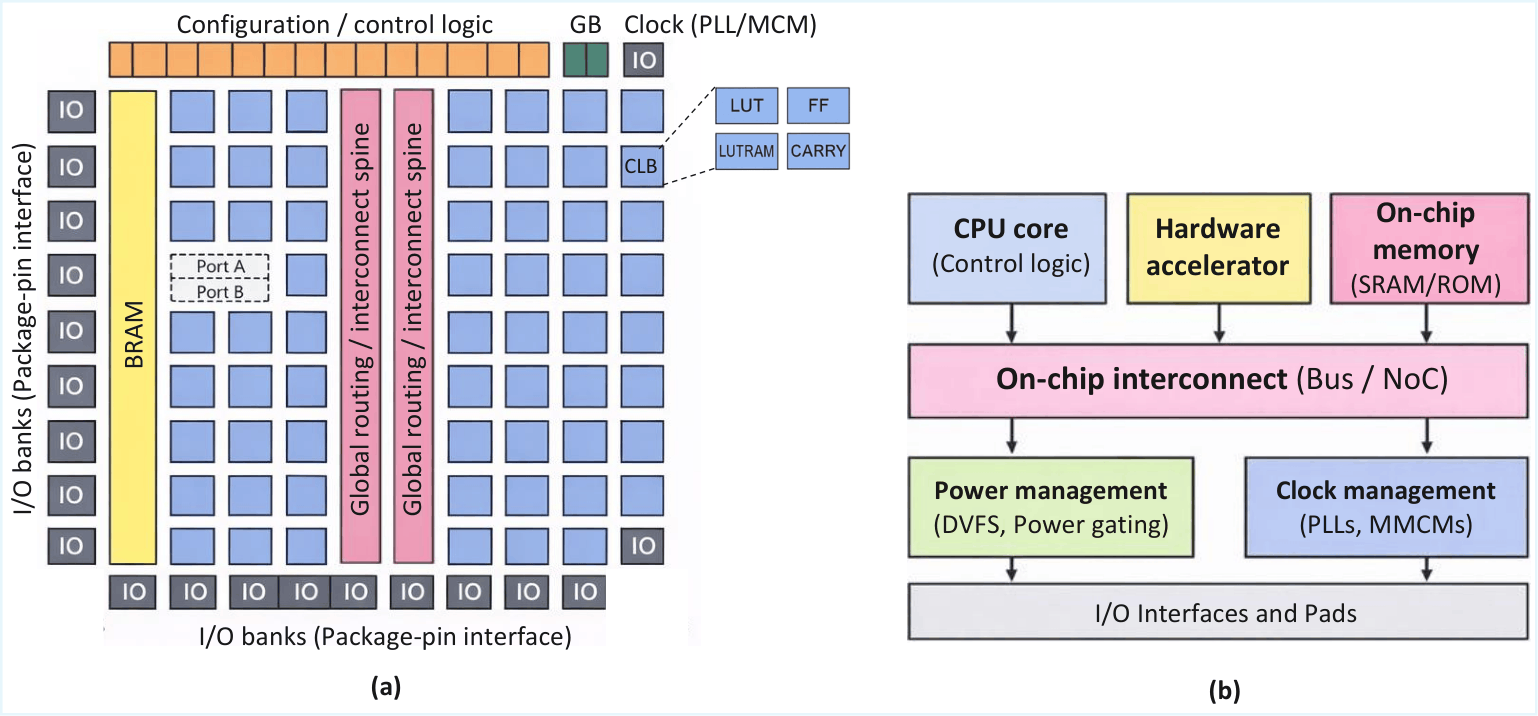}
    \caption{Overview of two hardware architecture representations that could be employed for \ac{RL} workloads. (a) A simplified \ac{FPGA} floorplan showing the distribution of on-chip resources—CLB logic tiles, \ac{BRAM} columns, global routing spines, configuration/control logic, global buffers, clock management blocks, and I/O banks—with a CLB slice highlighting LUT, FF, LUTRAM, and carry logic elements. (b) A conceptual \ac{IC}/\ac{SoC}  architecture illustrating functional subsystems relevant to \ac{RL} execution, including the CPU core, hardware accelerator, on-chip memory, interconnect (Bus/NoC), power management, clock management, and I/O interfaces and pads.}
    \label{fig:FPGAIC}
\end{figure}

Surrounding and interleaved with the \ac{CLB} array are specialized resources. Columns of \acp{BRAM} offer larger, synchronous memories with configurable width and depth, used for data buffering, \acp{FIFO}, and on-chip caches. Dedicated \ac{DSP} slices provide high-speed arithmetic units (typically multipliers, add–accumulate units, and pre/post-adders), which are essential for signal processing, \ac{ML}, and other compute-intensive workloads. At the periphery of the device, I/O banks host programmable input/output blocks that interface the \ac{FPGA} fabric with external signals and support various standards and voltage levels. These computation and storage blocks are connected by a rich hierarchy of routing resources. Local and global interconnect networks, including long lines and dedicated global routing (often driven by global buffers, GBs), allow signals to traverse the chip and distribute clocks and resets. Clock-management tiles, such as \ac{PLL} and \ac{MMCM}, generate and condition internal clocks, enabling frequency synthesis, phase shifting, and jitter reduction. Configuration and control logic, typically including a configuration memory interface and state machines, orchestrate device programming, partial reconfiguration, and global control signals. Together, \acp{CLB} (with \ac{LUT}, \ac{FF}, \ac{LUTRAM}, and carry logic), \ac{BRAM}, \ac{DSP} slices, I/O banks, global routing, clock-management, and configuration/control logic form the heterogeneous substrate on which hardware-aware \ac{RL} and \ac{MDP}-based techniques can operate to dynamically optimize performance, power, and resource utilization. Figure \ref{fig:FPGAIC} (a) illustrate a typical \ac{FPGA} board resources.  

Unlike reconfigurable \acp{FPGA}, an \ac{IC} implements a fixed hardware architecture that is defined at fabrication time. Modern \acp{ASIC} or \ac{SoC} devices integrate multiple subsystems on a single die, including general-purpose \ac{CPU} cores, hardware accelerators, on-chip memories (\ac{SRAM}/read-only memory (ROM)), I/O interfaces, and analog or mixed-signal blocks. The digital logic is typically built from standard cells (gates, flip-flops, etc.) placed and routed once during design, with no post-silicon reconfiguration capability.

These \acp{IC} also embed rich infrastructure for power and clock management, such as clock trees, \acp{PLL}, and \ac{DVFS} controllers, as well as on-chip interconnects like buses or \acp{NoC}. In hardware-aware \ac{RL} and \ac{MDP}-based schemes, the \ac{IC} is therefore treated as a fixed substrate whose tunable parameters include operating points (voltage/frequency domains), power-gating states, and scheduling or resource-allocation policies for the embedded accelerators and memories. This contrasts with \acp{FPGA}, where \ac{RL} may additionally control spatial reconfiguration of the logic fabric itself. Figure \ref{fig:FPGAIC} (b) illustrate a typical \ac{IC} resources.

The unified abstraction hierarchy highlights the structural similarities and key differences between \ac{FPGA} and \ac{IC} design flows. At the  \ac{SLD} level, both platforms rely on high-level algorithm modeling using MATLAB, Simulink, or SystemC to explore functionality and performance. {Beyond this level, \acp{DSC} have increasingly emerged as an additional abstraction layer between high-level \ac{ML}/\ac{RL} models and low-level hardware implementation. Unlike general-purpose \ac{HLS} tools, which mainly translate C/C++/SystemC descriptions into hardware, \acp{DSC} operate on domain-aware representations such as tensor graphs, dataflow graphs, neural-network operators, or accelerator templates. This enables compiler-level optimizations such as operator fusion, memory tiling, scheduling, quantization, and hardware-aware mapping before generating \ac{HLS} code, \ac{RTL}, or intermediate hardware descriptions. In the context of \ac{RL}-based \ac{FPGA}/\ac{IC} systems, \acp{DSC} are particularly relevant for deep \ac{RL} methods such as \ac{DQN}, \ac{PPO}, and \ac{DDPG}, where policy/value networks, replay buffers, and learning-update pipelines introduce repeated computational patterns that can be systematically optimized.}

\Ac{HLS} then bridges software and hardware by converting C/C++/SystemC descriptions into \ac{RTL} for both technologies, although \ac{ASIC} tools emphasize early power and area exploration. \ac{RTL} design remains common to both flows, relying on Verilog or \ac{VHDL} to define datapaths, control logic, and finite-state machines. Divergence appears at the physical level: \ac{FPGA} implementation performs placement, routing, and timing closure over a fixed reconfigurable fabric, while \ac{ASIC} flows involve full custom physical design, including floorplanning, clock-tree synthesis, routing, sign-off timing analysis, and DRC/LVS checks before tape-out. This unified view clarifies how \ac{RL} hardware—such as \ac{MDP} or \ac{QL} accelerators—can be designed coherently for both \ac{FPGA} prototyping and \ac{IC} deployment. Table \ref{tab:fpga_ic_levels} provides a unified view of abstraction layers common to both \ac{FPGA} and \ac{IC} design flows, highlighting their shared stages and key distinctions.

\begin{table*}[htbp]
\centering
\scriptsize
\caption{Unified abstraction levels for \ac{FPGA} and \ac{IC} design flows}
\label{tab:fpga_ic_levels}
\renewcommand{\arraystretch}{1.25}
\begin{tabular}{|m{1.5cm}|m{7cm}|m{7cm}|}
\hline
\textbf{Design Level} & \textbf{\ac{FPGA} Workflow} & \textbf{\ac{IC} / \ac{ASIC} Workflow} \\ \hline

\textbf{\ac{SLD}} 
& Algorithm modeling, dataflow simulation, partitioning (MATLAB/Simulink, SystemC, Python, PYNQ)
& Same high-level modeling and system exploration (MATLAB/Simulink, SystemC, C/C++) \\
\hline
\textbf{{\Ac{DSC}}} & {RL graph-level optimization, operator fusion, quantization, memory tiling, dataflow scheduling, accelerator mapping; possible generation of \ac{HLS}/\ac{RTL} or accelerator intermediate representation} & {Similar graph/dataflow-level compilation targeting \ac{ASIC}/\ac{SoC} accelerators; supports early exploration of data reuse, memory hierarchy, and hardware mapping} \\

\hline
\textbf{\ac{HLS}} 
& C/C++/SystemC $\rightarrow$ RTL (Vitis HLS, Intel HLS, CatapultC); directives for unrolling, pipelining, memory partitioning
& C/C++/SystemC $\rightarrow$ RTL (Cadence Stratus, Mentor Catapult, Synopsys HLS); early power/performance design-space exploration \\
\hline

\textbf{\ac{RTL}} 
& Verilog/VHDL/SystemVerilog RTL for datapaths, \ac{FSM}s, controllers; synthesizable design for  \acp{LUT}, FFs, \ac{DSP}s, BRAMs
& Verilog/SystemVerilog RTL for logic synthesis, optimization, timing-driven transformations \\
\hline

\textbf{Gate-Level} 
& FPGA-specific placement, routing, timing closure (Vivado, Quartus); bitstream generation for predefined fabric 
& ASIC physical design including floorplanning, placement, routing, CTS, STA, IR drop analysis, DRC/LVS; GDSII tape-out \\
\hline
\end{tabular}
\end{table*}

\subsection{Platforms and Simulators}
Beyond conventional design environments such as MATLAB/Simulink, HDL Coder, and vendor-specific toolchains including AMD-Xilinx Vivado, a growing number of research-oriented simulators and platforms have been proposed to enable flexible integration of learning algorithms with hardware-aware evaluation. 
These platforms aim to decouple \ac{RL} agent development from low-level hardware implementation details, support rapid design-space exploration, and provide meaningful performance, energy, and latency feedback prior to physical deployment. For example, Chen et al. \cite{chen2023vpr} introduced \emph{VPR-Gym}, a Python-based platform built on top of the OpenAI Gym framework to enable systematic exploration of \ac{AI} techniques for \ac{FPGA} placement optimization. By decoupling learning agents from the underlying C++ implementation of the VPR toolchain, VPR-Gym lowers the barrier for integrating modern \ac{RL}, evolutionary strategies, and other AI-based optimizers into the \ac{FPGA} computer-aided design (CAD) flow. The platform supports multiple learning paradigms, flexible reward formulations, and parallel environment execution, making it particularly suitable for rapid experimentation and comparative evaluation of placement strategies while preserving compatibility with industrial-grade \ac{FPGA} CAD tools. Beyond \ac{FPGA} placement, simulation frameworks targeting emerging hardware paradigms have also been proposed. Rakka et al. \cite{rakka2025hdrlpim} presented HDRLPIM, a cycle- and energy-aware simulator for hyper-dimensional \ac{RL} implemented on processing-in-memory (PIM) architectures. The simulator models associative processor-based accelerators using SRAM and ReRAM technologies and focuses on accelerating inference in hyper-dimensional \ac{QL}. By abstracting the hardware execution of core \ac{RL} operations—such as Q-value computation and action selection—HDRLPIM enables detailed evaluation of latency, energy consumption, and energy--delay product, offering a valuable tool for early-stage design-space exploration of non-von-Neumann \ac{RL} accelerators. In the context of power electronics, Xu et al. \cite{xu2023novel} proposed an \ac{FPGA}-based circuit simulator designed to accelerate \ac{RL}-driven power converter design. The simulator offloads time-domain circuit simulation to an \ac{FPGA} accelerator deployed on cloud-based platforms, significantly reducing the training time of \ac{RL} agents compared to conventional software simulators. By packaging the accelerator within an OpenAI Gym-compatible environment, the framework enables seamless integration with standard \ac{RL} algorithms while allowing on-the-fly modification of circuit parameters without regenerating bitstreams. This work highlights the importance of hardware-aware simulators in making \ac{RL}-based design methodologies practical for real-world engineering applications. Complementary to these simulation platforms, Canese \emph{et al.} \cite{canese2022automatic} presented a MATLAB/Simulink-based toolchain that automatically generates \ac{FPGA} intellectual property (IP) cores for tabular \ac{QL} accelerators. The framework produces synthesizable \ac{VHDL} code and integrates directly with the AMD-Xilinx Vivado flow, enabling users to customize Q-table dimensions, numerical precision, and target \ac{FPGA} devices. By leveraging HDL Coder, the tool reproduces established hardware architectures while significantly reducing development time and facilitating design-space exploration. Although the framework demonstrated full compatibility with existing \ac{QL} accelerators, support for Intel \ac{FPGA} platforms and \ac{ASIC} design flows was identified as future work.

\section{Taxonomy of RL-based FPGA and IC platform}
\label{sec3}

This section provides the essential concepts for understanding \ac{RL}, a taxonomy of existing \ac{RL}/\ac{DRL} techniques, and references to works that have already applied \ac{RL} in \ac{FPGA} accelerators within each category. A detailed conceptual distinction between \ac{RL} and \ac{DRL} is presented in the taxonomy of previous survey~\cite{kheddar2024reinforcement}. In the present paper, the term \ac{RL} is used in a broad sense that encompasses both classical and deep \ac{RL} methods, as they share the same underlying \ac{MDP} formalism and notation adopted here.

A wide variety of \ac{RL} models have been deployed on \ac{FPGA} platforms, spanning from classical tabular methods to deep, continuous-control and multi-agent schemes. To structure this design space, we propose a taxonomy that organizes RL-based \ac{FPGA} platforms according to the underlying learning paradigm (model-based \acp{MDP}, value-based methods such as \ac{QL} and \ac{DQN}/\ac{DDQN}, policy-gradient and \ac{AC} families including \ac{PPO} and \ac{FPGA}, \ac{MC} approaches, and \ac{MARL}), as well as their target problem scenarios, hardware advantages, limitations, and real-time capabilities. This taxonomy highlights how different algorithms map to distinct reconfiguration, scheduling, and power–performance trade-off problems in heterogeneous fabrics, and it guides designers in choosing an appropriate \ac{RL}  backbone for a given application class, as summarized in Table~\ref{tab:RLFPGAcompar}.

\renewcommand{\arraystretch}{1.30}
\begin{table*}[ht!]
\centering
\caption{Refined comparison of \ac{RL}- based \ac{FPGA} algorithms covering key applications} \label{tab:RLFPGAcompar}
\scalebox{0.78}{
\begin{tabular}{m{1cm}m{4cm}m{4cm}m{4cm}m{4.5cm}m{3cm}}
\hline
\textbf{Model} & \textbf{Problem scenario  suitable for} & \textbf{Advantages} & \textbf{Limitations} & \textbf{Typical applications} & \textbf{Real-time capability} \\
\hline
\textbf{MDP} & 
Model-based control with known or identifiable dynamics; 
small/medium discrete $S,A$ &
Interpretable, analyzable; 
formal guarantees on optimal policy and safety &
Requires accurate transition model; 
poor scalability, mainly offline planning &
Test-configuration generation, 
interconnect coverage, 
coarse \ac{DVFS} or power-state selection &
\textbf{Low--moderate} 
(offline planning, light online lookup) \\
\hline
\textbf{QL} & 
Discrete, low-dimensional state/action spaces; 
no explicit model available &
Simple tabular updates; 
easy to implement in fixed-point hardware &
Explodes with large $S,A$; 
slow exploration and convergence &
Mode selection for \ac{DVFS}, 
cache / memory policy, 
reconfiguration of small regions &
\textbf{Moderate} 
(for compact tables in BRAM) \\
\hline
\textbf{SARSA} &
Streaming control tasks requiring smooth policy updates; 
tasks sensitive to abrupt action changes &
Stable incremental learning; 
updates follow executed behavior; 
well-suited for hardware-friendly \ac{TD} updates &
Still tabular with limited scalability; 
performance depends strongly on exploration strategy &
Thermal/power throttling, 
adaptive subsystem tuning, 
sequential reconfiguration tasks &
\textbf{Moderate}
(lightweight updates suitable for real-time streams) \\
\hline
\textbf{DQN} & 
High-dimensional state, discrete actions; 
rich sensor/monitor signals on \ac{FPGA} &
Handles large $S$ with deep nets; 
good for complex policies &
Training cost, replay memory, 
stability and quantization issues on-chip &
Placement / routing guidance, 
task scheduling, \ac{NoC} routing, 
adaptive accelerator tuning &
\textbf{High for inference} 
(training mostly off-chip) \\
\hline
\textbf{DDQN} & 
Same as \ac{DQN} where value overestimation 
hurts reliability &
Reduces Q-overestimation; 
more stable value estimates &
Still heavy training; 
needs two networks and memory resources &
Latency–power trade-off control, 
adaptive mapping of kernels, 
thermal-aware control &
\textbf{High for inference} 
(training offline, batched) \\
\hline
\textbf{AC} & 
Mixed continuous/discrete control; 
need both policy and value estimates &
Supports continuous actions; 
good sample efficiency vs. pure policy gradient &
Coupled \ac{AC} training; 
sensitive to hyperparameters &
Fine-grained \ac{DVFS}, 
analog front-end tuning, 
throughput–latency control &
\textbf{Moderate--high} 
(inference light, training heavier) \\
\hline
\textbf{Monte-Carlo} & 
Episodic tasks with clear episode end; 
small to medium spaces &
Unbiased return estimates; 
conceptually simple, few assumptions &
Requires full episodes; 
high variance, slow convergence &
Offline tuning of configuration policies, 
design-space exploration of mappings &
\textbf{Low} 
(primarily offline learning) \\
\hline
\textbf{PPO} & 
Complex, high-dimensional problems 
with continuous or discrete actions &
Stable policy-gradient updates; 
amenable to parallel training &
On-policy, sample-inefficient; 
training expensive for embedded devices &
Global power–performance optimisation, 
multi-objective resource management, 
\ac{DPR} &
\textbf{High for inference} ( 
training off-chip / batch mode) \\
\hline
\textbf{DDPG} & 
Continuous-action control with 
fine-grained knobs (voltages, clocks, gains) &
Handles continuous $A$; 
off-policy with replay buffer &
Very sensitive to noise, 
exploration, and normalization &
Analog tuning, \ac{DVFS} curve shaping, 
continuous parameterized routing or 
power-gating levels &
\textbf{High for inference} ( 
training typically off-chip) \\
\hline
\textbf{MARL} & 
Large \acp{FPGA} decomposed into regions/tiles; 
distributed decisions needed &
Scales with fabric size; 
local agents close to hardware, 
supports modular training &
Non-stationarity, coordination overhead; 
global optimality not guaranteed &
Per-region \ac{DVFS}, 
\ac{NoC} traffic management, 
multi-tenant \ac{FPGA} resource sharing &
\textbf{High} (light local agents; 
training mostly offline) \\
\hline
\end{tabular}}
\end{table*}

{Table~\ref{table:4} provides a comparative summary of representative \ac{RL}- based \ac{FPGA}/\ac{IC}  proposed schemes and their main implementation and performance characteristics. A qualitative comparison with \ac{CPU}/\ac{GPU}- based implementations further clarifies the practical role of \ac{RL}- based \ac{FPGA}/\ac{IC}  systems. As shown by the summarized works, the reported gains are often measured relative to \ac{CPU}/\ac{GPU} baselines, mainly in terms of speedup, throughput, or power efficiency. In general, \acp{GPU} remain highly effective for flexible development and large-scale \ac{RL} training because of their mature software ecosystems and strong support for massively parallel computation. In contrast, \ac{FPGA} and \ac{IC} implementations are generally more advantageous in latency-critical and energy-constrained deployment settings, where deterministic timing and workload-specific customization are important. However, these benefits are often accompanied by higher design complexity, tighter resource and memory constraints, and lower post-deployment flexibility. Hence, the surveyed works indicate that \ac{GPU} \ac{FPGA}, and \ac{IC} platforms are best regarded as complementary solutions rather than direct substitutes.}

\begin{table*}[ht]
\centering
\caption{Overview of representative \ac{RL}- based \ac{FPGA}/\ac{IC}  proposed schemes, highlighting hardware platforms, resource consumption, implementation level, performance results, and limitations.}
{\fontsize{6.5}{7}\selectfont
\begin{tabular}{
m{0.2cm}  
m{0.5cm}  
m{1.8cm}  
m{0.5cm}  
m{0.5cm}  
m{0.8cm} 
m{0.5cm}  
m{0.5cm}  
m{0.1cm}  
m{0.1cm}  
m{0.5cm}  
m{0.5cm}  
m{0.3cm}  
m{1.5cm}  
m{4cm}  
}
\label{table:4} \\
\hline
\multirow{2}{*}{\textbf{Ref}} & \multirow{2}{*}{\textbf{RL}} & \multirow{2}{*}{\textbf{Hardware}} & 
 \multicolumn{5}{c}{\textbf{Resource consumption (\%)}} & \multirow{2}{*}{\textbf{N}} & \multirow{2}{*}{\textbf{A}} & \multirow{2}{*}{\textbf{CLK}} & \multirow{2}{*}{\textbf{PW}} 
& \multirow{2}{*}{\textbf{IL}} 
& \multirow{2}{*}{\textbf{Best result}} & \multirow{2}{*}{\textbf{Limitations}} \\\cline{4-8}
 & Used & & \ac{BRAM} & LUTS & LUTRAM  & FF & DSP  &  & &  &     &   & \\
\hline
\hline

\cite{leiva2022fpga} & MDP & Zynq-7000 & -- & 15.35 & -- & 1.85 & -- & 8 & -- &  100 & -- 
  & HLS 
& 7.8$\times$ acceleration 
& HLS MDP generator; limited scope; no BRAM/DSP data  \\\hline

\cite{liu2022fpga} & QL & Zynq UltraScale+ MPSoC & -- & 0.056 & 0.010 & 0.012 &  0.231 & 8&  4 &   240& 0.038     & RTL & Usage: -50\% in resource  & Limited to small action spaces, offline training, no large-scale benchmarks. \\\hline

\cite{spano2019efficient} & QL & Zynq UltraScale+ MPSoC & -- & 0.08 & 0.03&  0.03 & 0 & 8 &  4 & 222 & 0.037    & RTL & --  & Uses small Q-tables, limited exploration strategies, simplified multiplier logic, and no dynamic environment testing.  \\\hline

\cite{sutisna2022architecture} & QL  & PYNQ-Z1 &  24.29 & 9.77 &  1.90 &  6.34 & 0 & 32 & 4 &   100 & 1.75     & RTL  & 1400$\times$ speedup & Limited state modeling, no multi-intersection support, and high \ac{BRAM} usage for large environments.\\\hline

\cite{sutisna2022farane}  & QL & Zynq MPSoC & 4.63 & 7.03 & 1.12 & 4.19 & 0.28 & 16 & 4 &  125 & 67  & RTL & 
Throughput= 148.55 MSps & 
Requires \ac{FPGA} PL; performance depends on N and \ac{BRAM} growth \\\hline

\cite{sutisna2022reinforcement}  & QL & Ultra96-V2 & 9.26 & 1.45 & 0.09 & 0.61 & 0 & 16 & 4 &  100 & --  & RTL & Usage: -40\% LUT, -14\% FF & Scalability limited by N; \ac{DSP}-free design; power results missing.\\\hline

\cite{da2018parallel} & QL & Virtex-6 \newline (6×4 Q-table)& --  & 1 & --  & <1 & 4 & 10 & 4 &  26.42 & 0.003  & SLD & Throughput= 26.42 MSps & Evaluated only on a few robot tasks; generalization remains untested.\\\hline

\cite{zhang2023deep} & \ac{DQN} & Virtex-7 VC707  & 32.71 & 67.41  & -- & 36.65 & 14.96 & 32 & 2 &  200 & 7.36 & RTL & 19.05$\times$ faster than GPU & High  resource usage, limited scalability, and dependence on external host.\\\hline

\cite{rothmann2024fpga} & \ac{DQN} & Versal ACAP VCK190  & -- & 25  & -- & 13 & 27 & 32 & 2 &  312.5 & -- & HLS & 2.5$\times$ speed-up than CPU & High \ac{BRAM} demand, four weight copies, external host dependence, limited scalability.\\\hline

\cite{le2022fpga} & \ac{DQN} & Virtex-7 VC707  & -- & 4.64  & 0.39 & 1.95 & -- & 32 & 2 &  172.41 & 0.363  & RTL & Max reward ($\sim$200 points) & Small network, slow convergence, limited scalability, and  increased complexity.\\\hline

\cite{sumarudin2025low} & \ac{DQN} & Ultra96-V2  & 1.43 & 22.42  & 4.25 & 12.95 & 1.11 & 32 & 4 &  70 & 2.061 & RTL & 671$\times$ and 287$\times$ inference and train speedups  & Limit scalability; maze environment small and non-generalizable.\\\hline

\cite{nai2022design} & \ac{DQN} & PYNQ-Z1  & -- & 0.93  & -- & 0.22 & 2.27 & 8 & 4 &  -- & -- & HLS & 1.84$\times$ speedup & Small maze task, tiny network, no timing data, no power or clock reported.\\\hline

\cite{cho2019fa3c} & A3C & UltraScale+ VU9P  & 40.6 & 57.3  & -- & 37 & 34.3 & 32 & 18 &  180 & 18 & RTL & 1.62$\times$ better than GPU & Hard to adapt to other \ac{RL} models; requires complex RTL design and tuning\\\hline

\cite{ge2024fpga} & A3C & Zynq-7000 SoC  & 72.1 & 75.2  & -- & 46.8 & 24.6 & 32 & 4 &  200 & 10.03 & RTL & 2.39$\times$ speedups than GPU& High \ac{BRAM} demand limits scalability of distributed A3C design\\\hline

\cite{li2022a3c} & A3C & Virtex-7 XC7VX485T  & 39.4 & 77.8  & 15.4 &  43.6 & 42.7 & 32 & 4 & 200 & 29 & RTL & 54$\times$ speedup than GPU & FP32 computations increase power and restrict resource efficiency\\\hline

\cite{taha2024heppo} & PPO & Zynq UltraScale+ MPSoC ZCU106  & 9 & 4.69 & -- & 9.91 & 30.48 & 8 & -- &  300 & -- & RTL & 1.5$\times$ increase rewards, 30\% speedup & No measured energy efficiency, lacks broader platforms.\\\hline

\cite{meng2021ppoaccel}& PPO & Alveo U200  & 100 & 89  & -- & 89 & 62 & -- & -- &  285 & -- & RTL & Up to 26$\times$ throughput & High \ac{BRAM} usage and limited scalability for large models; Depend on CPU.\\\hline

\cite{waseem2023fpga} & PPO & Avnet Ultra96-V2 
& 6 & 27 & 2 & 11.2 & 9.7 & 32 & -- &  100 & 2.053 
& RTL 
& 98.321\% faster than CPU 
& Limited model size; constrained on-chip memory; moderate throughput \\ \hline

\cite{yang2021fixar}  & DDPG & Alveo U50  & 57.6 & 58.4  & -- & 23.5 & 38.8 & 32 & 2--6 &  164 & 20.4 & RTL & 2.7$\times$ higher throughput than CPU-GPU & Scalability remains restricted by memory bandwidth and resource usage.\\\hline

\cite{guo2024fpga} & DDPG & Intel Stratix-V  & 74.52 & 54.71  & -- & 60.65 & 97.45 & 32 & 5 &  200 & -- & RTL & 38$\times$ faster than CPU (PyTorch) & IO bandwidth bottleneck critically restricts full \ac{FPGA} acceleration.\\\hline

\cite{zhang2023framework} & DDPG & Alveo U200  & 35 & 46  & -- & 33 & 38 & 32 & 3 &  300 & -- & HLS & 1005$\times$  training throughput & Relies on full-system power, obscuring \ac{FPGA} efficiency analysis\\\hline

\cite{hu2022td3lite} & TD3 & Alveo U50  & 39.2 & 57.1 & -- & 21.4 & 38.6 & 12 & 3 &  160 & 24 & HLS & Energy eff.= 110.3 FIPS/W & BFP quantization reduces precision, risking learning performance degradation.\\\hline
{\cite{ge2025e2cap}} & DQN & Xilinx VCU118 & 92.31 & 47.95 & -- & 35.47 & 17.38 & 512 & -- & 120 & 17.32 
  & RTL 
& 33.01$\times$ (CPU), 17.72$\times$ (GPU)
& DRL-specific accelerator with host dependency, limited to DQN evaluation, and high BRAM usage \\\hline
{\cite{bazmalah2026resource}
} & QL & Kintex-7 \newline (Genesys 2) & -- & 0.35 & 0.10 & 0.19 & -- & 1024 & -- & 73 & 36 
  & RTL 
& 
816.4 MSPS; 4.3$\times$ speedup 
& 
Q-table scalability limits; $\epsilon$-greedy local optima; Grid-World-only evaluation \\\hline
\end{tabular}}
\begin{flushleft}
\begin{scriptsize}
\textbf{Abbreviation:} Clock (CLK) [MHz]; Implimentation level (IL); Power (PW) [W]; Not available (NA); Number of bits (N); Action (A)    
\end{scriptsize}
\end{flushleft}
\end{table*}

\subsection{MDP}

An \ac{MDP} $(S,A,P,R,\gamma)$ formalizes sequential decision-making for hardware-aware \ac{RL}, where $s \in S$ compactly represents the current hardware and workload configuration (e.g., buffer utilization, timing slack, power states) and $a \in A$ encodes reconfiguration choices such as clock scaling, datapath parallelism, resource allocation, or power-management tuning. The transition kernel $P(s' \mid s,a)$ models stochastic workload dynamics, noise, and hardware variability, while the reward function $R(s,a)$ aligns learning with design goals including latency minimization, energy and performance-per-watt optimization, throughput maximization, and accuracy or error-rate targets; the discount factor $\gamma$ balances short- versus long-term performance. On \acp{FPGA}, \ac{MDP}-driven \ac{RL} implementations exploit fine-grained reconfigurability, parallelism, pipelining, low-precision arithmetic, and optimized memory hierarchies, whereas \ac{IC}-based realizations emphasize area efficiency, deterministic timing, and ultra-low energy operation. Across both platforms, the $(S,A,P,R,\gamma)$ formalism supports rigorous analysis of convergence, stability, and safety, making \ac{MDP}-guided \ac{RL} a robust paradigm for real-time, self-optimizing \ac{FPGA} and \ac{IC} architectures. 

The agent is built around an \ac{RL} controller that interacts with the environment and a Planner/Policy block. After an action $a_t$ is applied, the environment produces the next state $s_{t+1}$ and reward $r_{t+1}$. The state $s_{t+1}$ is forwarded to the Planner/Policy block, which uses it to select the next action $a_{t+1}$. In parallel, a delay element $z^{-1}$ stores $(s_{t+1}, r_{t+1})$ so that, at the following control cycle, the controller receives the previous transition pair as $(s_t, r_t)$. Together with the discount factor $\gamma$ and learning rate $\alpha$, these signals are used by the controller to update its value functions and/or policy parameters. The updated policy is then made available to the Planner/Policy block, which outputs the action to be applied to the environment, thereby closing the \ac{MDP} control loop.

As a preliminary contribution, the paper \cite{ruan2019reinforcement} adopted a model-based \ac{RL} perspective on the nondeterministic polynomial-time complete  problem of full interconnect coverage in \ac{SRAM} \acp{FPGA}. It formulated configuration generation as a \ac{MDP} whose states are programmable interconnect point  terminals and actions select wire-type transitions over a generic interconnect model. Using \ac{DP}, the approach computed optimal state values and policies that yield near-theoretical-minimum configuration sets with 100\% interconnect resource coverage, replacing ad hoc graph heuristics with mathematically grounded planning. However, the scheme faced high offline computation, limited scalability, and no online adaptive deployment. Extending the discussion, the paper \cite{leiva2022fpga} tackles slow sparse \ac{MDP} model construction on \acp{CPU} by offloading AI-Toolbox’s \texttt{SparseModel()} routine to a  \ac{FPGA} using Vivado \Ac{HLS}. The accelerator generates nonzero transition and reward entries, exploiting loop flattening, pipelining, fixed-point arithmetic, and AXI-Stream interfaces to fill large sparse matrices efficiently. The design treats \acp{FPGA} as generic, reusable \ac{MDP}/\ac{RL} coprocessors tightly coupled to high-level libraries, prioritizing portability and development productivity over deeply customized, application-specific \ac{RL} accelerators. It bridges algorithmic \ac{RL}  frameworks and hardware efficiency.

\subsection{Q-learning}

Unlike \ac{MDP}-based planning, which assumes access to the transition model, \ac{QL} focuses on estimating the optimal action–value function $Q^\ast(s,a)$ without requiring an explicit model of $P$ or $R$. In hardware-aware settings, a tabular or compactly parameterized $Q(s,a)$ is stored in on-chip memories (e.g., \ac{BRAM}/\ac{LUTRAM}) and updated according to Equation \ref{eqQL}: 
\begin{equation}
\label{eqQL}
Q_{t+1}(s,a) = Q_t(s,a) + \alpha \bigl[r + \gamma \max_{a'} Q_t(s',a') - Q_t(s,a)\bigr],  
\end{equation}

\begin{figure}[ht!]
    \centering
    \includegraphics[width=0.65\linewidth]{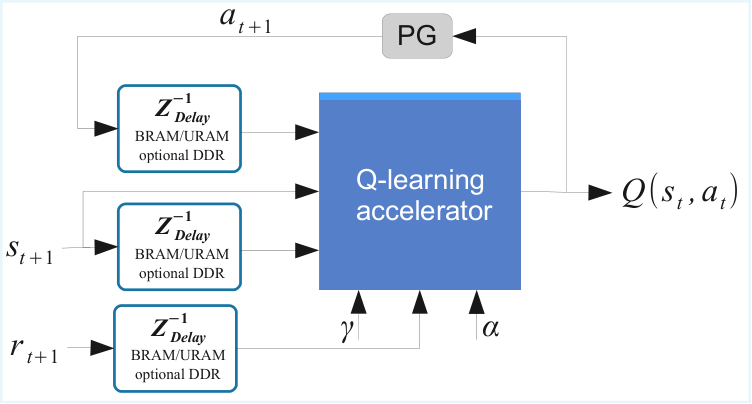}
    \caption{High-level architecture of the \ac{QL} agent mapped to \ac{FPGA}/\ac{IC}  hardware.
The \ac{QL} accelerator executes the tabular update rule using \ac{DSP}-based MAC units
and fixed-point arithmetic, with Q-values stored in BRAM/URAM. The learning rate
$\alpha$ governs the magnitude of the parameter update, while the discount factor
$\gamma$ scales the contribution of future reward terms. The \ac{PG},
implemented with \ac{LUT} logic and a lightweight pseudo-random module, produces the next
action $a_{t+1}$ to support stochastic exploration. Temporal elements $z^{-1}$
act as register-based pipeline delays to synchronize $s_t$, $a_t$, and $r_t$
in hardware. The architecture highlights a practical mapping of \ac{QL} operators
to \ac{FPGA} resources (\ac{DSP}, \ac{LUT}, BRAM/URAM) without enforcing a device-specific
implementation~\cite{spano2019efficient}.}
    \label{fig:QL}
\end{figure}

\noindent Where $\alpha$ is the learning rate. On \acp{FPGA}, this update naturally maps to fixed-point arithmetic pipelines and parallel comparators for the $\max_{a'}$ operation, while the table lookups are accelerated via multi-port memory and carefully scheduled access patterns. The $\varepsilon$-greedy or Boltzmann exploration policy can be realized as a lightweight control \ac{FSM}, enabling on-line adaptation of \ac{DVFS} modes, region-level reconfiguration, or scheduling decisions with modest resource overhead. Compared to purely model-based \ac{MDP} planning, \ac{QL} trades analytical guarantees for simpler hardware implementation and the ability to learn directly from runtime measurements when transition probabilities or reward statistics are difficult to characterize a prior. Figure \ref{fig:QL} summarize \ac{QL} working principle.  

The \ac{QL} agent shown in Figure \ref{fig:QL} is composed of two
main blocks: the \ac{PG} and the \ac{QL}
accelerator \cite{spano2019efficient,sutisna2022architecture}. The accelerator receives the current state $s_t$ and the reward $r_{t+1}$ from the environment, together with
the next state $s_{t+1}$ delayed by one time step through the $z^{-1}$ registers. Internally, it updates the Q-matrix according
to the chosen learning rule, using the discount factor $\gamma$ and learning rate $\alpha$ provided as external parameters.
The updated Q-values are then read by the \ac{PG}, which implements an action-selection strategy and outputs the next
action $a_{t+1}$ to be applied to the environment. Because the \ac{PG} is application-independent, it can be implemented either in software or hardware. Within this generic template, several \ac{FPGA}-based \ac{QL} accelerators have explored different points in the design space in terms of resource usage, latency, and flexibility.

At the low-resource end of this spectrum, \cite{liu2022fpga} proposed a serial \ac{FPGA} accelerator for \ac{QL} that targets low resource consumption and deterministic timing. The authors implemented the design using Vivado \ac{HLS}, achieving a significant reduction in logic usage compared with a previous parallel architecture while fixing the Q-update latency to four cycles. The modular design with RAM-based Q-tables and LFSR-driven random policy is therefore well suited to embedded, real-time agents. Moving toward higher-throughput, fully parallel architectures, the paper \cite{spano2019efficient} addressed the problem of executing \ac{QL} efficiently in hardware, given that conventional \ac{CPU}- and \ac{GPU}-based implementations suffered from high power consumption and limited real-time performance. It proposed an \ac{FPGA}-based accelerator that used distributed action \acp{RAM}, a max-tree for rapid action selection, and an optimized Q-Updater that reduced multiplier usage through power-of-two arithmetic. The design achieved superior speed and energy efficiency, making the resulting accelerator suitable for embedded, \ac{IoT}, and multi-agent real-time \ac{RL} applications. Extending these ideas to heterogeneous \ac{SoC} platforms, the paper \cite{sutisna2022farane} addressed the challenge of executing \ac{QL} under strict real-time and energy constraints, where fast state–action updates and scalable memory structures were crucial. To overcome this, the authors introduced FARANE-Q, a heterogeneous architecture that exploited the dual nature of Zynq devices by combining a \ac{SoC}-based \ac{PS} for configuration and control with \ac{FPGA} \ac{PL} for hardware acceleration. Their design restructured \ac{QL} through parallel and pipelined computation in the \ac{PL}, supported by approximate multipliers, BRAM-based Q-storage, and dynamic $\varepsilon$-greedy exploration. This coordinated \ac{PS}–\ac{PL} strategy enabled efficient, scalable, and reconfigurable \ac{RL} execution for embedded intelligent systems. Complementary to these architectural innovations, \cite{sutisna2022reinforcement} focused on a hardware-optimization strategy centered on numerical precision. Whereas earlier \ac{FPGA}-based \ac{QL} accelerators typically relied on full-precision fixed-point arithmetic, neural-network–based Q-value estimation, or approximate multipliers without systematic precision exploration, this work examined how fractional and integer bit-widths affected Q-value accuracy and showed that reducing the datapath from 32 to 16 bits preserved optimal policy behavior. Their approach integrated precision tuning with barrel-shifter–based approximate multipliers, enabling a \ac{DSP}-free Q-Updater and substantial reductions in \ac{LUT} and \ac{FF} usage.

\subsection{SARSA}

Unlike \ac{QL}, which was an off-policy method relying on the greedy estimate $\max_{a'}Q(s',a')$, \ac{SARSA} performed an on-policy update in which the next action $a'$ was sampled from the agent’s behavior policy (e.g., $\varepsilon$-greedy). Its update rule was:
\begin{equation}
Q_{t+1}(s,a) = Q_t(s,a) + \alpha \bigl[r + \gamma Q_t(s',a') - Q_t(s,a)\bigr],
\end{equation}
where $(s,a,r,s',a')$ denoted the full interaction tuple. The absence of the $\max_{a'}$ operator made \ac{SARSA}’s bootstrapping dependent on the \emph{actual} executed action rather than a hypothetical greedy one.

From a hardware perspective, this distinction had several implications. Eliminating the $\max_{a'}$ comparator tree simplified the datapath and reduced critical-path delays, \ac{LUT} footprints, and \ac{DSP} utilization—an advantage for deeply pipelined fixed-point accelerators on \acp{FPGA} and resource-constrained \ac{ASIC} \ac{SoC} s. The update logic became dominated by a single table lookup for $Q(s',a')$ instead of a parallel scan across all actions, which enabled smaller multi-port \ac{BRAM}/\ac{LUTRAM} banks or even banking-free single-cycle access in low-action-count settings. Furthermore, the on-policy nature required the \ac{PG} to expose the selected action $a'$ back to the learner, creating a tighter coupling between the \ac{PG} and the update engine. This feedback was typically implemented through a lightweight handshake interface or shared memory queue, and, unlike in \ac{QL}, the \ac{PG} could not be fully decoupled from the update hardware.

Recent \ac{FPGA} accelerators reinforced these advantages. QTAccel \cite{meng2020qtaccel} demonstrated that \ac{SARSA} reused exactly the same pipeline stages as \ac{QL}—address generation, memory banks, and \ac{ALU} pipelines—requiring only that the policy-selected action be forwarded to the update stage. This eliminated the need for additional action-selection logic and greedy comparator trees, thereby reducing control complexity and easing timing closure as the number of actions increased. QTAccel also showed that \ac{SARSA} introduced no additional \ac{LUT}, \ac{DSP}, or \ac{BRAM} overhead, and that its update latency remained stable and independent of action-space width. Moreover, \ac{SARSA} aligned naturally with architectures in which the environment or policy logic was integrated into the same hardware substrate. This is exemplified by the FAQ accelerator \cite{rothmann2022faq}, which implemented environment dynamics and Q-updates within a configurable hardware framework, provided such an example. Because transitions, rewards, and actions were produced directly on the \ac{FPGA}, the executed action $a'$ could be forwarded immediately to the learner. This coherence between execution and learning avoided the divergence that off-policy methods like \ac{QL} had to correct, simplified scheduling, supported deterministic replay-free operation, and reduced interface traffic.

This analysis prove that \ac{SARSA} offered simpler datapaths, action-count-independent update latency, and tighter coupling between policy execution and learning, making it an attractive alternative to \ac{QL} for tightly integrated \ac{FPGA}/\ac{IC}-based Q-table \ac{RL} systems.

\subsection{Deep Q-learning}

Unlike tabular \ac{QL}, which directly stores and updates a discrete table of $Q(s,a)$ values, a \ac{DQN} approximates the action–value function using a deep neural network $Q(s,a;\theta)$ and learns its parameters from sampled transitions. This removes the need for an explicit transition model while enabling scalability to high-dimensional state spaces. Instead of the single-step update in Equation~\ref{eqQL}, the \ac{DQN} agent minimizes a \Ac{TD} loss over minibatches extracted from an experience replay buffer:
\begin{equation}
\mathcal{L}(\theta) = \mathbb{E}{(s,a,r,s')\sim \mathcal{D}}
\Bigl[ \bigl(r + \gamma \max{a'} Q(s',a';\theta^{-}) - Q(s,a;\theta)\bigr)^2 \Bigr],
\end{equation}
where $\theta$ are the parameters of the main Q-network and $\theta^{-}$ are the periodically updated parameters of the target network.

\noindent On hardware platforms such as \acp{FPGA}, this architecture introduces new building blocks compared with tabular \ac{QL}. Neural network forward propagation is mapped to pipelined matrix–vector multipliers, activation blocks, and quantized arithmetic units, while the experience replay buffer is implemented using on-chip or hybrid BRAM/\ac{DDR} circular memory structures that support random minibatch sampling. The target Q-network is instantiated as a secondary neural module that shares the same topology but updates at a slower rate, improving training stability. The exploration policy (e.g., $\varepsilon$-greedy) remains a lightweight control \ac{FSM}, but action selection now operates on the network’s vector of Q-values rather than table entries.

\begin{figure}[!htbp]
    \centering
    \includegraphics[width=0.7\linewidth]{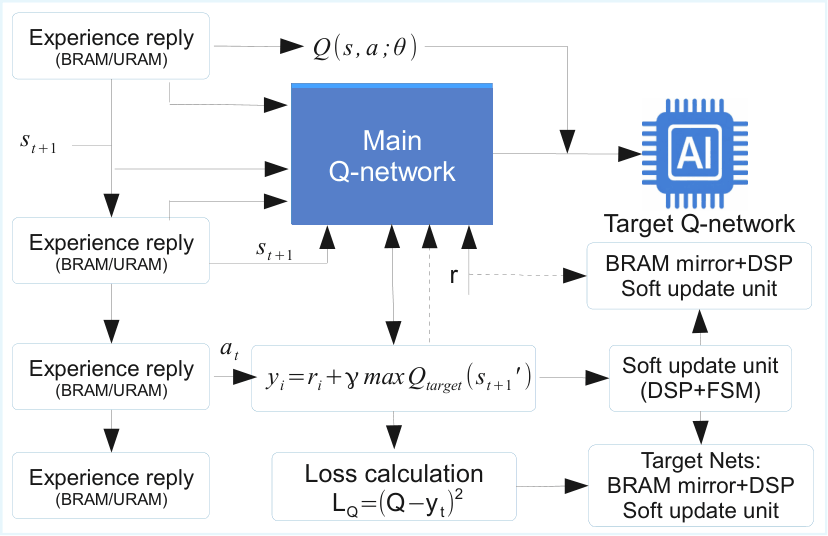}
    \caption{High-level architecture of a \ac{DQN} agent on \ac{FPGA}/\ac{IC} . The Main Q-network is
implemented using \ac{DSP} arrays and fixed-point MAC units, with parameters stored in
BRAM/URAM. Experience replay is mapped to BRAM/URAM or external DDR depending on
capacity. The Target Q-network is maintained as a \ac{BRAM} mirror and updated via a 
soft-update unit (\ac{DSP} + \ac{FSM}). The loss $L=(Q-y)^2$ is computed using the \ac{TD} target
$y=r+\gamma Q_{\text{target}}(s_{t+1},\arg\max_a Q(s_{t+1},a))$. }
    \label{fig:DQN}
\end{figure}

\noindent Compared with hardware-friendly \ac{QL}, which relies mainly on fixed-point comparators and simple update operators, \ac{DQN} trades architectural simplicity for generalization capability and expressiveness—enabling learning directly from complex, high-dimensional sensory streams. Figure~\ref{fig:DQN} illustrates the high-level structure of the basic \ac{DQN} agent, highlighting its three core components: the main Q-network, the target network (annotated with an AI hardware icon), and the experience replay buffer feeding minibatch samples to the learning pipeline.

Beyond the algorithmic principles of \ac{DQN}, recent FPGA-based implementations demonstrate how the learning pipeline can be decomposed, parallelized, and quantized to meet the constraints of embedded and edge platforms. While the baseline \ac{DQN} formulation requires forward inference, loss evaluation, backpropagation, and periodic target updates, mapping these stages to hardware has motivated several architectural innovations. A representative example is the heterogeneous \ac{DQN} accelerator of Zhang et al. \cite{zhang2023deep}, which divides the training process into five strictly separated computing stages—forward pass, loss evaluation, error backpropagation, gradient generation, and parameter update—each implemented as an independent hardware unit. This phase-specialized decomposition eliminates resource conflicts and improves pipeline utilization, especially when floating-point arithmetic is required. Their design exploits \ac{FPGA} IP cores for 32-bit floating-point \acp{MAC} and adopts multi-mode matrix multipliers to reduce the cost of weight transposition. The result is a heterogeneous dataflow pipeline capable of supporting full \ac{DQN} training while outperforming \ac{CPU} and \ac{GPU} baselines in energy efficiency. A different approach is adopted in Rothmann et al.’s STANN-RL accelerator \cite{rothmann2024fpga}, which focuses on modularity and reusability of \ac{DQN} components. Here, neural-network training blocks (dense layers, optimizers, minibatch generators) are drawn from the STANN library and extended with \ac{RL}- specific operators such as target-network synchronization and prioritized minibatch sampling. Implemented on a Xilinx Versal device and interfaced with Gymnasium environments over Ethernet, the accelerator supports low-latency training with batch sizes typical of \ac{DQN} (often as small as 1–8). Notably, the inclusion of an \ac{FPGA}-based Adam optimizer improves convergence compared with traditional fixed-point stochastic gradient descent (SGD), while maintaining a speed-up than \ac{CPU} execution. \ac{FPGA}-based \ac{DQN} has also been explored in the context of pedagogical or benchmark environments. Tran et al. \cite{le2022fpga}, provide a full training and inference pipeline for the Cartpole task, using 32-bit floating-point arithmetic and Cocotb-based HW/SW co-simulation. Their design highlights the importance of careful scheduling of neural-network operations, since even a small network demands efficient pipelined \ac{MAC} arrays and memory interfaces to sustain real-time training throughput. Although the architecture is less optimized than more recent designs, it exposes essential hardware building blocks—forward-prop datapaths, target-network shadow modules, and online gradient-update logic—that underlie almost all later \ac{DQN} accelerators. More recent research targets resource-constrained edge platforms. Sumarudin et al. \cite{sumarudin2025low}, propose a dual-mode accelerator capable of supporting both inference and on-device training within a single configurable processing element. By reusing datapaths across forward and backward passes, employing piecewise-linear approximations for activation functions, and adopting fixed-point arithmetic with quantization-aware updates, the design achieves significant reductions in  \acp{LUT}, \ac{DSP}s, and power consumption. The accelerator—implemented on an Ultra96 platform—delivers remarkable speedups with a low power budget, illustrating how \ac{DQN} training can be pushed onto edge-class \acp{SoC}.

\subsection{DDPG}

\begin{figure}[!htbp]
    \centering
    \includegraphics[width=0.8\linewidth]{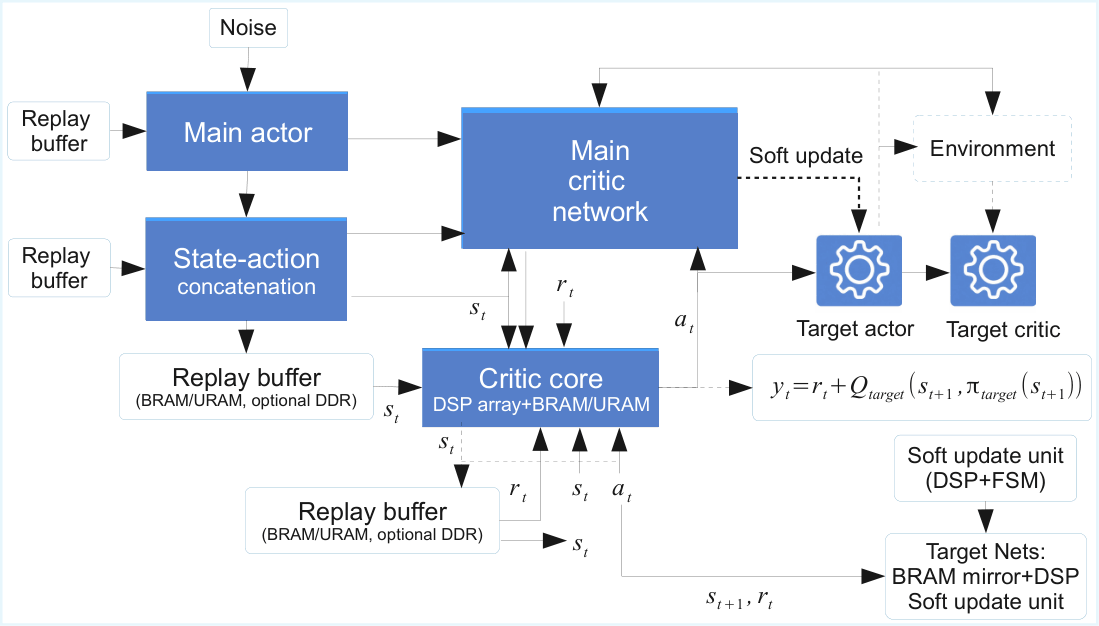}
    \caption{High-level functional architecture of a \ac{DDPG} agent. 
The Main Actor generates continuous control actions, with exploration noise injected during interaction with the environment. 
The Main Critic evaluates state--action pairs and computes the \ac{TD} target using the Target Actor and Target Critic, whose parameters are updated through soft (Polyak) averaging. 
A replay buffer enables off-policy learning by storing and supplying past transitions $(s_t, a_t, r_t, s_{t+1})$ for both actor and critic updates. 
State--action concatenation indicates where state and action vectors are combined prior to critic evaluation. 
\textbf{Solid lines} denote forward dataflow paths used during online interaction and gradient-based learning, whereas \textbf{dashed lines} indicate auxiliary or delayed pathways used exclusively for target value computation and parameter synchronization.
\textbf{Target networks} are passive, auxiliary models not used for online control, and are invoked only during TD-target evaluation and soft updates. 
Learning hyperparameters (e.g., $\alpha$, $\gamma$, $\tau$) are applied during loss computation and update rules and are not explicit network inputs.
}
\label{fig:ddpg}
\end{figure}

While \ac{DQN} extends tabular \ac{QL} to high-dimensional discrete control, \ac{DDPG} generalizes the same off-policy and target-network principles to continuous action spaces. Instead of computing $\arg\max_{a'}Q(s,a')$, \ac{FPGA} introduces a differentiable deterministic actor $\mu(s;\theta_\mu)$ whose output directly produces a continuous action. The critic $Q(s,a;\theta_Q)$ evaluates these actor-generated actions, forming the continuous-control analogue of \ac{DQN}. This eliminates discrete-action maximization and its wide comparator trees, but requires an \textit{action–state fusion} layer in the critic to concatenate the actor output with the state vector. These algorithmic differences impose clear architectural implications on hardware accelerators. First, the critic must evaluate $Q(s,a)$ for actor-generated actions, increasing \ac{MAC} count relative to discrete-action \ac{DQN} evaluators. Second, the actor and critic jointly participate in each training step: the critic learns via a \ac{TD} loss, while the actor receives gradients from $\nabla_a Q(s,a)$, requiring a dual forward–backward pipeline. Third, exploration is performed by injecting lightweight noise (e.g., OU or uniform) at the actor output, avoiding stochastic sampling logic or $\varepsilon$-greedy units. Figure \ref{fig:ddpg} illustrates the high-level architecture and learning principle of the \ac{DDPG} agent, showing the interaction between the actor, critic, and target networks.

Across \ac{FPGA} and \ac{IC} designs, these features make \ac{FPGA} highly amenable to acceleration. Replay memory can be stored in \ac{BRAM} or DDR using circular buffers, and the \ac{AC} pair can share matrix–vector engines through time-multiplexed scheduling because \ac{FPGA} does not require a discrete argmax. Furthermore, \ac{FPGA} typically uses smaller batch sizes than deep value-based algorithms, easing memory-bandwidth pressure and enabling deterministic low-latency compute pipelines.

Recent implementations further refined this mapping. FIXAR \cite{yang2021fixar} demonstrated that \ac{DDPG} benefited from fixed-point quantization-aware training, where data precision was reduced mid-training without degrading policy quality. This reduced arithmetic cost and enabled adaptive parallelism in \ac{FPGA} datapaths. Similarly, Guo and Luk’s \ac{FPGA}-accelerated sim-to-real platform \cite{guo2024fpga} showed \ac{DDPG}’s suitability for continuous-control robotics accelerators, where the gradient-computation pipeline was the dominant workload and achieved substantial speedups through \ac{FPGA}-based parallelism. In heterogeneous \ac{CPU}–\ac{GPU}–\ac{FPGA} settings, Zhang et al. \cite{zhang2023framework} included \ac{DDPG} as one of the algorithms supported in their prioritized replay framework, indicating that critic-centric training operations mapped particularly well to on-chip parallel sampling modules. Finally, \ac{TD3}-lite \cite{hu2022td3lite}—a successor of \ac{DDPG}—illustrated how structural and representation optimizations (shared networks, block floating-point) directly reduced the resource footprint inherent to \ac{AC} pipelines. {More recently, the authors  \cite{ramasamy2026deep} applied \ac{DDPG} to torque-ripple minimization in switched reluctance motors for electric-drive control, where the actor generated continuous firing-angle offsets and the critic guided policy updates from torque, current, speed, position, and flux-linkage feedback. }

Taken together, the hardware literature positioned \ac{DDPG} as a well-structured target for \ac{FPGA}/\ac{IC}  acceleration: its deterministic policy, moderately sized networks, absence of discrete maximization, and predictable update rules yielded balanced compute and memory characteristics suited to both single-accelerator and heterogeneous-system designs.

\subsection{Actor-critic}

Unlike value-based methods that update a single action--value estimator, \ac{AC} algorithms maintain two coordinated function approximators: an \emph{actor} network that represents a policy $\pi(a \mid s;\theta)$ and a \emph{critic} network that evaluates states through a value function $V(s;\phi)$ or an advantage estimate.
At each interaction step, action selection, value evaluation, and parameter updates are jointly executed, forming a tightly coupled learning pipeline that is well suited for hardware realization. The end-to-end learning sequence of a baseline \ac{AC} agent is summarized in Equation~\eqref{eq:AC-seq-hw}.
The actor performs policy inference to select an action, the environment returns the immediate reward and next state, and the critic evaluates the \ac{TD} error that drives both policy and value-function updates.

\begin{equation}
\label{eq:AC-seq-hw}
\begin{aligned}
s_t
&\xrightarrow{\text{state input}}
\pi_\theta(\cdot|s_t)
\;{\scriptstyle[\text{DSP/MAC array}]}
\xrightarrow{\text{action sampling}}
a_t \xrightarrow{\text{envir.}}
&(r_t, s_{t+1})
\xrightarrow{\text{critic evaluation}}
V_\phi(s_t),\,V_\phi(s_{t+1})
\;{\scriptstyle[\text{DSP/MAC array}]} \\
&\xrightarrow{\text{TD error}}
\delta_t
\;{\scriptstyle[\text{fixed-point ALU}]}
\xrightarrow{\text{gradient computation}}
(\nabla_\theta J_\pi, \nabla_\phi J_V)
&\;{\scriptstyle[\text{gradient engine}]} \xrightarrow{\text{parameter update}}
(\theta,\phi)
\;{\scriptstyle[\text{BRAM/URAM write-back}]}
\end{aligned}
\end{equation}

The interaction between actor and critic is governed by the \ac{TD} error is:

\begin{equation}
\label{eq:TDAC}
\delta_t = r_t + \gamma V_\phi(s_{t+1}) - V_\phi(s_t),
\end{equation}

\noindent which simultaneously drives value-function regression and policy-gradient updates. From a hardware perspective, Equation~\eqref{eq:TDAC} maps to a lightweight arithmetic unit that feeds both gradient paths, enabling synchronized learning with minimal control overhead.

{In a neuromorphic context, the authors in  \cite{matsushima2026novel} demonstrated that this tightly coupled \ac{AC} interaction can be efficiently realized using an ergodic sequential logic (ESL)-based \ac{SNN}, where both actor and critic are implemented as discrete-state neuron arrays with local update rules. Their \ac{FPGA} implementation shows that the ESL-based design preserves the functional behavior of conventional spiking \ac{AC} models while significantly reducing hardware cost in terms of logic elements and power. However, the evaluation is limited to small-scale navigation task. }

While classical \ac{AC} refers broadly to any architecture combining policy gradients with a value baseline, most hardware-oriented implementations adopt the \ac{A3C} formulation, which extends Equation~\eqref{eq:AC-seq-hw} to multiple asynchronous learners.
Each learner independently executes a local \ac{AC} pipeline and generates gradients that are asynchronously accumulated into shared global parameters, as formalized in Equation~\eqref{eq:A3C-seq-hw}.

\begin{equation}
\label{eq:A3C-seq-hw}
\begin{aligned}
\Big\{
& s_t^{(k)}
\xrightarrow{\text{local policy}}
\pi_{\theta_g}(\cdot|s_t^{(k)})
\;{\scriptstyle[\text{worker DSP pipeline}]}
\xrightarrow{\text{action}}
a_t^{(k)} \xrightarrow{\text{envir.}_k}
(r_t^{(k)}, s_{t+1}^{(k)})
\xrightarrow{\text{local critic}}
V_{\phi_g}(s_t^{(k)}),\,V_{\phi_g}(s_{t+1}^{(k)}) \xrightarrow{\text{advantage}}
A_t^{(k)}\\
&\;{\scriptstyle[\text{ALU}]}
\xrightarrow{\text{local gradients}}
(\nabla \theta^{(k)}, \nabla \phi^{(k)})
\Big\}_{k=1}^{K} \xrightarrow{\text{asynchr. accumulation}}
(\theta_g,\phi_g)
\;{\scriptstyle[\text{global update engine + BRAM/URAM}]}
\end{aligned}
\end{equation}

The advantage estimate computed by each learner is given by

\begin{equation}
\label{eq:A3C-adv}
A_t^{(k)} = r_t^{(k)} + \gamma V_{\phi_g}(s_{t+1}^{(k)}) - V_{\phi_g}(s_t^{(k)}),
\end{equation}

\noindent which mirrors the TD-error formulation in Equation~\eqref{eq:TDAC} while enabling fully asynchronous gradient generation across parallel workers.
In \acp{FPGA}, this structure naturally maps to multiple replicated compute pipelines sharing a global parameter memory without strict synchronization.

More recently, \ac{SAC} has emerged as an entropy-regularized extension of \ac{AC} learning that improves robustness and exploration.
In contrast to the on-policy sequences in Equations~\eqref{eq:AC-seq-hw} and \eqref{eq:A3C-seq-hw}, \ac{SAC} performs off-policy learning using a replay buffer, a stochastic Gaussian actor, and dual critic networks.
The resulting learning sequence is summarized in Equation~\eqref{eq:SAC-seq-hw}.

\begin{equation}
\label{eq:SAC-seq-hw}
\begin{aligned}
s_t
&\xrightarrow{\text{stochastic policy}}
\pi_\theta(\cdot|s_t)
\;{\scriptstyle[\text{DSP + RNG}]}
\xrightarrow{\text{action sampling}}
a_t\xrightarrow{\text{envir.}}(r_t, s_{t+1})
\xrightarrow{\text{replay storage}}
(s_t,a_t,r_t,s_{t+1})
\;{\scriptstyle[\text{BRAM/DDR buffer}]} \\
&\xrightarrow{\text{mini-batch sampling}}
\mathcal{B}
\;{\scriptstyle[\text{DMA + address generator}]}
\xrightarrow{\text{twin critics}}Q_{\phi_1}, Q_{\phi_2}\;{\scriptstyle[\text{parallel DSP pipelines}]} 
\xrightarrow{\text{entropy-regularized update}}(\theta,\phi_1,\phi_2,\alpha)\\
&\;{\scriptstyle[\text{update controller}]}
\end{aligned}
\end{equation}

The \ac{SAC} objective augments the expected return with an entropy maximization term,

\begin{equation}
\label{eq:SAC-obj}
J(\pi) =
\mathbb{E}_{(s,a)\sim\mathcal{B}}
\left[
Q(s,a) - \alpha \log \pi(a|s)
\right],
\end{equation}

which introduces additional hardware requirements such as noise generation, log-probability computation, and temperature adaptation.

On \acp{FPGA}, Equations~\eqref{eq:AC-seq-hw}–\eqref{eq:SAC-seq-hw} translate into dual- or multi-pipeline architectures capable of executing concurrent inference, \ac{TD} or advantage computation, and gradient updates.
The TD-error unit in Equation~\eqref{eq:TDAC} is realized as a compact arithmetic module feeding both actor and critic update paths, while support for \ac{A3C} and \ac{SAC} requires parallel gradient-generation pipelines, global update controllers, and memory-backed replay buffers.

One of the earliest and most influential designs is FA3C~\cite{cho2019fa3c}, which implements a full \ac{A3C} accelerator supporting both inference and on-line training. The architecture introduces a compute-unit-pair structure in which forward and backward engines operate in a tightly pipelined manner, while tensor decomposition, quantization, and a two-level parameter cache mitigate off-chip bandwidth limitations. FA3C achieved  improvement in training throughput over \ac{GPU}- based \ac{A3C}, demonstrating that full \ac{AC} training can be mapped efficiently to \ac{FPGA} fabric. Building on this foundation, the distributed \ac{A3C} accelerator proposed by Ge \textit{et al.}~\cite{ge2024fpga} targets multi-agent \ac{A3C} with system-level optimization across heterogeneous nodes. Their architecture introduces a resource-optimal deployment strategy that avoids redundant replication of actor and critic networks across agents, enabling significantly higher concurrency on mid-sized \ac{FPGA} platforms. The design separates learning phases into dedicated loss-gradient computation units, back-propagation engines, gradient-compression modules, and synchronized parameter-update blocks. Experimental results show  speedup over \ac{GPU}- based \ac{A3C} while maintaining high resource efficiency, making it one of the most advanced recent contributions in distributed \ac{AC} hardware acceleration. A complementary approach is presented in the \ac{A3C}-HC accelerator~\cite{li2022a3c}, which implements \ac{A3C} on a heterogeneous \ac{CPU}- -\ac{FPGA} platform. In this design, environment simulation, high-level control, and synchronization remain on CPU cores, while the \ac{FPGA} executes the forward inference,  \ac{TD} computation, and gradient updates for both actor and critic networks. This partitioning aligns with modern edge-AI deployment strategies in which the \ac{FPGA} acts as a low-latency learning co-processor. The accelerator incorporates quantized feed-forward computation blocks, BRAM-based gradient accumulation buffers, and a timing-aligned dual-network update controller. Results indicate substantial reductions in training time and improved power efficiency compared with \ac{CPU}- only implementations.

Collectively, these efforts show that \ac{AC} algorithms are now viable for real-time, on-chip training, with architectures ranging from high-performance compute-unit-paired accelerators to scalable distributed \ac{A3C} systems and heterogeneous \ac{CPU}- -\ac{FPGA} co-processors. Remaining challenges include coordinating dual-network synchronization, managing memory bandwidth for gradient updates, and balancing resource usage across actor and critic datapaths. Nonetheless, modern \ac{FPGA} fabrics---with abundant \ac{DSP} blocks, multi-port \ac{BRAM}, and fine-grained parallelism---are increasingly capable of supporting continuous, low-latency \ac{AC} learning in embedded and edge-intelligent platforms.

\subsection{Monte-Carlo}

\Ac{MC} can be viewed as a simplified extension of \ac{AC} methods in which no critic or \ac{TD}-error is used. Instead of bootstrapping from value estimates, MC updates rely solely on the empirical return $G_t$ accumulated over an entire episode. This removes the need for \ac{TD}-error circuitry but requires hardware support for storing complete trajectories and computing return values at episode boundaries.

On \ac{FPGA}, MC accelerators replace the \ac{TD}-error unit with a lightweight \emph{return-accumulation block} implemented using fixed-point accumulators and either FIFO-based reward stacks or BRAM-backed trajectory buffers. Because gradients are computed only after episode termination, hardware must support sequential reward writes during sampling and a backward scan or prefix-sum operation to compute $G_t$, both of which map efficiently to pipelined adders or parallel-prefix engines. Policy-gradient \ac{MC} methods also require a \emph{trajectory buffer} to store state–action pairs and, optionally, log-probabilities, which are replayed through the policy network for a single episode-level update. Unlike \ac{A3C} or \ac{SAC}, \ac{MC} pipelines do not include dual networks, entropy-temperature terms, or asynchronous \ac{TD} updates, but they incur higher \ac{BRAM}/\ac{DDR} demand due to full-trajectory retention. As a result, many MC designs offload the trajectory buffer to external memory while keeping the policy network and gradient logic on \ac{FPGA} for efficient fixed-point computation. Compared with \ac{TD}-based accelerators, MC implementations trade frequent, low-latency updates for simpler control logic and more regular memory access patterns. The episodic update regime reduces responsiveness but remains attractive for tasks with short or deterministic episodes where unbiased return estimates are preferable to bootstrapped targets.

The work of Meng et al. \cite{meng2023framework} proposed one of the first \ac{FPGA}-capable \ac{MC} tree search (MCTS) accelerators that supports fully dynamic tree growth without reconfiguration, addressing a key limitation of earlier static-tree \ac{FPGA} designs. The authors introduced a butterfly-based interconnection network and a bank-assignment algorithm to minimize runtime access conflicts and enable arbitrary tree shapes. A hybrid \ac{CPU}–\ac{FPGA} execution model exploits \ac{FPGA} pipeline parallelism for in-tree operations while delegating simulations to \ac{CPU} threads. Their framework achieves up to 6.8× throughput improvement over CPU baselines while maintaining algorithmic performance across Atari benchmarks.

\subsection{PPO}

Compared with classical \ac{AC} algorithms, which updated the actor using instantaneous \ac{TD}-errors, \ac{PPO} introduced a clipped surrogate objective that restricted the magnitude of policy updates. This clipping term
\[
L^{\text{CLIP}}(\theta)
= \mathbb{E}\bigl[
\min\bigl(
r_t(\theta) A_t,\,
\text{clip}(r_t(\theta),1-\epsilon,1+\epsilon)A_t
\bigr)
\bigr],
\]
modified the actor update rule without altering the critic structure, and therefore differed fundamentally from both \ac{AC} and \ac{MC}: the algorithm required neither bootstrapped \ac{TD} errors for the actor nor full-episode returns, but instead relied on advantage estimates accumulated over truncated rollout segments.

On hardware platforms, these algorithmic changes had direct architectural consequences. Unlike \ac{AC}, where the actor received immediate \ac{TD}-errors, \ac{PPO} required a \emph{ratio block} to compute 
$r_t(\theta)=\pi_\theta(a_t\mid s_t)/\pi_{\theta_{\text{old}}}(a_t\mid s_t)$,
necessitating storage of the previous policy’s log-probabilities and additional arithmetic units for division or log-subtraction. 
The clipping path introduced a second comparison-and-select stage that was absent in \ac{AC} and \ac{MC}, and required dedicated fixed-point saturators or \ac{LUT}-based comparators to implement the bounded update. 
The algorithm’s minibatch-based optimization further demanded repeated passes over the same rollout data---in contrast to the single-sweep online updates used in \ac{AC} and the episodic updates in \ac{MC}. 
As a result, \ac{FPGA} designs incorporated \ac{BRAM}-backed rollout buffers and address generators for minibatch iteration, which increased memory pressure compared with on-line \ac{TD} updates.

Unlike \ac{MC}, which computed policy gradients only at episode boundaries, \ac{PPO} operated on truncated horizons, enabling moderate responsiveness while avoiding the high-variance returns that \ac{MC} required. 
This design required hardware support for advantage computation---typically a backward scan accumulating temporal differences---and for storing the value predictions required for generalized advantage estimation (GAE).  These blocks were mapped to pipelined subtract--multiply--accumulate engines or to parallel-prefix units, depending on throughput targets. In contrast to asynchronous \ac{A3C} designs, which relied on many loosely coupled learners, \ac{PPO} favored synchronized optimization over multiple minibatches, which, in hardware, translated into a more regular update schedule but a tighter coupling between the actor, critic, and rollout memory. 
Recent \ac{FPGA} and \ac{IC} studies showed that this structured optimization enabled efficient time-multiplexing of matrix--vector units between actor and critic networks, since the clip-and-ratio computation, rather than the neural layers themselves, became the dominant performance bottleneck.

All in all, \ac{PPO} differed from both \ac{AC} and \ac{MC} by introducing ratio evaluation, policy-clipping logic, and minibatch-based training, all of which required dedicated arithmetic blocks, larger rollout storage, and multi-pass memory access. 
Although these additions increased control complexity, \ac{PPO} offered more stable learning and predictable convergence behavior, making it suitable for \ac{FPGA}/\ac{IC} accelerators targeting real-time but non-stationary environments.

The HEPPO \cite{taha2024heppo} accelerator specifically targeted \ac{PPO}’s Generalized Advantage Estimation (GAE), a component absent in both \ac{AC} and \ac{MC} implementations. The design introduced a deeply pipelined GAE computation unit with reward standardization, block standardization, and uniform 8-bit quantization. This standardization pipeline stabilized \ac{PPO}’s advantage estimates and reduced on-chip memory footprint by $4\times$, addressing a bottleneck unique to \ac{PPO}. Unlike \ac{AC} designs that only evaluated $V(s)$ inline, HEPPO implemented backward GAE recursion using dedicated arithmetic chains, minimizing memory accesses. Its single-\ac{SoC} integration eliminated \ac{CPU}--\ac{GPU} communication and delivered up to $30\%$ speed improvement, demonstrating that \ac{PPO} could be efficiently executed without large off-chip buffers. PPOAccel \cite{meng2021ppoaccel} addressed the heavy training workload of \ac{PPO} by accelerating \emph{both} inference and backpropagation on systolic arrays. The work exploited the structural similarity between \ac{MLP} and \ac{CNN} policies and implemented a unified GEMM engine with shared forward/backward datapaths---a capability unnecessary in \ac{AC} or \ac{MC} designs that used lighter function approximators. PPOAccel also introduced a novel memory-blocked data layout enabling stall-free streaming during multi-epoch updates, directly addressing \ac{PPO}’s strict requirement to reuse the same batch across $K$ training epochs. The accelerator achieved up to $27\times$ throughput improvement over \ac{CPU}/\ac{GPU} baselines, mainly by eliminating kernel-launch overheads and training/inference imbalance typical in \ac{PPO}. Building on these \ac{PPO}-specific architectural requirements target the ratio, clipping, and GAE computation pipelines. For example, PPOAccel framework \cite{meng2021ppoaccel} proposed a CPU–\ac{FPGA} heterogeneous accelerator for \ac{PPO} that jointly speeds up policy inference and training through a unified systolic-array engine applicable to both \ac{MLP}- and \ac{CNN}-based policies. The design carefully profiled \ac{PPO}’s sample-generation and model-update phases, then introduced optimized dataflow mapping, memory-blocked data layouts, and systolic compute-sharing to reduce data-movement overhead and balance utilization. Evaluations on standard robotics and Atari benchmarks show that PPOAccel effectively exploits \ac{PPO}’s structure while preserving its learning dynamics.

\subsection{Comprehensive Performance Metrics and Cross-Design Analysis}
\label{subsec:comprehensive_metrics}

Although the surveyed \ac{FPGA}/\ac{IC} implementations report several raw hardware indicators in Table~\ref{table:4}, and application-level performance improvements in Table~\ref{table:7}, these metrics alone do not provide a complete view of implementation efficiency. This is particularly important for \ac{RL} hardware, where different works accelerate different execution phases, including policy inference, value-function update, replay-buffer sampling, backpropagation, environment interaction, and full training iterations. Therefore, a comprehensive comparison requires normalized metrics that jointly capture performance, energy consumption, and hardware cost.

In this review, the comprehensive evaluation is organized around three complementary metric families: energy efficiency, resource efficiency, and latency--energy trade-off. Energy-efficiency metrics evaluate how much useful \ac{RL} computation is achieved per unit of power. Depending on the reporting style of each study, the useful computation may correspond to inferences per second, training updates per second, samples per second, throughput, or \ac{FIPS}. Among these, \ac{FIPS}/W is especially informative for full \ac{RL} training accelerators because it accounts for the complete learning loop rather than only isolated inference or update kernels.

Resource-efficiency metrics complement energy-efficiency analysis by relating the achieved performance to the amount of \ac{FPGA} or \ac{IC} hardware consumed. This is necessary because two designs may report similar throughput or speedup while using very different fractions of \ac{BRAM}, \acp{LUT}, \acp{FF}, \ac{LUTRAM}, or \ac{DSP} resources. A design with moderate throughput but very low utilization may be preferable for embedded or multi-agent deployment, whereas a design with high throughput but near-saturated \ac{BRAM} or \ac{DSP} usage may face limited scalability. For this reason, both the average resource footprint and the dominant bottleneck resource should be considered when comparing hardware efficiency.

The latency--energy trade-off is another important dimension, particularly for real-time \ac{RL} applications such as robotics, networking, dynamic voltage and frequency scaling, autonomous control, and hardware security. Low latency alone does not necessarily imply an efficient design if it is achieved at the cost of excessive power consumption. Conversely, a low-power design may not be suitable for strict real-time control if the decision latency is too high. Energy per operation and energy--delay behavior therefore provide a more balanced view of whether a hardware accelerator is suitable for latency-critical and energy-constrained deployment.

Based on the raw implementation characteristics summarized in Table~\ref{table:4}, the normalized evaluation criteria defined in Table~\ref{tab:5}, and the application-level performance comparisons reported in Table~\ref{table:7}, the reviewed works show substantial variation in workload granularity, hardware cost, energy efficiency, resource efficiency, and reported performance objectives. Lightweight tabular \ac{QL} accelerators generally exhibit very small resource footprints and low power consumption, making them attractive for embedded control tasks with compact state--action spaces. For example, the \ac{QL} implementations in~\cite{liu2022fpga,spano2019efficient} consume only a small fraction of \ac{LUT}, \ac{FF}, \ac{LUTRAM}, and \ac{DSP} resources and operate at tens of milliwatts. These characteristics indicate strong suitability for low-power embedded RL, although their efficiency should be interpreted within the context of compact tabular workloads rather than compared directly with neural-network-based deep \ac{RL} accelerators.

\begin{table*}[t]
\centering

\caption{Comprehensive normalized metrics for cross-design comparison of \ac{RL}-based \ac{FPGA}/\ac{IC} implementations.}
\label{tab:5}
\renewcommand{\arraystretch}{1.25}
\begin{tabular}{p{3cm} p{3cm} p{9cm} p{1.2cm}}
\hline
\textbf{Metric} & \textbf{Expression} & \textbf{Interpretation} & \textbf{Preferred trend} \\
\hline

Throughput-based energy efficiency &
$\mathrm{EE_T = T/P}$ &
Measures the amount of useful \ac{RL} computation achieved per watt. The throughput $\mathrm{T}$ may represent inferences/s, updates/s, samples/s, or full iterations/s, depending on the reported workload. &
Higher \\

FIPS/W &
$\mathrm{\mathrm{FIPS}/P}$ &
Measures full \ac{RL} iterations processed per watt. This metric is particularly suitable for accelerators that execute both decision-making and learning-update stages. &
Higher \\

Energy per operation &
$\mathrm{E_{\mathrm{op}} = P/T}$ or \newline $\mathrm{E_{\mathrm{op}} = P \times L}$ &
Estimates the energy consumed by one inference, update, sample, or full \ac{RL} iteration. It is useful when comparing low-power and real-time accelerators. &
Lower \\

Average resource footprint &
$\mathrm{RU_{\mathrm{avg}} = \frac{1}{|\mathcal{R}|}\sum_{r \in \mathcal{R}} U_r}$ &
Represents the mean utilization of the reported \ac{FPGA} resources, where $\mathcal{R}$ may include \ac{BRAM}, \ac{LUT}, \ac{LUTRAM}, \ac{FF}, and \ac{DSP} usage. &
Lower \\

Bottleneck resource utilization &
$\mathrm{RU_{\max} = \max_{r \in \mathcal{R}} U_r}$ &
Identifies the most heavily used hardware resource. This metric helps reveal scalability limits even when the average resource usage appears moderate. &
Lower \\

Resource efficiency &
$\mathrm{RE_T = T/RU_{\mathrm{avg}}}$ &
Relates useful \ac{RL} throughput to average resource utilization. It helps compare designs that achieve similar performance with different hardware costs. &
Higher \\

Energy--resource efficiency &
$\mathrm{ERE_T = T/(P \times RU_{\mathrm{avg}})}$ &
Combines throughput, power, and resource utilization into a single efficiency indicator. It is useful for identifying designs that are both energy-efficient and hardware-efficient. &
Higher \\

Energy--delay product &
$\mathrm{EDP = E_{\mathrm{op}} \times L}$ or \newline $\mathrm{EDP = P \times L^2}$ &
Captures the latency--energy trade-off. It is particularly relevant for real-time \ac{RL} systems where low latency and low energy must be achieved simultaneously. &
Lower \\

\hline
\end{tabular}
\begin{flushleft}
\scriptsize{\textit{\textbf{Note:}} The normalized metrics in this Table are intended to complement the raw hardware indicators summarized in Tables~\ref{table:4} and~\ref{table:7}, including resource utilization, clock frequency, power consumption, implementation level, and reported best results. Because the reviewed studies use different workload granularities and baseline platforms, derived metrics should be interpreted within the same execution category, such as inference-only, training-only, replay-buffer acceleration, or full \ac{RL} iteration. Speedup-based comparisons are treated qualitatively because \ac{CPU}/\ac{GPU} baselines differ across studies.}
\end{flushleft}
\end{table*}

In contrast, \ac{DQN}, \ac{PPO}, \ac{DDPG}, and \ac{AC} accelerators provide higher learning capability and support more complex environments, but they typically require substantially more \ac{BRAM}, \ac{DSP}, and \ac{LUT}  resources due to neural-network inference, replay-buffer access, target-network storage, and gradient computation. For example, \ac{DQN} accelerators such as~\cite{zhang2023deep,sumarudin2025low} report large speedups over \ac{CPU} or \ac{GPU} baselines, but their resource profiles differ significantly. High-throughput \ac{DQN}   implementations often consume larger fractions of \ac{LUT}  and \ac{DSP} resources, whereas compact \ac{DQN}   accelerators trade model complexity and generality for lower power and reduced resource usage. This demonstrates that speedup alone is insufficient for cross-design comparison unless it is interpreted together with power, latency, and resource utilization.

\ac{AC} and \ac{A3C}-based accelerators further illustrate the need for comprehensive metrics. Designs such as~\cite{cho2019fa3c,ge2024fpga,li2022a3c} report considerable acceleration over \ac{GPU} or \ac{CPU} baselines, but they also introduce high resource pressure because actor and critic networks, gradient engines, local workers, and global update logic must be supported concurrently. These designs are therefore better evaluated using combined performance--resource and performance--power indicators rather than raw speedup alone. In particular, replicated worker pipelines may improve throughput but can also increase \ac{BRAM}, \ac{DSP}, and routing pressure, which affects scalability on smaller \ac{FPGA} devices.

Among the reviewed works, TD3lite~\cite{hu2022td3lite} provides one of the most explicit examples of comprehensive performance reporting. It reports full-iteration throughput together with measured power and resource utilization, allowing direct evaluation of \ac{FIPS}/W. This makes it a strong example of balanced energy and resource efficiency for full \ac{RL} training. The design also demonstrates how block floating-point quantization and structural simplification can reduce the resource footprint of \ac{AC} pipelines while maintaining high training throughput. Compared with works that emphasize only speedup or throughput, TD3lite provides a clearer basis for evaluating whether the accelerator is efficient from both energy and hardware-cost perspectives.

The latency--energy trade-off should also be interpreted according to the target application. Many \ac{RL}-based \ac{FPGA}/\ac{IC} systems are designed for real-time or energy-constrained environments, where the suitability of a design depends not only on its average throughput or reported speedup, but also on whether each decision, update, or training iteration can be completed within the required timing and power budget. Therefore, latency-related metrics should be interpreted together with reported power consumption and the type of accelerated \ac{RL} workload. This is especially relevant for applications such as robotics, networking, autonomous control, hardware security, and adaptive power management, where both fast response and low energy consumption are required.

A further challenge is that the surveyed studies use different performance units and baseline platforms. Some works report speedup over \ac{CPU}, others report speedup over \ac{GPU}, while others report throughput, samples per second, inference time, power, or reward improvement. Since \ac{CPU} and \ac{GPU} baselines differ across studies, speedup-based comparisons should be interpreted qualitatively rather than as absolute efficiency rankings. Similarly, throughput-based comparisons are meaningful only when the accelerated workload is clearly specified, such as inference-only, training-only, replay-buffer acceleration, or full \ac{RL} iteration.

In global, the comparative analysis indicates that existing \ac{RL}-based \ac{FPGA}/\ac{IC} studies demonstrate clear improvements in speed, throughput, power efficiency, and resource utilization. However, because the literature reports heterogeneous workload granularities and baseline platforms, normalized metrics should be used to make comparisons more consistent and interpretable. Studies should report decision latency, training-update latency, throughput, measured board or chip power, energy per operation, resource utilization, and \ac{FIPS}/W when full \ac{RL} iterations are accelerated. They should also indicate whether the reported power corresponds to device-level, board-level, or full-system consumption. Such reporting would enable more accurate comparison of energy efficiency, resource efficiency, and latency--energy behavior across future \ac{RL}-based \ac{FPGA} and \ac{IC} implementations.

\color{black}

\section{Research Areas in RL-based FPGA and IC    Platforms}
\label{sec4}
\ac{RL} has stimulated a diverse range of innovations within \ac{FPGA} {and \ac{IC}} platforms, extending beyond acceleration to influence configurability, optimization, and system-level intelligence. As \ac{RL} algorithms evolve—from classical tabular methods to deep, \ac{AC}, and continuous-control models—their hardware realizations on \acp{FPGA} {and \acp{IC}} increasingly leverage parallelism, quantization, and reconfigurable {or specialized} logic to achieve real-time, energy-efficient decision-making. The existing body of work {can be broadly divided according to two principal roles of \ac{RL}: (i) \ac{RL} as the workload, where hardware platforms are designed to accelerate \ac{RL} execution, and (ii) \ac{RL} as the optimizer, where learning agents are used to improve hardware configuration, runtime adaptation, and design automation}. {Within these two categories,} the reviewed studies naturally cluster into several major research directions, which are discussed comprehensively in the subsections that follow. This organization provides a clear structure for understanding how \ac{RL} is used to accelerate computation, guide hardware--software co-design, enable adaptive reconfiguration, improve energy efficiency, and support large-scale coordinated learning across \ac{FPGA} {and \ac{IC}} platforms. To guide the reader, Table~\ref{tab:6} summarizes the principal research areas surveyed in this section.

\renewcommand{\arraystretch}{1.25}
\begin{table*}[ht!]
\centering
\scriptsize
\caption{{Overview of the two principal roles of \ac{RL} in \ac{FPGA}/\ac{IC} research and their associated sub-domains, highlighting the primary focus, typical objectives, and key challenges of each area.}}
\label{tab:6}
\begin{tabular}{m{0.5cm}m{2cm} m{4.5cm} m{4.5cm} m{4.5cm}}
\hline
&\textbf{Research Area} & \textbf{Primary Focus} & \textbf{Typical Objectives} & \textbf{Key Challenges} \\
\hline

\multirow{1}{*}{\rotatebox{90}{{\textbf{\ac{RL} as the  Workload}}}} &\textbf{RL-based \newline acceleration} &
Mapping \ac{RL} computation (value updates, NN inference, gradients) onto \ac{FPGA} pipelines &
Reduce latency; increase throughput; exploit parallelism; support on-device training &
Resource constraints; precision tuning; balancing compute pipelines \\ 

&\textbf{Multi-Agent RL} &
\ac{FPGA}-based architectures supporting distributed or local \ac{MARL} agents &
Scalable learning; cooperative control; reduced centralized bottlenecks &
Non-stationarity; coordination overhead; gradient synchronization \\

&\textbf{Multi-FPGA \ac{RL} \newline platforms} &
Distributed \ac{RL} execution across multiple interconnected \ac{FPGA}s &
Large-scale RL; higher throughput; extended memory capacity &
Communication latency; consistency of updates; partitioning strategies \\

& {\textbf{Energy-Efficient \ac{RL} Acceleration}} & 
{Energy-efficient \ac{RL} execution on \acp{FPGA} using neuromorphic, mixed-precision, and hierarchical designs} &
{Reduce energy consumption of \ac{RL} learning/inference; improve GOP/s/W; enable low-power on-device adaptation} &
{Task specificity; limited online learning flexibility; memory and precision trade-offs} \\ \hline

&\textbf{HW--SW \newline co-exploration} &
Joint exploration of algorithmic and hardware design parameters using \ac{RL}-based search &
Automated design-space exploration; precision selection; mapping and scheduling optimization &
Large search spaces; long exploration time; generalization to unseen workloads \\

\multirow{1}{*}{\rotatebox{90}{{\textbf{\makecell{\ac{RL} as the  Optimizer}}}}} &\textbf{Reconfig. \ac{RL} \newline approaches} &
Runtime reconfiguration controlled by \ac{RL} agents (DPR, adaptive tuning) &
Adaptive performance; robustness under workload/environment variations; real-time control &
Reconfiguration cost; policy stability; ensuring safe reconfiguration \\

&\textbf{RL-based Optimization} &
System-level optimization of scheduling, routing, task allocation, and hardware parameters &
Improve efficiency; balance workloads; optimize global system performance &
Scalability; convergence stability; multi-objective formulation \\

&\textbf{High-Level \& \newline Logic synthesis} &
Integrating \ac{RL} into HLS, logic optimization, placement, and routing flows &
Improve QoR; reduce compile time; guide EDA decision-making &
High dimensionality of EDA flows; reproducibility; integration with proprietary tools \\

& {\textbf{RL-based Energy Management}} & {\ac{RL}-based platform-level energy control, including power management, \ac{DVFS}, and core activation} &
{Reduce platform energy consumption; adapt to workload variations; improve runtime energy efficiency} &
{Reward-design sensitivity; restricted state representations; stability and transferability across workloads and platforms} \\

&{\textbf{RL-guided Multi-\ac{FPGA}/\ac{IC}  \newline infrastructures}} &
{Using \ac{RL} to manage communication, routing, or control in multi-device \ac{FPGA}/\ac{IC}  platforms} &
{Reduce congestion; improve interconnect efficiency; shorten backend/runtime overhead} &
{Large state/action spaces; reward-design complexity; limited policy transferability} \\

\hline
\end{tabular}
\end{table*}

\subsection{RL as the Workload}
\color{black}

\subsubsection{RL-based Acceleration}

\ac{RL}-based acceleration focuses on mapping the computational primitives of \ac{RL}—including value updates, neural-network inference, gradient computation, and memory operations—onto \ac{FPGA} fabrics to exploit fine-grained parallelism and low-latency dataflow. Across the literature, three dominant architectural strategies emerge: (i) replay-centric acceleration, (ii) massively parallel environment execution, and (iii) application-specific compute pipelines. These approaches collectively aim to improve throughput, reduce latency, and enhance resource efficiency through pipelining, quantization, and heterogeneous \ac{SoC} integration.

Early works primarily targeted heterogeneous \ac{CPU}–\ac{GPU}–\ac{FPGA} frameworks, where the main objective was to offload critical bottlenecks such as experience replay and learning updates. In this context, replay-centric designs introduced parallel data structures and pipelined update mechanisms to accelerate memory-intensive operations. or example, Meng et al. \cite{meng2022fpga} proposed a replay accelerator based on a parallel $K$-ary sum-tree, achieving $21$–$40\times$ speedup in replay operations, illustrating the significant impact of memory-centric optimization.  Building on this idea, later work \cite{meng2025acceleration} extended the approach to system-level acceleration by modeling \ac{RL}  components—Actors, Learners, and Replay Managers—as a task graph for hardware/software co-partitioning. In parallel, environment-level parallelism has been explored as an orthogonal strategy, where Li et al. \cite{li2025pearl} demonstrated that scaling to thousands of parallel environment instances can significantly improve data-generation throughput and training efficiency.

A complementary direction focuses on integrating deep \ac{RL} models directly into \ac{FPGA} pipelines for embedded inference and training. These works highlight the importance of fixed-point arithmetic, deep pipelining, and heterogeneous execution. For instance, compact \ac{DQN} accelerators have demonstrated real-time inference with low power consumption \cite{thinh2022hardware}, while heterogeneous \ac{CPU}–\ac{FPGA} frameworks \cite{li2019implementation} show that partial offloading of neural inference already yields deterministic timing advantages. Additionally, lightweight \ac{RL} models such as bandit-based approaches have been mapped to hardware pipelines \cite{radovic2021hardware}, illustrating the flexibility of \ac{FPGA} acceleration across varying algorithmic complexity.

As the field matured, research increasingly shifted toward application-specific \ac{RL} accelerators, where hardware architectures are co-designed with target domains and system constraints. These designs emphasize tightly coupled compute pipelines, efficient on-chip memory usage, and real-time interaction with physical systems. Representative examples include \ac{RL}- based resource-allocation engines \cite{gerogiannis2022deep}, real-time control systems operating at microsecond scale \cite{wang2021accelerated}, and policy-gradient accelerators for robotics applications \cite{shao2018towards}. More recent work has extended these ideas to multimodal \ac{RL} systems integrating vision, language, and control pathways \cite{shiri2021hardware}, further demonstrating the versatility of hardware-aware \ac{RL} design.

{Collectively, a clear evolution can be observed from isolated component-level optimizations toward fully integrated, domain-specific \ac{RL} systems. Despite differences in application and algorithmic choice, common architectural patterns consistently emerge, including on-chip replay buffers, pipelined learning engines, high-parallelism environment execution, fixed-point computation, and dataflow-oriented neural-network implementations. These patterns form the foundation of modern hardware–\ac{RL} co-design and highlight the convergence of architectural strategies across the literature.}

\subsubsection{Multi-agent RL–based Approaches}

\Ac{MARL} extends \ac{RL} to distributed \ac{FPGA} architectures in which multiple learning agents operate concurrently across a single or multi-tile fabric. Such systems exploit the fine-grained parallelism and heterogeneous resources of \ac{FPGA} devices to support cooperative, competitive, or decentralized policy learning at scale. From a hardware perspective, existing \ac{MARL} accelerators can be broadly categorized according to three dominant bottlenecks: (i) control latency, (ii) inter-agent communication, and (iii) deep-network computation.

Early works primarily addressed the control-latency bottleneck, focusing on fast closed-loop response rather than deep learning throughput. For instance, swarm-based multi-agent extensions of \ac{QL} have been implemented on \ac{FPGA} to regulate communication systems, achieving significant reductions in convergence time through low-latency updates and deterministic execution \cite{cardarilli2022fpga}. Similarly, tabular \ac{MARL} architectures leveraging centralized processing units and shared Q-tables demonstrated efficient resource utilization and sub-microjoule energy consumption, highlighting the suitability of lightweight multi-agent learning for real-time hardware deployment \cite{cardarilli2020fpga}.

As \ac{MARL} algorithms evolved toward centralized-training and decentralized-execution paradigms, inter-agent communication emerged as a dominant bottleneck. To address this challenge, recent designs introduced parallel \ac{AC} pipelines combined with efficient on-chip interconnects to enable scalable coordination among agents. For example, hardware-accelerated multi-agent \ac{DDPG} (MADDPG) frameworks \cite{wiggins2023accelerating} demonstrated significant improvements in training throughput by exploiting parallelism and lightweight communication mechanisms. This issue becomes even more pronounced in attention-based architectures, where centralized critics require dense information exchange, increasing memory pressure and data movement costs. Specialized accelerators incorporating attention modules and pipelined communication strategies have therefore been proposed to mitigate these overheads \cite{wiggins2024heterogeneous}.

A third class of \ac{MARL} accelerators targets the compute-bound regime, where deep neural-network training dominates execution time. In this context, sparse computation and model compression techniques have been employed to reduce computational complexity and improve efficiency. For instance, LearningGroup \cite{yang2022learninggroup} integrates learnable weight grouping with on-chip sparse-data generation, enabling real-time compressed training and achieving substantial performance and energy-efficiency gains over dense implementations.

{In light of these proposed solutions, a clear trend emerges in which \ac{MARL} hardware design is driven by the dominant system bottleneck rather than the specific learning algorithm. This bottleneck-oriented perspective provides a unified view of the design space and clarifies how different architectural strategies—ranging from low-latency control pipelines to communication-efficient interconnects and compute-optimized sparse engines—address the scalability challenges of multi-agent systems. Nevertheless, current \ac{MARL} accelerators still face important limitations. As the number of agents increases, communication overhead and memory requirements grow non-linearly, particularly in architectures with centralized critics or attention mechanisms. Furthermore, many existing designs rely on highly specialized pipelines tailored to specific algorithms, limiting portability across different \ac{MARL} formulations. These challenges highlight the need for more flexible and scalable hardware abstractions to support next-generation multi-agent \ac{RL} systems.}

\subsubsection{Multi-FPGA/IC Platforms for Scalable RL Execution}

When \ac{RL} is treated as the \emph{workload}, multi-\ac{FPGA} and multi-\ac{IC} platforms are primarily designed to scale learning and inference beyond the capacity of a single device. In this setting, the main objective is to distribute \ac{RL} computation across multiple interconnected chips in order to increase throughput, expand memory capacity, and support large numbers of parallel agents. Such systems therefore rely on efficient inter-device communication, synchronized updates, and effective workload-partitioning strategies to maintain scalability.

A representative example is the IBM Neural Computer used by Asseman~\textit{et al.}~\cite{asseman2021accelerating}. Unlike single-device accelerators, this system distributes the \ac{RL} workload across 416 interconnected Zynq-7045 nodes organized in a high-bandwidth, low-latency 3D mesh. Each node implements the full fitness-evaluation pipeline in \ac{RTL}, including the Atari~2600 environment, image preprocessing, and a quantized \ac{CNN}, allowing 832 parallel agents to be evaluated simultaneously. This architecture exemplifies how distributed execution enables scalable, fine-grained parallelism, particularly for population-based \ac{RL} methods such as deep neuroevolution, where large numbers of candidate policies must be evaluated independently. By avoiding centralized bottlenecks and leveraging mesh-based communication for synchronization, such systems achieve substantial improvements in evaluation throughput and training time.

{Across these developments, multi-device \ac{RL} acceleration highlights a fundamental trade-off between scalability and coordination overhead. While distributing computation enables significant throughput gains, it also introduces challenges in maintaining consistency, managing communication latency, and balancing workloads across devices. Despite these advances, multi-\ac{FPGA} platforms for scalable \ac{RL} execution still face important limitations, including complex cross-device synchronization, increasing communication overhead as the number of devices grows, and reliance on hand-tuned workload partitioning and fixed interconnect structures. These issues continue to limit portability and scalability, highlighting the need for more automated distribution strategies and flexible communication abstractions.}

\subsubsection{Energy-Efficient Acceleration of RL}

A first line of research addresses \ac{RL} as the \emph{computational workload} itself and focuses on reducing the energy cost of executing learning and inference on \acp{FPGA}. In this context, the main objective is not to optimize the hardware platform through \ac{RL}, but rather to design hardware architectures that implement \ac{RL} efficiently under strict power and resource constraints.

In this regard, a major research direction focuses on neuromorphic and spiking \ac{RL} architectures, which reduce energy consumption through sparse, event-driven computation and local learning rules. These designs typically replace dense neural updates with spike-based communication and exploit temporal sparsity to minimize switching activity and memory traffic. For example, spiking implementations based on leaky integrate-and-fire neurons and \ac{STDP} have demonstrated significant gains in both convergence speed and energy efficiency \cite{asgari2020low}. Complementary designs explore low-bitwidth neuron–synapse representations and lightweight hardware platforms to sustain low-power operation even for non-trivial workloads \cite{bahar2025low}. Further optimizations include multiplier-less arithmetic, coarse quantization, and temporal hysteresis, enabling substantial reductions in energy consumption while maintaining learning capability \cite{rasheed2024reduced}. In addition, biologically inspired reinforcement mechanisms based on reward-modulated \ac{STDP} eliminate the need for replay buffers and centralized value-function computation, thereby reducing memory overhead and improving efficiency \cite{quintana2022bio}.

Beyond neuromorphic designs, a second direction focuses on architectural and algorithmic optimizations to reduce the energy cost of \ac{RL} training and inference. Jo et al.~\cite{jo2021energy} proposed an energy-efficient \ac{RL} accelerator that incorporates a heterogeneous replay buffer to reuse past experiences, selective layer retraining to limit backpropagation costs, and a mixed-precision compute architecture that safely replaces a large portion of FP32 operations with FXP16 arithmetic. This combination led to up to 90\% reduction in adaptation iterations and 12\% reduction in computational workload, with more than 27\% of \ac{RL} operations executed in low precision without degrading the learned policy. In addition to flat policy architectures, several researchers explored structured \ac{RL} formulations that inherently reduce exploration cost and thus energy consumption. Shiri et al.~\cite{shiri2022e2hrl} introduced a hierarchical \ac{RL} accelerator (E2HRL) in which complex behaviors are decomposed into multi-level subgoals, enabling agents to learn with far fewer state--action transitions. Their \ac{FPGA} implementation on Artix-7 demonstrated 1.1--11.4~GOP/s/W energy efficiency, corresponding to 8.8$\times$--11$\times$ improvements compared to \ac{GPU}- based \ac{RL} execution. Related work by the same authors extended the concept to multimodal \ac{RL} by incorporating structured language instructions into the decision-making process~\cite{shiri2020energy}. The resulting configurable hardware architecture achieved a low-power operating mode below 139~mW while offering a high-throughput option reaching 217~FPS at approximately 1.2~mJ per decision.

{Overall, these studies show that event-driven computation, lightweight arithmetic, hierarchical decomposition, and precision-aware design form the main architectural foundations for energy-efficient acceleration of \ac{RL} on reconfigurable hardware. Nevertheless, current solutions remain constrained by task specificity, limited online learning flexibility, and memory/precision trade-offs that may restrict their generality for more complex \ac{RL} workloads.}

\subsection{RL as the Optimizer}
\color{black}
\subsubsection{RL-Based Reconfigurable Hardware Approaches}

Reconfigurable hardware has increasingly been coupled with \ac{RL} to enable adaptive, resource-aware, and autonomously self-optimizing embedded systems. The literature in this domain broadly converges toward three main research directions: (i) \ac{RL}-guided dynamic architectural reconfiguration on \acp{FPGA}, (ii) \ac{RL}-enhanced reconfigurable cyber--physical components, and (iii) \ac{RL}-driven distributed embedded devices. Together, these directions illustrate how learning-based control can bridge hardware adaptability with real-time decision making in highly dynamic environments.

A significant line of work focuses on \ac{RL}-driven \textit{dynamic reconfiguration} of \ac{FPGA} accelerators, where learning algorithms adapt hardware structure and operating parameters at runtime. These approaches typically combine partial reconfiguration, precision tuning, and workload-aware scheduling to match accelerator configurations to changing computational demands. For example, dynamically reconfigurable neural architectures have been explored in which evolutionary or \ac{RL}- like feedback guides structural adaptation of FPGA-based networks \cite{garcia2020dynamically}. Similarly, \ac{RL} has been applied to optimize sparsity, pruning, and precision in neural accelerators, enabling hardware to autonomously adjust its performance--energy trade-off across workloads \cite{pal2025dynamic}. Other works extend this concept to runtime scheduling, where \ac{RL} determines optimal switching strategies among reconfigurable modules to improve overall system efficiency \cite{gong2020improving}. These studies collectively demonstrate that \ac{RL}- driven reconfiguration enables fine-grained, workload-aware adaptation of hardware resources, significantly improving flexibility compared to static accelerator designs.

Beyond \ac{FPGA} compute fabrics, \ac{RL} techniques are increasingly applied to \textit{reconfigurable cyber--physical} components whose behavior evolves under environmental or operational variations. In these systems, \ac{RL} serves as a real-time control mechanism to maintain performance under non-stationary conditions. For instance, \ac{RL}- based controllers have been used to dynamically recalibrate analog circuits such as power amplifiers, compensating for temperature-induced drift and preserving system reliability \cite{xu2024reinforcement}. Similarly, \ac{RL} has been applied to optimize high-dimensional analog control parameters in reconfigurable radio frequency systems, where learning-based strategies effectively manage complex interference patterns \cite{peng2024active}. In another domain, \ac{RL} has been employed to regulate dynamic battery topologies in electric vehicles, enabling adaptive state-of-charge balancing across heterogeneous cells \cite{baccari2024deep}. These works collectively highlight \ac{RL}'s effectiveness in controlling physical systems characterized by nonlinear dynamics, uncertainty, and continuously varying operating conditions.

Another important research direction examines \ac{RL}-driven \textit{reconfiguration in distributed embedded devices} equipped with small \ac{FPGA} fabrics. These systems typically operate under tight energy and resource constraints, requiring decentralized decision-making and adaptive workload management. For example, \ac{RL}- based strategies have been used in self-aware sensor nodes to decide when to reconfigure hardware, execute tasks locally, or offload computation to neighboring nodes \cite{burger2021developing}. Extensions of this approach incorporate more complex edge scenarios, including job batching, dynamic offloading, and cooperative execution \cite{burger2020reconfigurable}. These approaches demonstrate how \ac{RL} can effectively manage energy--performance trade-offs in distributed systems while enabling emergent cooperative behavior.

{Across these developments, several unifying observations emerge. First, \ac{DPR} consistently appears as a key mechanism enabling \ac{RL} agents to adapt hardware structure with minimal runtime disruption. Second, model-free \ac{RL} approaches are particularly effective in handling nonlinear, uncertain, and environment-dependent hardware behavior. Third, many systems directly incorporate hardware performance measurements—such as latency, interference levels, temperature drift, or voltage imbalance—into the reward function, making physical operating conditions an integral part of the learning process. Finally, scalability challenges are often addressed through architecture-aware strategies such as module clustering, coefficient partitioning, and hierarchical control. Despite these promising capabilities, several limitations remain. Existing frameworks often rely on simplified \ac{RL} models with restricted state representations, limiting robustness and applicability to large and highly dynamic hardware design spaces. In addition, runtime reconfiguration overhead and training instability introduce practical constraints when deploying these methods in real \ac{FPGA} systems. The absence of standardized evaluation benchmarks further hinders fair comparison and limits the generalizability of current solutions across different architectures and deployment scenarios.}

\subsubsection{RL-based Energy and Power Management}

Another line of research treats \ac{RL} not as the workload being accelerated, but as the \emph{optimizer} used to regulate the energy behavior of \ac{FPGA}-based systems. In this setting, the learned policy acts as a runtime controller that dynamically adapts operating modes, resource utilization, or voltage/frequency configurations in response to workload variations and platform conditions. These approaches are particularly attractive for managing complex and non-linear energy--performance trade-offs that are difficult to capture using static heuristics.

Existing works in this area primarily focus on dynamic power management and adaptive resource control. For example, early hardware implementations of \ac{RL} controllers demonstrated the feasibility of learning-based power management, where temporal-difference learning was used to select optimal operating policies based on observed workload patterns \cite{prabha2007hardware}. More recent approaches extend this concept by incorporating system-level awareness and advanced training strategies. In particular, digital-twin–assisted frameworks enable \ac{RL} agents to be trained in high-speed virtual environments, reducing the need for costly on-device exploration while preserving adaptability to real-world conditions \cite{laad2025digital}. Such approaches allow runtime controllers to efficiently optimize \ac{DVFS} policies, core selection, and workload distribution across heterogeneous \ac{FPGA} platforms.

{A common pattern across these approaches is, \ac{RL}- based energy management highlights a shift from static and heuristic-based control toward adaptive, data-driven optimization. By continuously interacting with the system, \ac{RL} agents can capture complex dependencies between workload characteristics and hardware behavior, enabling more responsive and efficient power management strategies. Despite these advantages, several challenges remain. Current approaches are often sensitive to reward design and rely on limited state representations, which can restrict generalization across workloads and platforms. In addition, ensuring stable and predictable runtime behavior remains difficult, particularly in safety-critical or latency-sensitive applications. These limitations highlight the need for more robust training methodologies, transferable policy representations, and tighter integration between \ac{RL} algorithms and hardware monitoring infrastructures.}

\subsubsection{RL in Multi-FPGA/IC Infrastructures}

When \ac{RL} is treated as the \emph{optimizer}, the objective shifts from accelerating learning to improving the behavior of large multi-\ac{FPGA} or multi-\ac{IC} infrastructures. In this setting, the hardware platform itself forms the environment, while the \ac{RL} agent learns policies that enhance communication efficiency, reduce congestion, and shorten backend design cycles. Such approaches are particularly relevant for large-scale systems where traditional heuristics struggle to capture complex interactions within dense interconnect fabrics.

Existing work in this area primarily focuses on communication-aware optimization and design-flow acceleration. For example, \ac{RL}-based routing frameworks have been proposed to dynamically select congestion-aware paths in multi-\ac{FPGA} interconnects, improving both latency and routing efficiency \cite{farooq2024reinforcement}. These approaches leverage online reward feedback to adapt routing decisions to varying traffic patterns and resource constraints. Reported results indicate that such learning-based methods can significantly reduce routing time and alleviate communication bottlenecks, leading to measurable improvements in overall backend design cycles.

{A key observation from these studies is that \ac{RL} enables adaptive, data-driven optimization of system-level behavior, replacing static heuristics with policies that can respond to workload and topology variations. This is particularly important in multi-device environments, where the interaction between communication, placement, and routing becomes increasingly complex as system scale grows. Despite these promising results, \ac{RL}-based optimization for multi-\ac{FPGA}/\ac{IC} platforms remains constrained by several challenges. These include the difficulty of designing effective reward functions, the large state and action spaces induced by complex interconnect topologies, and the limited generalizability of learned policies across different platforms and workloads. Addressing these challenges will require more scalable learning formulations and transferable optimization frameworks for next-generation reconfigurable systems.}

\color{black}

\subsubsection{RL-based System and Design Optimization}

\ac{RL} has emerged as a powerful paradigm for optimizing \ac{FPGA} systems in domains traditionally dominated by heuristics. Rather than accelerating \ac{RL} or focusing solely on energy efficiency, these approaches use \ac{RL} to improve computer-aided design \ac{QoR}, including placement, routing, scheduling, compiler optimization, and system-level runtime control. In this context, the \ac{FPGA} architecture itself becomes part of the environment, enabling agents to learn policies that navigate complex structural constraints and design trade-offs.

Existing work in this area can be broadly categorized into four main directions: (i) placement optimization, (ii) scheduling and runtime control, (iii) compiler and design-space exploration, and (iv) system-level optimization.

\textit{Placement optimization} represents one of the earliest and most extensively studied applications of \ac{RL}. These approaches replace manually tuned heuristics such as simulated annealing with adaptive policies that guide placement decisions. For example, \ac{RL}- based placement strategies have demonstrated significant runtime reductions while maintaining or improving \ac{QoR} \cite{murray2019adaptive}. Subsequent work incorporated timing and wirelength awareness into the learning process, further improving placement efficiency and solution quality \cite{elgammal2021rlplace}. More recent approaches integrate graph neural networks into the \ac{RL} loop to better capture structural dependencies within \ac{FPGA} netlists, enabling improved placement quality and faster convergence \cite{yang2025drlgofpga}. These works collectively highlight the effectiveness of \ac{RL} in navigating the large combinatorial design space inherent to placement problems.

Beyond placement, another important direction focuses on \textit{scheduling and runtime optimization}. In these approaches, \ac{RL} agents learn to manage execution order, resource allocation, and system timing under dynamic constraints. For instance, \ac{RL}-based schedulers have been used to optimize real-time digital simulation by modeling hardware constraints such as memory usage and communication availability within the learning environment \cite{guan2020frtds}. Complementary work has explored architectural support for stable \ac{RL} execution, including mechanisms such as Q-table memoization and reuse to improve convergence and runtime behavior \cite{mazaya2024reinforcement}. These studies demonstrate how \ac{RL} can adapt scheduling policies to complex and time-varying hardware conditions.

A third research direction addresses \textit{compiler and design-space optimization}. Here, \ac{RL} is used to explore large combinatorial spaces such as compiler pass ordering and hardware–software co-design configurations. For example, \ac{RL}-based methods have been applied to select optimal LLVM/\ac{HLS} compiler pass sequences, significantly improving performance compared to static optimization levels \cite{shahzad2022reinforcement}. Similarly, hardware–software co-exploration frameworks leverage \ac{RL} to jointly optimize neural architectures and accelerator parameters, enabling improved accuracy and reduced runtime \cite{lu2022hardware}. These approaches highlight the strength of \ac{RL} in guiding complex multi-dimensional optimization problems. {Recent work extends \ac{RL} to \ac{FPGA} logic synthesis by formulating optimization command selection as a design-space exploration problem. An archive-driven approach combined with \ac{PPO} enables efficient exploration of synthesis flows, leading to improved mapping quality in terms of \ac{LUT} reduction while satisfying timing constraints \cite{zhu2026archive}.} 

Finally, \textit{system-level optimization} focuses on using \ac{RL} to dynamically control system behavior during runtime. In these cases, the agent monitors hardware performance metrics and adapts system policies accordingly. For example, \ac{RL}-based runtimes have been used to dynamically select coherence modes in heterogeneous \acp{SoC}, improving performance and reducing memory traffic without modifying accelerator architectures \cite{zuckerman2021cohmeleon}. These approaches illustrate how \ac{RL} can optimize system-level behavior beyond individual hardware modules.

{Across these developments, a consistent theme is that \ac{RL} enables data-driven exploration of complex design spaces that are difficult to address using traditional heuristics. By learning from interaction with the system, \ac{RL} agents can adapt to workload variations and structural constraints, leading to improved efficiency and design quality. Nevertheless, \ac{RL}-based optimization on \ac{FPGA}/\ac{IC}  platforms remains constrained by several challenges, including training overhead, limited scalability to large state spaces, sensitivity to reward design, and restricted generalization under dynamic workloads. Addressing these limitations will require more scalable learning techniques and tighter integration between \ac{RL} models and hardware-aware design frameworks.}

\begin{figure}[!htbp]
    \centering
    \includegraphics[width=0.65\linewidth]{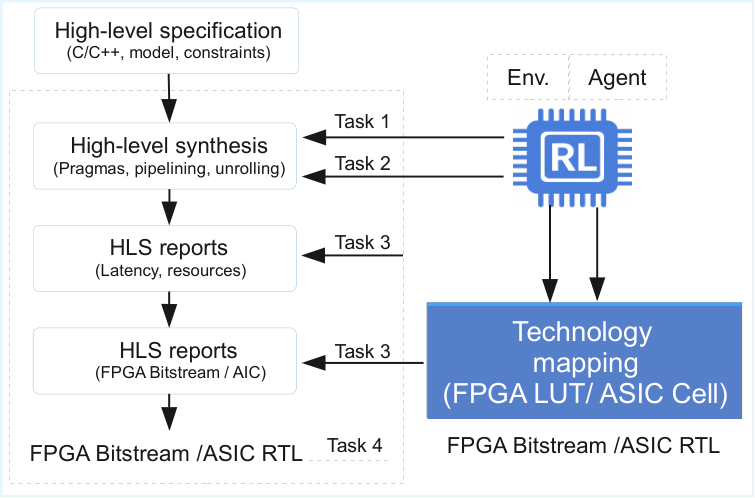}
    \caption{Unified \ac{RL}-assisted synthesis workflow covering both \ac{HLS} and logic-synthesis stages. \ac{RL} agents interact with several points in the design flow: Task (1) optimizing \ac{HLS} transformations; Task (2) incorporating feedback from specifications and \ac{HLS} reports; Task (3) guiding logic optimization; and Task (4) assisting \ac{FPGA} and \ac{ASIC} technology mapping. This illustrates how \ac{RL} provides an adaptive optimization layer across the full IC/\ac{FPGA} synthesis pipeline.}
    \label{fig:HLSLS}
\end{figure}

\subsubsection{RL-based High-Level and Logic Synthesis}

\ac{RL}- assisted synthesis has emerged as a significant research direction spanning logic optimization, \ac{FPGA}/\ac{ASIC} technology mapping, and \ac{HLS} directive tuning (Figure~\ref{fig:HLSLS}). In contrast to traditional synthesis flows based on fixed heuristic scripts, these approaches use \ac{RL} agents to explore large design spaces and automatically derive optimization strategies tailored to specific circuits and workloads.

Existing work in this area can be broadly categorized into three main directions: (i) logic-level optimization and technology mapping, (ii) \ac{HLS} directive and compiler optimization, and (iii) hybrid design-space exploration frameworks.

\textit{Logic-level optimization and mapping} represents one of the earliest applications of \ac{RL} in synthesis. These approaches replace manually crafted optimization sequences with learned policies that adapt to circuit characteristics. For example, \ac{RL}- based methods have demonstrated the ability to select circuit-specific logic-optimization sequences that reduce \ac{LUT} utilization and outperform traditional heuristic flows \cite{zhou2023area}. More recent frameworks extend this concept by unifying logic optimization and technology mapping within a single \ac{RL}- driven environment, enabling flexible synthesis strategies across both \ac{FPGA} and \ac{ASIC} targets \cite{qian2024efficient}. These works highlight the effectiveness of \ac{RL} in navigating the highly combinatorial optimization space of logic synthesis.

A second major direction focuses on \textit{\ac{HLS} optimization}, where \ac{RL} agents are used to automatically tune compiler directives, pragmas, and transformation strategies. Systems such as AutoAnnotate \cite{shahzad2024autoannotate} and AnnotationGym \cite{shahzad2025annotationgym} leverage \ac{RL} to generate optimized \ac{HLS} configurations for Vitis and Intel toolchains, achieving significant performance improvements over baseline and manually tuned designs. Further enhancements include improved policy-agent models for directive selection and ordering, which increase optimization stability and consistency across workloads \cite{munafo2023improved}. These approaches demonstrate how \ac{RL} can effectively replace manual tuning in complex \ac{HLS} workflows.

A third line of research explores \textit{hybrid learning frameworks} for multi-objective design-space exploration. In these systems, \ac{RL} is combined with predictive models to guide exploration more efficiently. For instance, \ac{RL} agents integrated with graph neural network predictors can generate Pareto-optimal trade-offs among latency, resource utilization, and timing during \ac{HLS} compilation \cite{wu2022ironman}. Similarly, \ac{RL} has been applied to fine-grained hardware parameter tuning, such as bit-width optimization in \ac{FPGA} accelerators, enabling automated balancing of accuracy, performance, and resource consumption \cite{jang2024reinforcement}. These approaches illustrate the growing role of \ac{RL} in managing complex, multi-dimensional optimization problems in synthesis flows.

{In light of these proposed approaches, a clear trend emerges in which \ac{RL} shifts synthesis from static, heuristic-driven processes toward adaptive, data-driven optimization. By learning directly from design feedback, \ac{RL} agents can tailor optimization strategies to specific circuits, toolchains, and performance objectives, improving both efficiency and solution quality across the \ac{FPGA} and \ac{IC} design stack. Despite these advances, \ac{RL}- based synthesis still faces notable limitations. Training remains computationally expensive, as agents must explore vast design spaces with limited sample efficiency. Generalization across circuits, workloads, and toolchains is often weak, requiring frequent retraining. Moreover, \ac{RL} policies can be opaque and unstable, posing challenges for reliability, verification, and integration into industrial electronic design automation workflows.}

\section{Applications of RL-based FPGA/IC platform}
\label{sec5}

\begin{figure}[!htbp]
    \centering
    \includegraphics[width=0.8\linewidth]{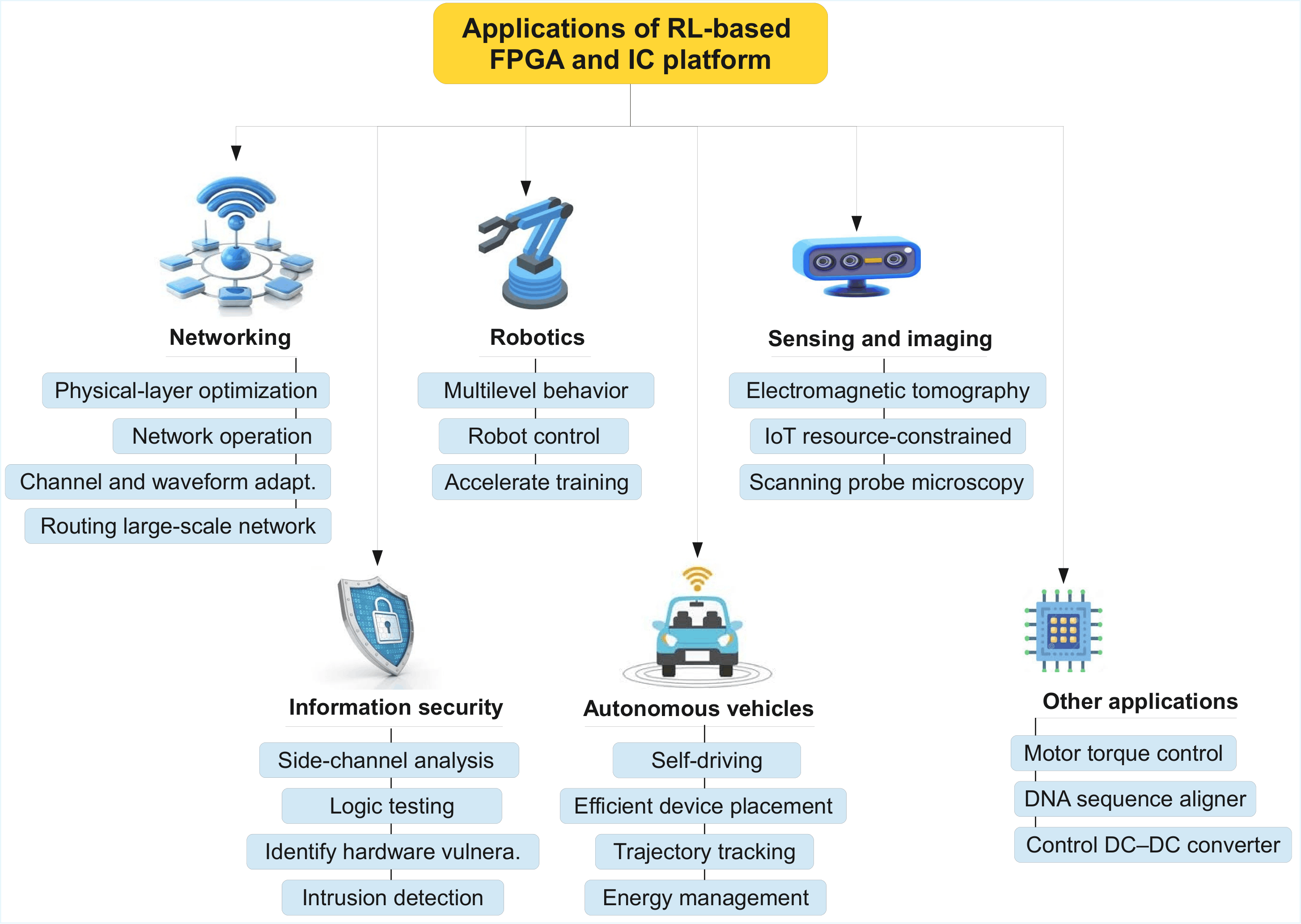}
    \caption{Application domains of \ac{RL}- based \ac{FPGA} and \ac{IC} platforms.}
    \label{fig:taxoApps}
\end{figure}

This section reviews the main application domains in which \ac{RL} has been deployed on \ac{FPGA} and \ac{IC} platforms, as summarized in Figure~\ref{fig:taxoApps}. The literature is organized by problem domain, highlighting how application-specific constraints—such as latency, energy efficiency, and real-time adaptability—shape \ac{RL} algorithm selection and hardware realization. Representative domains including networking, cybersecurity, intelligent sensing and imaging, and embedded control are discussed, with emphasis on reported performance gains and practical limitations. This domain-oriented perspective complements the earlier taxonomy and architectural analysis by clarifying where \ac{RL}-enabled \ac{FPGA}/\ac{IC} systems have demonstrated tangible real-world impact. A broader application-oriented summary of well-investigated FPGA-based \ac{RL} implementations, including target boards, dedicated tasks, limitations, and performance improvements over baseline methods, is presented in Table~\ref{table:7}.

{To further complement the domain-wise organization of this section, Table~\ref{tab:cross_domain_rl_fpga} provides a cross-domain comparison of RL algorithm selection and \ac{FPGA}/\ac{IC} implementation strategies. This comparison highlights how application requirements influence both the selected \ac{RL} family and the corresponding hardware realization. For instance, networking and autonomous systems generally favor \ac{DQN}, dueling \ac{DQN}, \ac{TD3}, \ac{PPO}, and \ac{MARL} methods because they require fast decision-making under dynamic and uncertain conditions. In contrast, hardware security applications frequently employ \ac{QL}, \ac{DQN}, \ac{PPO}, \ac{AC}, and \ac{HDC}-based \ac{RL} to support adaptive detection, verification, and vulnerability exploration. Robotics and motor-control applications commonly rely on \ac{QL}, \ac{DQN}, \ac{AC}, and \ac{DDPG}-like approaches due to their real-time feedback requirements, whereas sensing and imaging systems tend to use deep \ac{RL}, post-decision state  learning(PDS), \ac{SAC}, or \ac{PG} methods for adaptive acquisition and signal reconstruction. Across these domains, \ac{FPGA}/\ac{IC} implementations typically exploit pipelined datapaths, fixed-point or quantized arithmetic, \ac{BRAM}/\ac{LUTRAM}-based memory structures, hardware/software co-design, and parallel inference engines. }

\begin{table}[t]
\centering

\scriptsize
\caption{Cross-domain comparison of RL algorithm selection and FPGA/IC implementation strategies.}
\label{tab:cross_domain_rl_fpga}
\renewcommand{\arraystretch}{1.25}
\begin{tabular}{p{2cm} p{2cm} p{4.5cm} p{4cm} p{3.4cm}}
\hline
\textbf{App. domain} & 
\textbf{Typical RL algo.} & 
\textbf{Hardware implementation strategy} & 
\textbf{Main objective} & 
\textbf{Key constraints} \\
\hline

Networking & 
QL, DQN, dueling DQN, TD3, MARL & 
FPGA-accelerated routing, CNN/RL inference, distributed Q-tables, and pipelined decision engines & 
Low-latency routing, spectrum access, waveform adaptation, and traffic control & 
Dynamic traffic, scalability, communication overhead, and real-time response \\

\hline

Hardware and \newline embedded security & 
QL, DQN, PPO, AC, RL-RNN, HDC-based RL & 
RTL-level verification engines, FPGA-based detection pipelines, HDC/RL accelerators, and adaptive test engines & 
Vulnerability detection, intrusion detection, side-channel analysis, and security verification & 
Large state spaces, adversarial behavior, verification complexity, and limited generalization \\

\hline

Robotics & 
QL, DQN, AC, DDPG & 
Real-time FPGA control pipelines, fixed-point policy inference, and hardware-in-the-loop control & 
Fast feedback control, adaptive behavior, and RL training acceleration & 
Strict timing, limited resources, control stability, and sensor uncertainty \\

\hline

Autonomous \newline vehicles & 
PPO, AC,  DQN, MARL & 
CPU--GPU--FPGA task placement, FPGA-accelerated policy inference, and hardware-in-the-loop validation & 
Self-driving control, UAV tracking, trajectory optimization, and energy-aware operation & 
Safety, latency, energy consumption, and multi-objective optimization \\

\hline

Intelligent sensing and imaging & 
Deep RL, PDS-RL, SAC, PG methods & 
Lightweight accelerators, stochastic computing, FPGA--CPU hybrid control, and adaptive sensing pipelines & 
Feature selection, signal reconstruction, adaptive acquisition, and sensing optimization & 
Sensor noise, low-power operation, real-time imaging, and data uncertainty \\

\hline

Other applications & 
QL, DQN, model-free RL & 
FPGA-based inference/control cores, edge-training pipelines, and safety-filtered decision engines & 
Motor torque control, DNA sequence alignment, power converter control, and embedded optimization & 
Safety constraints, domain specificity, limited benchmarks, and deployment reliability \\
\hline
\end{tabular}
\end{table}

\renewcommand{\arraystretch}{1.30}
\begin{table*}[htbp]
\tiny
\centering
\caption{Summary of well-investigated FPGA-based \ac{RL} applications, including \ac{FPGA} boards used, dedicated tasks, limitations, and performance improvements over baseline methods.}

\label{table:7}
\begin{tabular}{m{0.2cm}m{0.5cm}m{2cm}m{1.2cm}m{6cm}m{1.5cm}m{2cm}m{1.5cm}}
\toprule
\textbf{Apps.} & \textbf{Ref.} & \textbf{FPGA used} & \textbf{RL} & \textbf{Dedicated task and limitations} & \textbf{Compared to} & \textbf{Result} & \textbf{Improvement} \\
\midrule\hline
& {\cite{choudhary2026scalable}}  & Nexys A7 / PYNQ-Z2 & QL 
& Proposes a QL-based fault- and congestion-aware NoC routing framework with a compact Q-table; however, it relies on offline training, synthetic evaluations, and lacks formal deadlock guarantees.
& HARAQ, RELAR 
& 0.95 delivery rate at 20\% faults
& -- \\\cline{2-8}
 & \cite{kim2020reinforcement} & PXIe-7975 &  REINFORCE & Employs \ac{RL} to develop a flexible duplex system, targeting maximization of area spectral efficiency. However, the system's reliance on the OP-map for transmission decisions limits it in dynamic environments.& Full duplex OP-map & Better  spectral efficiency & -- \\\cline{2-8}

\multirow{6}{*}{\rotatebox{90}{Networking}}& \cite{lent2020evaluating}  & Zynq Z7020 & Model-free \ac{RL} &  The work targeted accelerated \ac{RL}- based routing in delay-tolerant networks using FPGA, but faced scalability constraints,  and communication overhead.  & No \ac{RL} (CPU) & Response Time $\approx$ 2.5s \newline Throughput $\approx$ 41 bundles/s & $\times$32 faster\\\cline{2-8}
& \cite{d2021can} & Xilinx xc7z045ffg900-2 &  TD3 & Real-time waveform synthesis improves resilient wireless signal classification, but faces complexity, \ac{FPGA} usage, training instability, and sensitivity to channel noise. & No-FIR \newline DeepRadioID  &  Acc. (High SNR) =100\%  & 4.1$\times$ (No-FIR) \newline 1.9$\times$ (DeepRadioID) \\\cline{2-8}
& \cite{kim2024fpga} & PYNQ-Z2 &  Dueling-\ac{DQN} & Enabled real-time Dueling-\ac{DQN} routing on LEO satellites using FPGA-accelerated CNNs.  But faced higher design complexity, resource limits, and scalability constraints. & \ac{CPU}- only  & Inf. time =0.3 s & 3.10$\times$ faster\\\cline{2-8}

& \cite{jagadheesh2022reinforcement} & Kintex-7 FPGA KC705 & QL-MARL& It learned adaptive fault-tolerant routing paths in mesh NoCs, but required per-router Q-tables and added hardware overhead with higher resource usage. & QL–based routing& Packet delivery +7.1\%& 2.7\% fewer hardware resources and 0.89\% less power \\\cline{2-8}

& {\cite{chen2025mooo}} & Cyclone IV FPGA & RDQN 
& RDQN-based SDN controller placement with FPGA-based monitoring; however, DRL training is software-based and evaluated mainly via simulations. 
& PSA 
& 59.39\% worst-case latency reduction 
& 42.49\% average latency reduction \\\cline{2-8}

& {\cite{yaseen2025neurodpd}} & FPGA-based 28 GHz testbed & DQN 
& DRL-based digital predistortion for nonlinear RF-chain compensation in 6G; nevertheless, validation is limited to controlled RF/PA test conditions and requires lookup-table memory for real-time adaptation.
& Conventional MP-DPD 
& 12.3 dB improvement 
& 36\% EVM reduction \\\hline

& \cite{chen2023adatest} & Zynq ZC706 &  \ac{RL} & Adaptive sampling for in-chip \ac{HT} detection. AdaTest suffer from high computational complexity, reliance on \ac{RL} tuning, and scalability challenges for larger circuits.  & MERO \newline TRIAGE  & -- \newline -- & DR= +15.61\% \newline 0.63× Speedup \\\cline{2-8}

\multirow{4}{*}{\rotatebox{90}{Security}}&  \cite{hashemi2018approximate} & Spartan6 &  RL-RNN & Reduce the time taken from capturing the image to encoding the iris. However,  the \ac{FPGA}'s resources may restrict the system's complexity and scalability. & Local search  & 48$\times$ Speedup & +6$\times$ Speedup  \\\cline{2-8}

& \cite{mondol2024rl} & RTL-level SoC benchmarks: AES, RSA, RS232 & 
PPO, AC & 
Automated pre-silicon security verification to detects vulnerabilities in confidentiality, integrity, and availability.  Evaluated only at RTL level, not yet extended to system/device level. & 
Cadence JasperGold & 
 66.6\% SAM, \newline 48\% RNM & 
-- \\\cline{2-8}

& \cite{issa2024cyberrl}  & Xilinx Alveo U50 & Double \ac{QL}  & Brain-inspired \ac{RL} using \ac{HDC} for adaptive attack/defense strategy learning in Markov games. However, it scalability testing limited to simulated environments. & \ac{DQN}  & 1.9$\times$ speedup in  training, 12.8$\times$ higher rewards & $\approx$+700$\times$ energy efficiency  \\\hline
&  \cite{dumesnil2017single} & Xilinx Spartan-6 & CC-OC \ac{RL} & Implementing adaptive learning for robotic control using conditioning behaviors. However, it suffer from  \ac{FPGA} resource constraints, limited scalability for complex tasks, and potential real-time processing delays.& --   & Very low resource demands & -- \\\cline{2-8} 
\multirow{4}{*}{\rotatebox{90}{Robotics}}& \cite{hwang2009modular}  & Virtex II &  AC & Develop a mobile robot capable of performing goal-seeking, cruising, and safety assurance tasks within an unstructured environment. The limitations are in generalizability, scalability, learning efficiency, and resource scarcity. & --  & Tasks done & -- \\\cline{2-8} 
& \cite{kadokawa2021binarized} & Avnet Ultra96-V2 &  BPN & Real-time control of robots using raw image data on \ac{FPGA}. \ac{BPN} is limited to discrete action spaces, restricting applicability to continuous control tasks. & \ac{DQN} \newline BQN  & IT=4 ms \newline TT= 11.6 ms & IT=-999 ms \newline TT= +6$\times$\\\cline{2-8}
& \cite{guo2019customisable} & Intel Stratix-V &  DDPG & Accelerate \ac{RL} for robotic control using \ac{FPGA}-based DDPG implementation. The method is restricted to \ac{FPGA} resource limits, affecting scalability for larger neural networks.& DDPG (CPU)  & NEAG = 600 Episodes & NEAG =  -200 Episodes \\\cline{2-8}
& \cite{gankidi2017fpga} & Xilinx Virtex-7  &  QL-MLP & \ac{FPGA} fixed-point accelerator for \ac{QL} in planetary robotics; no full rover task demo or on-board hardware validation  & CPU – Intel i5 2.3 \newline Virtex 7 (Floating)  & 95$\times$ (single neuron) and 43$\times$ (MLP) faster than CPU & $\approx$-29\% power vs floating-point MLP\\\hline

& {\cite{deng2026fpga}} & Intel Agilex 7 & DRL 
& Heterogeneous FPGA-CPU-GPU DRL framework for multi-environment decision-making; yet relies on offline scheduling and limited flexibility for major model changes.
& CPU-only 
& 41.28 GFLOPS/W energy efficiency 
& 3.88$\times$ (vs. CPU) \\\cline{2-8}
\multirow{4}{*}{\rotatebox{90}{\makecell{Autonomous  vehicles}}} & \cite{zheng2021device}  & ALINX &  PPO & RL-based device placement for autonomous vehicle DNNs minimizing latency/energy; only offline benchmarks, no real-vehicle deployment  & PG,\newline single-GPU \newline single-CPU & Average cost =0.7 (MNIST) & PPO= -0.3 (PG) \newline PPO=-0.8 (GPU) \newline PPO=-1.7 (CPU) \\\cline{2-8}
&  \cite{li2019fast}  & Altera 5CGX  &  AC & Framework for \ac{UAV} trajectory tracking through predefined waypoints. \ac{FPGA}'s limited resources cause latency issues, while \ac{RL} framework demands significant training time and resources.& PID approach  & PE= 0.2399 & PE= -0.3333\\\cline{2-8}

& \cite{shi2022online} & ZYNQ7020 & MARL & Online energy management of multi-stack fuel cell hybrid vehicle; limited scalability and offline training dependency & Fuel economy & 92.86 Fuel economy \newline 0.170 ms Computation time & +1.2 fuel economy \\\cline{2-8}

& \cite{damodarin2025smart} & AMD ZCU104 &  QL & A practical architecture for \ac{EV} charging within modern smart grid systems. Nevertheless, dynamic real-time conditions and user preferences are not explored. & Software-only \ac{RL} scheduling  & Latency = 34.3 ns \newline  Total power = 136 mW & --\\\hline

& {\cite{betancourt2026intelligent}}  & Xilinx Zynq UltraScale+ & QL 
& DRL-based power allocation for underwater MIMO-NOMA underwater visible light communication (UVLC) under realistic turbulence; yet relies on simulated oceanographic data and complex turbulence modeling.
& Water-filling 
& 50\% BER reduction 
& 7 dB SNR gain \\\cline{2-8}

\multirow{4}{*}{\rotatebox{90}{\makecell{Intelligent    sensing}}} & \cite{cao2025adaptive} & Not specified &  Double \ac{DQN} & Optimized adaptive EMT imaging using Double \ac{DQN} for feature selection; limited by low-resolution and absence of evaluation across diverse EMT configurations. & Traditional method & Image quality (10\% feature) > 85\% & Total time= -21\%\\\cline{2-8}

& 
\cite{sun2022hardware} & 32-nm Stochastic computing &  PDS-RL & Accelerated post-decision state \ac{RL} for delay-sensitive IoT sensor control; limited to tabular models, single scenario, ASIC-oriented evaluation only, without FPGA. & QL  & Delay= 98.76 ns \newline Power= 5.87 mW & 5.3$\times$ faster than QL\newline Path delay= -87.9\%\\\cline{2-8}

& \cite{wei2025novel} & cRIO-9064 &  Soft AC & FPGA-assisted \ac{RL} tuning of P–I gains in commercial SPM; only one SPM setup and sample set, generalization and fully on-board \ac{RL} untested & MFP-3D  & Mean deflection error=  0.108\,nm (SD 0.146\,nm) & Up to - 89.8\% in mean deflection error and -88.0\%  SD\\

\bottomrule\hline
\end{tabular}
\begin{flushleft}
Abbreviations: Position error (PE);  Inference time (IT); number of episodes to achieve goal (NEAG); classical conditioning (CC); operant conditioning (OC); opportunity (OP)
\end{flushleft}
\end{table*}

\subsection{Networking}

\ac{RL} has emerged as a powerful paradigm for optimizing decision-making across modern networking systems, supporting tasks such as traffic management, routing, interference mitigation, and network security. When combined with \ac{FPGA} acceleration, \ac{RL} gains substantial benefits—including fine-grained parallelism, reduced latency, and energy-efficient real-time operation—which are essential for coping with dynamic and resource-constrained communication environments. As networking demands continue to intensify, \ac{RL}-based \ac{FPGA} solutions offer scalable and high-performance architectures capable of adapting online to rapidly changing conditions.

To illustrate these advantages, several works explored \textit{physical-layer and spectrum-level optimization}, where real-time constraints are particularly stringent. For instance, persistent spectrum scarcity in 5G/B5G networks drives the need for advanced spectral-efficiency techniques such as \ac{DSA} and flexible duplexing. The study in \cite{kim2020reinforcement} addressed this challenge by introducing an \ac{RL}-driven flexible-duplex system operating in sensor-assisted mode, demonstrating that deploying \ac{RL} on real hardware could meet strict timing thresholds required for agile duplex switching. Building on the theme of \textit{network operation under uncertainty}, other works investigated \ac{RL}-enabled control for highly disrupted or delay-tolerant settings. The study in \cite{lent2020evaluating} examined routing in space networks characterized by long propagation delays, intermittent connectivity, and limited global state knowledge. The proposed cognitive controller used \ac{RL} with a \ac{SNN} implemented on \ac{FPGA} to enable faster-than-real-time routing decisions, showing how neural-based \ac{RL} hardware can sustain adaptability even in severely challenged environments. A related direction focused on \textit{channel-level compensation and waveform adaptation}, where the environment is highly dynamic and often unpredictable. The work in \cite{d2021can} introduced Chares, a \ac{TD3}-based real-time waveform synthesis system designed to counteract channel distortions and recover \ac{DL}-based wireless signal classification accuracy. By generating channel-specific FIR filters, the \ac{RL}-enabled \ac{FPGA} system improved accuracy by up to 4.1$\times$ under unseen conditions. While the approach offers microsecond-level reaction time and energy-efficient inference, it introduces additional complexity and sensitivity to noise dynamics, highlighting the practical challenges of deploying continuous-control \ac{RL} on hardware.

In parallel, several works explored \textit{\ac{RL}-accelerated routing for autonomous or large-scale network} infrastructures. In dynamic LEO satellite networks, \cite{kim2024fpga} developed an \ac{FPGA}-parallelized \ac{CNN} architecture enabling real-time Dueling-\ac{DQN} routing. The CPU–\ac{FPGA} co-design achieved substantial acceleration without degrading routing accuracy, demonstrating the viability of hardware-assisted \ac{RL} for onboard satellite decision making where deterministic latency is crucial. In the following,  at the architectural-level of chip interconnects, the work in \cite{jagadheesh2022reinforcement} employed a \ac{QL}-based \ac{MARL} strategy for resilient routing in \ac{NoC} environments. Each router maintained a two-level Q-table and autonomously learned forwarding decisions under faulty links or nodes. Implementing this scheme on an \ac{FPGA} enabled low-latency routing, high fault tolerance, and efficient hardware utilization, showcasing how distributed \ac{RL} agents can deliver scalable and self-optimizing network-on-chip behavior.

{From the perspective of the proposed taxonomy, networking applications span both roles of RL. Latency-critical tasks such as routing, spectrum access, and real-time waveform adaptation predominantly position \ac{RL} as the \emph{workload}, requiring \ac{FPGA}/\ac{IC}  acceleration to meet strict timing constraints. In contrast, higher-level network management and optimization problems—such as congestion control or adaptive routing policies in large-scale networks—lean toward \emph{\ac{RL} as the optimizer}, where the primary challenge lies in handling large and dynamic state/action spaces.}

\subsection{Hardware and embedded security}

\begin{figure}[!htbp]
    \centering
    \includegraphics[width=0.6\linewidth]{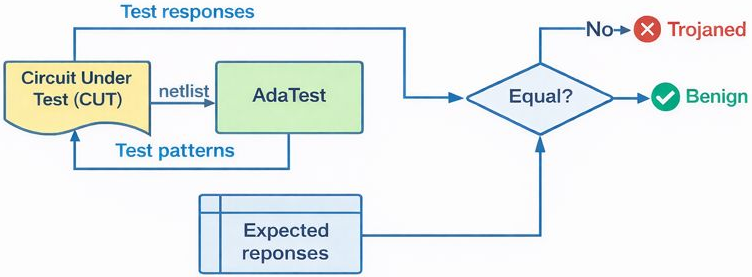}
    \caption{Operational flow of AdaTest, an \ac{RL}-based hardware Trojan (\ac{HT}) detection framework \cite{chen2023adatest}. 
Adaptive test patterns generated by the agent are applied to the \ac{CUT}, and the resulting responses are compared with expected outputs: mismatches indicate a \textit{Trojaned} circuit, while matching responses classify the design as \textit{Benign}.}
    \label{fig:HT}
\end{figure}

The design and production process of modern \acp{IC} are often delegated to external (untrusted) parties. This outsourcing in the supply chain leads to hardware security issues, including risks of sensitive data leakage, reduced performance, and potential copyright violations. At every stage of the \ac{IC} supply chain, malicious alterations, commonly known as \ac{HT} attacks, can be introduced. \ac{HT} detection methods are divided into two main types: \textit{\ac{SCA}}, which identifies physical alterations and  parameter changes, and \textit{logic testing}, which activates concealed triggers by using a variety of test patterns. For example, 
AdaTest \cite{chen2023adatest} is an automated, adaptive test pattern generation framework for \ac{HT} detection, using \ac{RL} in hardware accelerator design. It operates in two phases: circuit profiling, which identifies rare nodes using transition probability and \ac{SCOAP} testability measures, and adaptive test pattern generation, which employs a reward function to select test inputs. AdaTest effectively balances trigger coverage and test generation time through iterative testing. As illustrated in Figure \ref{fig:HT}, AdaTest is utilized to verify whether any \acp{HT} are embedded in the \ac{CUT}. It takes the netlist of the \ac{CUT} and user-defined parameters as inputs, subsequently generating a set of test vectors with high reward values as output.  

A related approach, \cite{hashemi2018approximate} utilizes an end-to-end iris recognition system integrating approximate computing with an \ac{FPGA}-based hardware/software pipeline. It employs \ac{RL} with a \ac{RNN} agent to optimize eight approximation knobs in the design. The \ac{FPGA} accelerates critical tasks, providing a 48$\times$ runtime reduction while meeting accuracy standards. Security is enhanced by processing sensitive biometric data locally, reducing exposure to threats.

The paper in \cite{mondol2024rl} proposes RL-TPG, a \ac{RL}–based framework for automated pre-silicon security verification on \ac{SoC} hardware at the \ac{RTL}. Using an \ac{RL} agent, it generates intelligent test patterns to \textit{identify vulnerabilities in hardware} modules such as advanced encryption standard (AES), Rivest–Shamir–Adleman (RSA), and RS232 cores. By combining static analysis, dynamic mutation, and multiple coverage metrics, RL-TPG achieves about 90\% verification coverage. In \cite{issa2024cyberrl}, a brain-inspired \ac{RL} framework based on hyperdimensional computing was proposed for efficient \textit{intrusion detection}. The approach combined double \ac{QL} with \ac{HDC} encoding, where states and actions were represented as high-dimensional hypervectors to enable fast, noise-tolerant learning and memory-like generalization. Each agent’s state vector $s_i$ was projected into high-dimensional space using the base hypervector matrix $B_i$ and its corresponding feature coefficients $\mathbf{d_i}$, which define the weighted contribution of each state feature. The encoded hypervector $\mathbf{H_s} = \sum_i \mathbf{d_i}B_i$ was then processed through kernel transformations $e^{jx} = \cos(x) + j\sin(x)$ and a systolic array to achieve rapid and energy-efficient Q-value computation.

{In this domain, \ac{RL} is primarily positioned as the \emph{optimizer}, where it is used to detect vulnerabilities, guide security policies, or adapt system behavior under adversarial conditions. These problems often involve complex and high-dimensional state spaces rather than strict real-time constraints. However, in certain cases such as real-time intrusion detection or side-channel analysis, aspects of \emph{\ac{RL} as the workload} emerge, requiring accelerated inference for timely threat mitigation.}

\subsection{Robotics}

In industrial applications, robotic controllers necessitate a low-latency computation process to meet stringent real-time constraints. This ensures prompt response to dynamic environments, allowing for precise manipulation and coordination of robotic tasks. The effectiveness of such systems hinges on their ability to process data rapidly and efficiently under varying operational conditions \cite{kebir2018smart}. For example, the proposed architecture in \cite{hwang2009modular} is tailored for hardware implementation in \textit{multilevel behavioral} robots, allowing them to autonomously perform intricate tasks such as goal-seeking, cruising, and ensuring safety. It consists of modules with similar structures, primitive actions, and compositional behaviors, all stemming from a basic template. Each module is designed to learn and collaborate in managing diverse tasks, utilizing a \ac{RL} approach based on an \ac{ARM} with a \ac{AC} model. This architecture is realized on a \ac{FPGA} chip that includes a \ac{CPU} core, which enables efficient parallel processing of neural networks. Tested on a mobile robot, the system effectively executes these tasks in unstructured environments, adeptly navigating obstacles and challenging scenarios.

Although \acp{FPGA} are more power-efficient than \acp{CPU} and \acp{GPU}, typical \ac{RL} methods are unsuitable for \ac{FPGA} implementation because \acp{FPGA} are composed of many logic blocks optimized for high-speed logical operations but not for real-number operations. To address this limitation, the scheme \cite{kadokawa2021binarized} introduces a \ac{BPN}, a novel \ac{RL} algorithm that leverages binarized \acp{CNN} for \textit{robot control} from raw images. The learning system involves three main steps: (i) \ac{FPGA} and \ac{CPU} collect datasets for learning by controlling the robot, (ii) a GPU server updates the \ac{BPN} network parameters based on collected data, and (iii) the \ac{FPGA} updates the binarized network parameters for real-time control. This architecture effectively enables high-speed control on the \ac{FPGA}, despite its real-number calculation limitations.

The research in \cite{guo2019customisable} recognizes that training robots directly through \ac{RL} can cause physical damage and be time-consuming. Although training robots in a simulated environment reduces costs and improves efficiency, the policy learning process still remains time-consuming. For this reason, the authors proposed an \ac{FPGA}-based hardware implementation of the \ac{DDPG} algorithm to \textit{accelerate training}. The \ac{FPGA}, specifically the Intel Stratix-V on the Maxeler MAX4 platform, handles gradient calculations and network parameter updates, allowing for faster and safer policy learning. This \ac{RL} approach enables training in continuous action spaces, providing more precise control for a 3D-printed robotic arm. The best result achieved a speedup of 4.07$\times$ for model one. However, the method is limited by I/O bottlenecks between the \ac{CPU} and \ac{FPGA}, which restrict potential speed gains. Similarly, \cite{gankidi2017fpga} addresses the challenges of implementing \ac{QL} onboard planetary rovers and spacecraft, which face constraints like limited processing power and radiation-hardening requirements. It proposes an \ac{FPGA}-based \ac{QL} implementation with \acp{ANN} to overcome these issues, leveraging \ac{FPGA}'s fine-grain parallelism to significantly reduce processing time. Using a Xilinx Virtex 7 \ac{FPGA}, the study demonstrates both single-neuron and \ac{MLP} \ac{QL} accelerators, achieving up to a 43-fold speedup compared to a conventional Intel i5 2.3 GHz CPU. The architecture’s performance and power consumption are further evaluated using the Symphony simulator and Xilinx compiler. However, the method's limitations include high power consumption, lack of pipelining for efficiency, and limited real-world testing on robot platforms.

{Robotics applications typically lie closer to \emph{\ac{RL} as the workload}, as they demand low-latency decision-making for control and interaction with physical environments. \ac{FPGA}/\ac{IC}  acceleration is therefore critical for enabling real-time inference and, in some cases, on-device training. Nevertheless, tasks involving high-level planning or policy learning in complex environments also reflect characteristics of \emph{\ac{RL} as the optimizer}, particularly when large state and action spaces are involved.}

\subsection{Autonomous vehicles}

Vehicles utilizing \ac{RL} on \ac{FPGA} can encompass both \textit{self-driving} cars and \acp{UAV}, offering real-time decision-making with efficient processing and enhanced autonomy. Self-driving vehicles are becoming a global trend, with manufacturers like Tesla investing heavily in autonomous systems that rely on real-time sensing, information fusion, and decision-making. \acp{DNN} are widely used for environmental sensing and control due to their success in other fields. As autonomous driving becomes commercialized, the need for powerful computing platforms to support tasks like environment perception, motion control, and communication within the Internet of vehicles grows.

\begin{figure}[!htbp]
    \centering
    \includegraphics[width=0.7\linewidth]{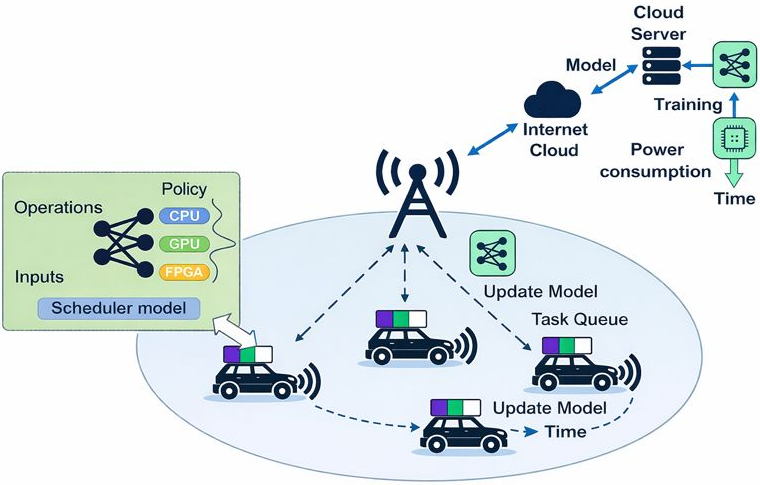}
    \caption{\ac{RL}- based Internet of vehicles model. The vehicle primarily consisting of a CPU, GPU, and \ac{FPGA} placement model, is trained on a cloud server with the objective of reducing both execution latency and energy consumption using \ac{RL} \cite{zheng2021device}.}
    \label{fig:IoV}
\end{figure}

To address these computational demands, \cite{zheng2021device} investigated the optimal allocation of \ac{AI} tasks in autonomous vehicles equipped with heterogeneous processing units, including \ac{CPU}, \ac{GPU}, and \ac{FPGA}. Their work proposed a \ac{PPO}-based \ac{RL} algorithm to determine \textit{efficient device placement} for neural network operations, as illustrated in Figure \ref{fig:IoV}. A sequence-to-sequence model was used for task assignment, rewarding configurations that minimized execution time and energy consumption. Experiments showed that this approach surpassed state-of-the-art \ac{RL} and traditional heuristics across several benchmarks. Building on the theme of real-time autonomous control, \cite{li2019fast} shifted the focus from ground vehicles to aerial platforms, employing an \ac{AC} framework for \ac{UAV} \textit{trajectory tracking} through predefined waypoints. Here, the \ac{FPGA} accelerated the control pipeline, delivering a throughput of 204.96 actions per second at only 33.57~$\mu$W of power. This hardware–accelerated \ac{RL} solution achieved notable gains, including a 58.14\% reduction in tracking error, 21.77\% lower power consumption, and a 9.23\% faster response time compared to baseline \ac{PID}-based controllers. Nevertheless, the study also highlighted practical challenges such as \ac{FPGA} resource limitations requiring time-multiplexing—potentially introducing latency—and substantial training requirements for the \ac{RL} model. Extending \ac{RL}-driven autonomy beyond navigation and toward energy optimization, \cite{shi2022online} explored independent \ac{QL} (IQL), a multi-agent \ac{RL} technique, for \textit{energy management} in multi-stack fuel cell hybrid vehicles. In this setting, each fuel cell operated as an autonomous agent balancing hydrogen consumption and battery condition. Their results demonstrated improved fault tolerance, real-time adaptability, and enhanced fuel economy, validated through hardware-in-the-loop experiments. This contribution further illustrates how \ac{RL}—combined with hardware acceleration—can support increasingly complex autonomous vehicle functions.

{Autonomous driving systems strongly emphasize \emph{\ac{RL} as the workload}, due to stringent real-time and safety requirements that necessitate fast and deterministic inference on hardware platforms. At the same time, elements such as trajectory planning and multi-objective decision-making introduce optimization challenges that align with \emph{\ac{RL} as the optimizer}, especially when dealing with complex environmental dynamics and large-scale state representations.}

\subsection{Intelligent sensing and imaging systems}

Intelligent sensing and imaging systems combine advanced sensors with AI-driven data processing to perceive, interpret, and reconstruct complex environments in real time. By integrating \ac{RL}, signal processing, and adaptive decision-making, these systems enhance accuracy, efficiency, and automation across applications such as industrial inspection, medical diagnostics, and autonomous systems.

Within adaptive imaging, Cao et al. \cite{cao2025adaptive} advanced \textit{electromagnetic tomography} (EMT) through a deep \ac{RL}-driven feature–selection strategy that identifies the most informative measurements for reconstruction. Their AMEMT-Net architecture incorporates historical actions and partial reconstructions to guide acquisition, cutting down redundant sampling while improving reconstruction fidelity. Both simulation and FPGA-based evaluation confirmed strong robustness to noise and reliable generalization from synthetic to real-world imaging conditions, illustrating the potential of \ac{RL}-adaptive sensing for high-efficiency imaging pipelines. Complementing these algorithmic advances, Sun et al. \cite{sun2022hardware} addressed the challenge of deploying \ac{RL} in \textit{resource-constrained wireless \ac{IoT} nodes} by designing lightweight hardware accelerators for postdecision state (PDS) learning. Their solution combines an arithmetic action-evaluation engine with a stochastic-computing architecture that replaces multipliers with bitwise operations, thereby reducing latency, area, and power consumption. By leveraging structured parallelism and compact state encoding, the proposed accelerator achieved substantially faster convergence and improved efficiency in delay-sensitive IoT scenarios when benchmarked against \ac{QL} and traditional baselines. This line of work highlights how energy-efficient hardware co-design can make embedded \ac{RL} feasible at the edge. Further downstream in precision instrumentation, Wei et al. \cite{wei2025novel} demonstrated how an \ac{RL}-enabled controller can enhance \textit{scanning probe microscopy} (SPM) through dynamic tuning of P–I gains. Their PICTS platform, implemented as a parallel FPGA–CPU architecture, used a \ac{SAC} agent to stabilize probe motion and improve image quality on difficult samples. The system achieved reductions in mean deflection error of nearly 90\% and in variance of about 88\% relative to a commercial fixed-gain controller. Nonetheless, evaluation was limited to a single SPM platform with a narrow sample set and partially relied on off-board training, leaving room for broader experimental validation and comparison to advanced nonlinear or model-predictive controllers.

{Applications in sensing and imaging exhibit a mixed positioning along the spectrum. Real-time perception and signal processing tasks typically require \emph{\ac{RL} as the workload}, benefiting from hardware acceleration to meet latency and energy constraints. Conversely, system-level adaptation, feature selection, or reconstruction strategies often place \ac{RL} in the role of \emph{optimizer}, where the complexity of the problem dominates over strict timing requirements.}

\subsection{Other applications}


The scope of \ac{RL}-based \ac{FPGA}/\ac{IC} acceleration has likewise broadened beyond the applications described earlier. For example, the work in \cite{schenke2023finite} proposed a safe \ac{DQN}–based finite-set direct \textit{torque control} for a permanent magnet synchronous motor, combining an edge-computing training pipeline with \ac{FPGA}-based \ac{DQN} realized as an \ac{MLP} inference for real-time torque control. A safeguarding layer uses online recursive least squares identification to block unsafe switching actions, and the best reported agent achieved fast torque step responses of about 5~ms while respecting current and voltage limits, with training converging in under 10~minutes on the MicroLabBox \ac{FPGA}. Yet, single motor/test bench, performance still below state-of-the-art model predictive control, high hardware/implementation complexity, and untested generalization to other drives or grid conditions. In a related effort, \cite{lall2023deep} proposed EdgeAlign, a compact dueling \ac{DQN}-based \textit{DNA sequence aligner} tailored for embedded edge devices (Jetson Nano and Arty A7-100T FPGA). Sequences are encoded as images and processed with a \ac{CNN}-based agent using BLAST score–driven rewards and a sliding-window scheme. AutoKeras compresses the model from $\sim$1.36M to $\sim$98k parameters, achieving 11.79\% mean BLAST-score error and 2.32ms/action on Jetson, and 1s/action on \ac{FPGA} at 100MHz. Nevertheless, the study was conducted only on influenza data, with non-optimal error, relatively slow \ac{FPGA} inference, and a fixed window size. In addition  \cite{chen2025resddqn} presented ResDDQN–FPGA, a \ac{DQN} with residual connections deployed on an \ac{FPGA} to \textit{control} a full-bridge LLC circuit resonant DC–DC converter.  The framework parallelizes the action–value function (Q) and target networks and partitions inputs to accelerate inference, improving dynamic response, steady-state accuracy and efficiency compared with sliding-mode control and model predictive control under input- and load-voltage variations. Experiments showed the fastest voltage recovery and lower energy consumption among the tested methods. However, some limitations noticed, including single converter platform, no detailed \ac{FPGA} resource/power analysis, and limited robustness and aging evaluation.

{This category encompasses diverse use cases with varying characteristics across the spectrum. Control-oriented tasks such as motor control or power electronics generally align with \emph{\ac{RL} as the workload}, due to their real-time constraints. In contrast, applications such as design-space exploration or sequence optimization tend toward \emph{\ac{RL} as the optimizer}, where scalability and decision complexity are the primary challenges.}

\begin{figure}[!htbp]
    \centering

    \begin{subfigure}{0.46\linewidth}
        \centering
        \includegraphics[width=\linewidth]{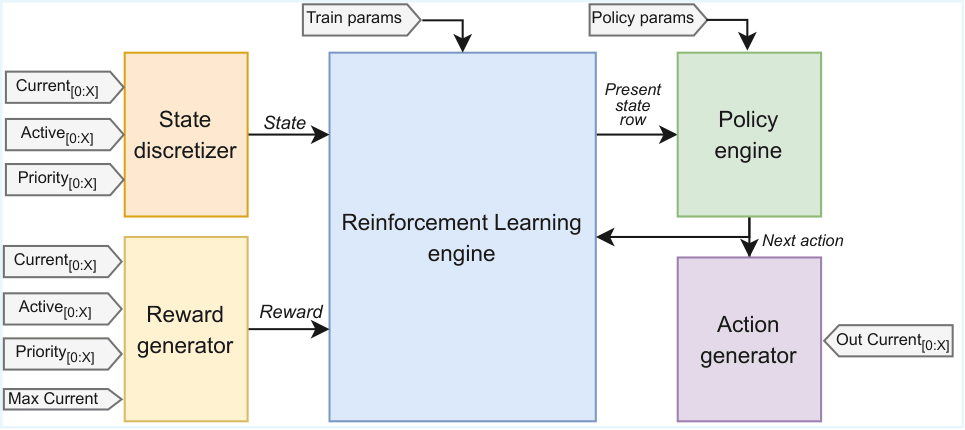}
        \caption{}
    \end{subfigure}\hfill
    \begin{subfigure}{0.46\linewidth}
        \centering
        \includegraphics[width=\linewidth]{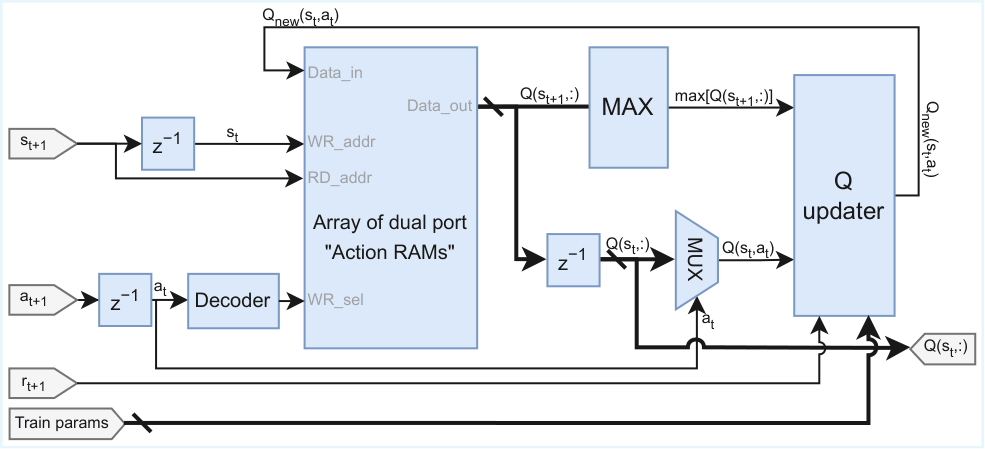}
        \caption{}
    \end{subfigure}

    \vspace{0.6em}

    \begin{subfigure}{0.48\linewidth}
        \centering
        \includegraphics[width=\linewidth]{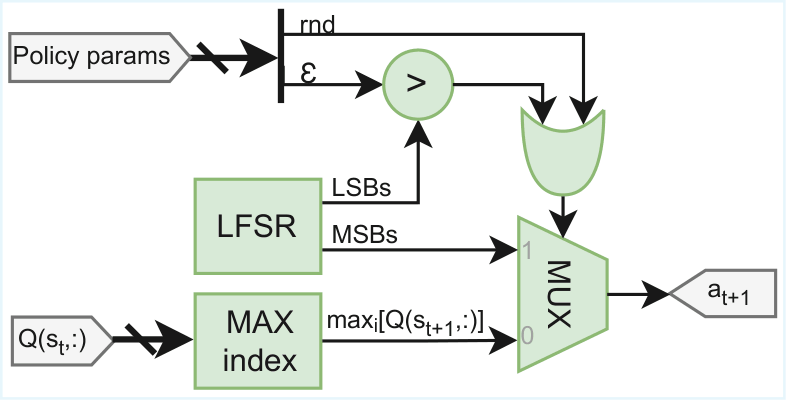}
        \caption{}
    \end{subfigure}\hfill
    \begin{subfigure}{0.48\linewidth}
        \centering
        \includegraphics[width=\linewidth]{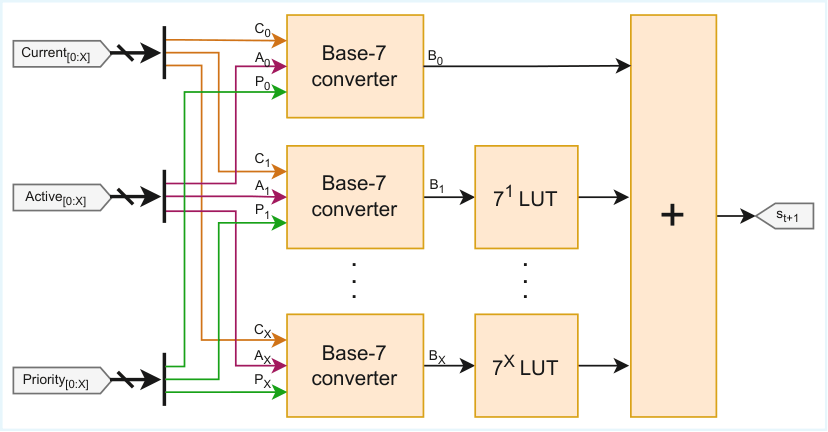}
        \caption{}
    \end{subfigure}

    \vspace{0.6em}

    \begin{subfigure}{0.48\linewidth}
        \centering
        \includegraphics[width=\linewidth]{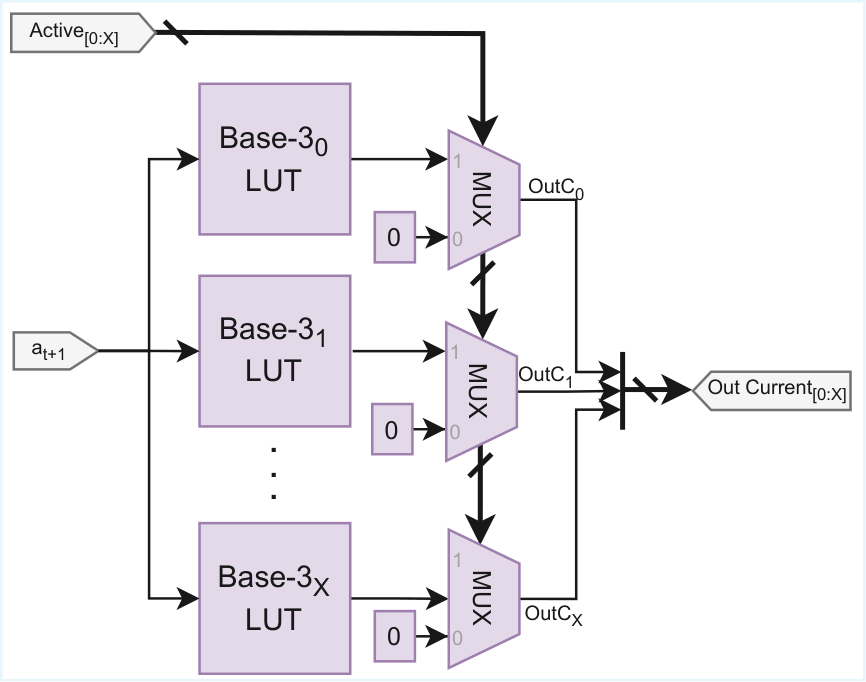}
        \caption{}
    \end{subfigure}\hfill
    \begin{subfigure}{0.48\linewidth}
        \centering
        \includegraphics[width=\linewidth]{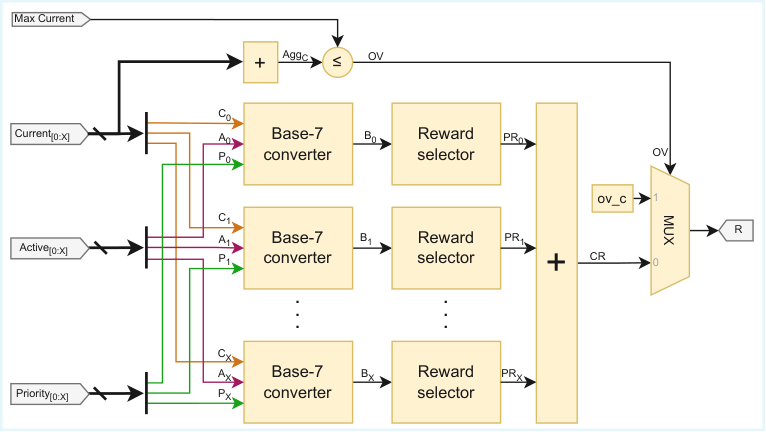}
        \caption{}
    \end{subfigure}

    \caption{RL-based \ac{FPGA} EV charging management system:
(a) overall \ac{FPGA} architecture;
(b) \ac{RL} engine;
(c) policy engine with LFSR-based exploration;
(d) state discretizer;
(e) action generator;
(f) reward generator
 \cite{damodarin2025smart}.}
    \label{fig:caseStudy}
\end{figure}

\subsection{Case Study: FPGA-Based RL for Smart EV Charging}

\ac{RL} has recently emerged as a promising paradigm for intelligent \ac{EV} charging management, particularly when real-time responsiveness and energy efficiency are required. Among the existing works, the \ac{FPGA}-based \ac{RL} architecture proposed in~\cite{damodarin2025smart} represents a key reference implementation, as it demonstrates how tabular \ac{QL} can be fully realized in hardware while meeting stringent embedded-system constraints.

To contextualize the contribution of this study, Figure \ref{fig:caseStudy} (a) presents the complete hardware
architecture of the \ac{EV} charging management system, organized as a modular and fully hardware-integrated control loop. Charger-level sensing inputs, including activity status, priority requests, and deployed current, provide the system perception and are processed by the \emph{state discretizer}. The resulting discrete state is supplied to the \emph{\ac{RL} engine}, which maintains and updates the learned knowledge. Action selection is carried out by the \emph{policy engine}, while the \emph{action generator}
translates the selected decision into charger-level control signals. In parallel, system performance is assessed by the \emph{reward generator}, which closes the learning loop.
Each of these components is integral to the overall operation and is discussed in detail in the following subsections, where their individual roles and hardware realizations are
analyzed. Building upon the system overview, Figure \ref{fig:caseStudy} (b) details the internal structure of the \ac{QL} engine. The Q-table is implemented as an array of dual-port \emph{Action RAMs}, enabling concurrent read and write operations. A dedicated MAX unit and a hardware Q-updater implement the learning rule using fixed-point arithmetic, ensuring deterministic timing and low power consumption. Action selection is performed by a dedicated policy engine, shown in  Figure \ref{fig:caseStudy} (c), which implements an $\epsilon$-greedy strategy entirely in hardware. A linear feedback shift register (LFSR) is employed to generate pseudo-random values, where the most significant bits (MSBs) are used to select candidate actions, while the least significant bits (LSBs) are compared against the exploration parameter to decide between exploration and exploitation, enabling efficient stochastic decision making with negligible
hardware overhead. To enable tabular learning, continuous sensor data are mapped into a discrete state space using a base-7 encoding scheme, as illustrated in Figure \ref{fig:caseStudy} (d). Each charger condition is converted into a digit, and weighted \ac{LUT}s generate a unique global state index. The action generator, depicted in Figure \ref{fig:caseStudy} (e), performs the inverse operation by decoding the selected action using base-3 look-up tables. This module translates abstract \ac{RL} decisions into concrete charger current levels while enforcing safety constraints. Finally, the reward generator shown in Figure \ref{fig:caseStudy} (f) evaluates system performance by assigning partial rewards to each charger state and aggregating them into a cumulative reward. Overcurrent conditions are explicitly penalized, ensuring safe and stable operation.

Beyond its architectural relevance, this study provides quantitative evidence of the practical viability of \ac{RL}-based \ac{FPGA} solutions for grid-edge control. The proposed system was implemented on an AMD ZCU104 platform (Zynq UltraScale+ XCZU7EV) and evaluated using a three-charger configuration, resulting in a discrete state space of 343 states. Fixed-point representations were employed, with Q-values encoded on 16 bits, states on 9 bits, and actions on 5 bits, enabling efficient hardware realization. Operating at a clock frequency of 84.7~MHz, the design achieves deterministic real-time behavior with a decision latency of approximately 34.3~ns. Experimental results report stable learning convergence, with an average reward of about 60, while hardware utilization remains limited, requiring less than 2.3\% of \ac{LUT}s and \ac{BRAM}s and below 0.2\% of \ac{DSP} resources. These results position this work as a representative reference for embedded, energy-efficient
\ac{RL} deployment in smart \ac{EV} charging infrastructures.

{The smart \ac{EV} charging case study primarily reflects \emph{\ac{RL} as the optimizer}, as the goal is to learn optimal charging policies under dynamic constraints such as energy demand, pricing, and grid conditions. The problem is characterized by a relatively large state space and multi-objective optimization. However, when deployed on \ac{FPGA} platforms, aspects of \emph{\ac{RL} as the workload} also arise, particularly when real-time decision-making and energy-efficient inference are required.}

\section{Challenges and Future directions}
\label{sec6}
Despite the demonstrated benefits of \ac{RL}-based \ac{FPGA} and \ac{IC} implementations, several challenges continue to limit their scalability, flexibility, and deployment maturity. These challenges stem from the mismatch between \ac{RL} algorithmic demands and hardware constraints. {To maintain consistency with the taxonomy introduced in Section~4, the challenges and future directions are organized according to the two principal roles of RL: as the workload and as the optimizer. When \ac{RL} is treated as the workload, challenges primarily arise from hardware constraints and the complexity of efficiently mapping learning algorithms onto \ac{FPGA}/\ac{IC}  platforms, whereas when \ac{RL} is used as an optimizer, challenges are mainly related to learning stability, system-level integration, and the complexity of applying \ac{RL} to dynamic hardware environments.} This section summarizes the key open challenges and outlines future research directions toward more scalable, robust, and reusable \ac{RL}-driven hardware systems.

\subsection{Limitations}

Despite the significant progress achieved by \ac{RL}-based \ac{FPGA} and \ac{IC} accelerators in terms of latency, energy efficiency, and real-time capability, several fundamental limitations remain that hinder their broader adoption and long-term scalability. These limitations span algorithmic, architectural, system-level, and methodological dimensions and are consistently observed across the surveyed works.

\subsubsection{RL as the Workload}

\color{black}

\begin{itemize}[leftmargin=10pt]

\item \textbf{Scalability and generalization:}
With respect to state and action spaces, scalability remains a primary limitation, particularly for tabular methods such as \ac{QL} and \ac{SARSA}. While these approaches map efficiently to on-chip memories, the quadratic growth of the Q-table with increasing state--action cardinality rapidly exhausts \ac{BRAM} or \ac{SRAM} resources. Even with aggressive quantization or compression, most implementations remain restricted to small or moderately sized environments, limiting applicability to complex real-world systems. Moreover, most implementations are validated on synthetic benchmarks or small-scale environments, with limited evidence of robustness under non-stationary workloads, hardware aging, or environmental uncertainty. This raises questions about the deployment readiness of current \ac{RL}-based \ac{FPGA}/\ac{IC}  systems in safety-critical or industrial contexts.

\item \textbf{On-device training and memory bandwidth constraints:}
On-device training remains challenging for deep \ac{RL} algorithms such as \ac{DQN}, \ac{PPO}, and \ac{AC} variants. Although inference pipelines are well supported on \ac{FPGA} fabrics, full training requires backpropagation, gradient accumulation, and frequent parameter updates, which introduce substantial computational and memory overhead. In addition, replay buffers, rollout memories, and gradient storage often exceed on-chip capacity, forcing reliance on external DDR memory. This dependency increases latency and power consumption and diminishes the theoretical energy-efficiency benefits of hardware acceleration.

\item \textbf{Exploration, numerical precision, and learning stability:}
Efficient hardware realization of exploration strategies and learning updates remains challenging due to tight constraints on randomness generation, numerical precision, and control complexity. Techniques such as $\epsilon$-greedy exploration, entropy regularization, and stochastic noise injection must balance exploration effectiveness with determinism and implementation simplicity. Hardware-friendly pseudo-random number generators may introduce bias, while more sophisticated exploration schemes increase resource usage and control overhead. In parallel, reduced-precision fixed-point arithmetic is commonly employed to improve efficiency and lower power consumption; however, aggressive quantization can distort value estimates, gradients, or policy distributions. These combined effects may degrade convergence, induce oscillatory behavior, or compromise policy stability, particularly in \ac{AC} and policy-gradient methods.

\item \textbf{Algorithm--hardware co-design and portability limitations:}
Algorithm--hardware co-design in \ac{RL}-based \ac{FPGA} and \ac{IC} systems remains highly specialized and lacks portability. Many accelerators are tightly optimized for a specific \ac{RL} algorithm, network topology, or application scenario, which limits reuse across different tasks. Adapting an existing architecture to alternative \ac{RL} formulations often requires substantial redesign, in contrast to the flexibility offered by software-based \ac{RL} frameworks.

\item \textbf{Limited \ac{IC}/SoC support for RL:} \ac{RL} implementations on \ac{IC}/\ac{SoC} platforms remain relatively limited compared to the extensive work on \acp{FPGA}. Unlike flexible and reconfigurable \ac{FPGA} solutions, \acp{IC} impose stricter constraints such as fixed architectures, tight power budgets, longer development cycles, and complex requirements including power gating, clock-tree synthesis, and memory hierarchy optimization. These limitations make algorithm–hardware co-design more challenging and less adaptable after fabrication. This gap highlights an important direction for future research, particularly in developing energy-efficient \ac{RL} accelerators, scalable on-chip learning mechanisms, and standardized design methodologies for integrating \ac{RL} into next-generation \ac{IC} platforms.
\color{black}

\end{itemize}

\subsubsection{RL as the Optimizer}

\color{black}

\begin{itemize}[leftmargin=10pt]

\item \textbf{Reconfiguration overhead and evaluation methodology:}
\ac{RL}-driven \ac{DPR} introduces non-negligible latency and energy overheads, which can offset performance gains if reconfiguration is frequent. Unstable or poorly trained policies may further trigger oscillatory reconfiguration behavior, raising concerns about system reliability and safety. In parallel, evaluation methodologies remain fragmented, with inconsistent reporting of performance, power, and area metrics. The absence of standardized benchmarks and evaluation protocols complicates fair comparison and obscures true design trade-offs.

\item \textbf{Toolchain complexity and design productivity:}
Developing \ac{RL}-based accelerators on \ac{FPGA} and \ac{IC} platforms remains a complex and labor-intensive process, requiring expertise across multiple abstraction levels, from \ac{RL} algorithm design and numerical optimization to \ac{HLS} directives, \ac{RTL} implementation, and hardware verification. While \ac{HLS} tools improve productivity, they offer limited native support for \ac{RL}-specific constructs such as experience replay, stochastic exploration, gradient synchronization, and dynamic policy updates. Consequently, designers often rely on manual optimizations and iterative design-space exploration to achieve acceptable performance, leading to long development cycles and reduced accessibility for non-hardware experts. This complexity hinders rapid prototyping, limits reproducibility, and slows the translation of emerging \ac{RL} algorithms into deployable hardware systems.

\item \textbf{Security and adversarial robustness:}
\ac{RL}-based \ac{FPGA} and \ac{IC} systems are potentially vulnerable to adversarial perturbations, reward manipulation, or malicious observation injection, particularly in online or adaptive learning settings. Despite increasing attention to secure and trustworthy \ac{AI}, security considerations remain largely unexplored in hardware-based \ac{RL} implementations, limiting their suitability for deployment in mission-critical or safety-sensitive environments.

\end{itemize}

\subsection{Future Directions}

Addressing the identified challenges in \ac{RL}-based \ac{FPGA} and \ac{IC} systems requires coordinated advances across algorithms, hardware architectures, system integration, and design methodologies. Future research should move beyond isolated accelerator prototypes toward holistic, reusable, and deployment-ready intelligent hardware platforms. Several promising research directions emerge from the surveyed literature, spanning scalability, learning efficiency, robustness, and application-driven deployment.

\subsubsection{RL as the Workload}
\color{black}

\begin{itemize}[leftmargin=10pt]
\item \textbf{Scalable and Generalizable \ac{RL} Architectures on FPGA:}
Future \ac{RL}-based \ac{FPGA} and \ac{IC} designs must prioritize scalability to large state and action spaces while ensuring robust generalization across tasks, environments, and operating conditions. A promising yet largely unexplored direction is the hardware realization of advanced \ac{RL} paradigms on \ac{FPGA}, including hierarchical, hybrid, and multi-stage architectures. In such designs, coarse-grained decision-making may be handled by lightweight or tabular \ac{RL} methods, while fine-grained and continuous control is delegated to deep \ac{RL} models, enabling efficient decomposition of complex control problems. Transfer learning \cite{lachenani2025advancing} in \ac{RL} further represents a key enabler for scalable hardware intelligence, allowing pretrained representations, value functions, or policy components to be reused across platforms and applications, thereby significantly reducing training cost and convergence time \cite{kheddar2024reinforcement}. Beyond classical formulations, emerging paradigms such as distributional \ac{RL}, risk-sensitive \ac{RL}, multi-objective \ac{RL}, and hierarchical \ac{MARL} offer richer decision criteria that better reflect real-world constraints and uncertainties. However, these paradigms remain largely uninvestigated in \ac{FPGA} and \ac{IC} implementations, presenting a fertile research opportunity for next-generation, scalable, and generalizable \ac{RL} hardware systems.

\item \textbf{Efficient On-Device and Lifelong Learning:}
Enabling efficient on-device and lifelong learning remains a pivotal research direction for \ac{RL}-based \ac{FPGA} systems. Future designs are expected to combine \emph{supervised}, \emph{unsupervised}, and \emph{self-supervised} learning paradigms in a complementary manner to reduce sample complexity, stabilize learning signals, and enable continual adaptation under strict latency, memory, and power constraints. Importantly, several recent algorithmic advances expose structural properties that are inherently amenable to \ac{FPGA} and custom \ac{IC} realization, irrespective of the specific application domain. \emph{Supervised} components, such as offline \ac{RL} and imitation-style learning from logged trajectories, enable most optimization to be conducted off-chip, yielding compact and stable policy representations for on-device deployment. This paradigm naturally fits FPGA-based systems, where deterministic inference, fixed-point arithmetic, and limited fine-tuning are preferred. \emph{Unsupervised RL}, in contrast, enables task-agnostic exploration and skill acquisition without reliance on external rewards, supporting long-term autonomy in nonstationary environments. Representative studies on unsupervised skill discovery and interaction-driven representation learning illustrate how intrinsic objectives, latent representations, and diversity measures can replace dense supervision and promote reusability of learned behaviors \cite{he2024robotic,chareyre2025unsupervised}. \emph{Self-supervised} learning further complements these paradigms by introducing structured predictive or reconstruction-based objectives that stabilize learning and reduce reliance on heuristic reward design. Recent works demonstrate that reconstruction-driven rewards and self-referencing mechanisms can mitigate reward nonstationarity and catastrophic forgetting, thereby improving sample efficiency and robustness during continual learning \cite{abbasi2025reinforcement,zhao2025self}. From a hardware perspective, such objectives are particularly attractive as they rely on regular, repeatable computations over temporal windows and latent features, which align well with pipelined and memory-aware \ac{FPGA} architectures. Finally, distributed operation motivates \emph{federated} and \emph{multi-agent} extensions in which multiple edge devices collaborate while preserving data locality and privacy. Recent decentralized \ac{RL} formulations for adaptive communication and collaboration provide concrete examples of how local decision-making can be achieved with minimal overhead and without centralized coordination \cite{wagle2025multi}. More broadly, these ideas suggest that future FPGA-based \ac{RL} systems may jointly optimize learning, communication, and resource utilization in a decentralized manner, enabling scalable and energy-efficient lifelong learning across heterogeneous edge platforms.

\item \textbf{Memory-, Dataflow-, and Precision-Aware Learning Architectures:}
As memory access increasingly dominates the energy and latency profile of deep \ac{RL} workloads, future \ac{FPGA} and \ac{IC} accelerators should adopt tightly integrated memory-centric and dataflow-aware design principles \cite{kaur2024comprehensive}. This includes hierarchical on-chip buffering, replay-buffer–aware memory organizations, near-memory or in-memory computation, and dataflow-specialized scheduling of \ac{RL} primitives such as policy evaluation, value updates, and experience sampling. By explicitly co-optimizing data movement and computation, such architectures can significantly alleviate bandwidth bottlenecks and improve energy efficiency. In parallel, precision-adaptive computation represents a key enabler for efficient and robust learning on hardware. Rather than relying on fixed numerical formats, future accelerators may employ mixed-precision or runtime-adjustable representations that adapt bit-widths according to learning phase, gradient dynamics, or policy confidence. When combined with memory-aware designs, adaptive precision can further reduce storage and communication overhead while preserving learning stability. Embedding numerical robustness indicators—such as overflow detection, quantization error monitoring, or confidence-aware scaling—directly into the learning loop offers a principled way to balance aggressive quantization with convergence reliability, making such architectures suitable for long-running and lifelong \ac{RL} workloads.

 \item \textbf{Edge and Deployment-Oriented RL Systems:}
Future \ac{RL}-based \ac{FPGA} platforms are expected to play a central role in edge and embedded applications that demand low latency, energy efficiency, and real-time adaptability. A primary research direction concerns \textit{\ac{RL}-driven \ac{UAV} navigation and swarm coordination} \cite{kheddar2025recent}, where recent advances in Transformer-enhanced \ac{RL} for \ac{UAV} decision-making highlight a growing need for hardware-efficient acceleration at the edge. Offline sequence-modeling approaches, such as the maneuver-conditioned decision Transformer (MCDT) \cite{jung2024maneuver} and the Transformer-weighted A* framework \cite{zhu2022uav}, rely heavily on attention-based inference, making them well suited to \ac{FPGA} acceleration through pipelined and parallel architectures. In parallel, online \ac{DRL} models integrating Transformers, including TOSRL \cite{jiang2024autonomous} and Transformer-enhanced MADDPG \cite{li2023trans}, further emphasize strict real-time constraints, partial observability handling, and energy efficiency, all of which align naturally with \ac{FPGA} capabilities. Moreover, multi-agent Transformer-based methods such as MA2DBT \cite{gao2023sequential} and MAT-V \cite{jiang2023short} expose the scalability challenges of attention-driven coordination, reinforcing the relevance of \ac{FPGA} parallelism. Finally, multimodal perception--decision pipelines such as ViTAL \cite{neves2024multimodal}, which demonstrate real-world \ac{UAV} deployment, implicitly motivate hardware-efficient co-design of perception and \ac{RL} inference. Beyond aerial robotics, these architectural insights extend naturally to other safety-critical domains, such as \textit{intelligent sensing and imaging for medical diagnostics} \cite{habchi2025advanced,habchi2025machine}. Recent studies show that \ac{RL}, particularly \ac{DQN}-based frameworks, can enhance medical imaging tasks through adaptive localization, segmentation, and classification, as evidenced by cervical cell localization \cite{khajuria2024active} and MRI carcinoma segmentation \cite{tao2022seqseg}. Such trends motivate FPGA-based acceleration to support low-latency and energy-efficient execution of \ac{RL}-enhanced perception pipelines. Furthermore, multimodal diagnostic systems indicate that FPGA-enabled co-design of sensing, deep perception, and \ac{RL} inference can facilitate robust, adaptive, and non-invasive diagnostic platforms for real-time clinical deployment. Similarly, \ac{RL} represents a promising yet underexplored direction for \textit{image-based high-energy physics detectors} \cite{kheddar2025image}, particularly when combined with \ac{FPGA} acceleration. Although existing studies remain limited, recent work such as \cite{kortus2023towards} demonstrates that label-free, graph-based \ac{RL} can achieve competitive accuracy and inference speed for charged-particle tracking, effectively replacing heuristic reconstruction in high-multiplicity detector environments. These characteristics strongly motivate future \ac{RL}-based \ac{FPGA} implementations, where massive parallelism, deterministic latency, and energy efficiency are essential for real-time detector pipelines. \ac{FPGA} acceleration could further enable online, trigger-level decision-making and scalable deployment of graph-based \ac{RL} models under strict throughput and timing constraints. Finally, \textit{automatic speech recognition} also emerges as a promising application domain \cite{djeffal2025robust,essaid2025deep,kheddar2024automatic}, where \ac{RL}-enabled \ac{FPGA} implementations could support adaptive inference and robustness in noisy, real-world environments. Collectively, these application-driven research avenues highlight the potential of \ac{RL}-based \ac{FPGA} systems to outperform static heuristics in dynamic, data-intensive, and latency-constrained settings.

\item \textbf{Generative AI and LLM-Driven \ac{RL} on  \ac{FPGA} Platforms:}
The convergence of \acp{LLM}, generative AI, and \ac{RL} represents a highly promising future direction for \ac{FPGA}-based intelligent systems, particularly for long-horizon, context-aware, and decision-intensive applications. Recent advances demonstrate that \textit{Transformer-based and generative models} can be efficiently mapped onto reconfigurable hardware. For instance, \cite{zhang2025fpga} showed that \ac{FPGA} acceleration of \ac{CNN}– Transformer architectures significantly reduces latency and energy consumption while preserving accuracy for real-time edge inference. Similarly, \textit{multimodal generative processing} has been explored in \cite{das2026hardware}, where a \ac{SoC} -\ac{FPGA} enables low-power, high-throughput medical image fusion across heterogeneous sensing modalities. Beyond perception-centric workloads, \textit{lightweight \acp{LLM}} are increasingly being used to guide or augment \ac{RL} agents through semantic reasoning and structured knowledge integration. Representative examples include interpretable and standards-compliant decision-making for tunnel maintenance \cite{yu2025lightweight}, as well as \ac{LLM}-guided \ac{MARL} with Transformer-based coordination for dynamic wafer batch scheduling \cite{xiang2025dynamic}. These studies highlight the potential of \ac{LLM} s to enhance exploration, policy interpretability, and coordination in complex environments. Nevertheless, a critical research gap remains: existing \ac{LLM} –\ac{RL} frameworks are predominantly evaluated on \ac{CPU}/\ac{GPU} platforms, while \ac{FPGA} implementations either exclude \ac{RL} entirely or focus solely on inference acceleration. Future research should therefore investigate tightly coupled \ac{RL}–\ac{LLM} co-design on \ac{FPGA}/\ac{IC} platforms, including hardware-aware training, on-chip memory optimization, quantized Transformer–\ac{RL} pipelines, and runtime reconfigurability. Addressing these challenges would enable fully autonomous, low-latency, and energy-efficient generative \ac{RL} systems suitable for edge and safety-critical applications.

\item \textbf{Quantum  RL on FPGA:}
Quantum and quantum-inspired \ac{RL} constitute a transformative long-term research direction, where \ac{FPGA}-based \ac{RL} is expected to play a pivotal enabling role. Recent breakthroughs have demonstrated that sub-microsecond-latency \ac{RL} agents can be deployed on \acp{FPGA} to control quantum systems in real time, achieving feedback decisions well within qubit coherence times \cite{reuer2023realizing}. Such results validate the suitability of \acp{FPGA} for hosting ultra-low-latency, deterministic \ac{RL} controllers capable of learning directly from physical measurements, without relying on explicit system models. Looking forward, hybrid quantum--classical \ac{RL} architectures represent a compelling perspective, where \acp{FPGA} serve as classical orchestration, control, and learning backends for near-term quantum hardware. In this paradigm, FPGA-based \ac{RL} agents can manage fast feedback loops, policy inference, and online adaptation, while higher-level quantum processing is delegated to quantum devices. This tight hardware--algorithm coupling opens the door to scalable quantum control, adaptive error mitigation, and autonomous calibration. 

\end{itemize}

\subsubsection{RL as the Optimizer}
\color{black}

\begin{itemize}[leftmargin=10pt]

\item \textbf{Modular, Portable, and Toolchain-Supported \ac{RL} Hardware Co-Design:}
To ensure portability, scalability, and reuse across evolving \ac{RL} algorithms, future hardware platforms should emphasize modular algorithm--hardware co-design. This entails decomposing \ac{RL} systems into standardized building blocks—such as policy inference, value estimation, exploration mechanisms, and learning updates—with well-defined interfaces and parameterizable implementations. Such modularity enables a single \ac{FPGA} or \ac{IC} architecture to support a wide spectrum of \ac{RL} methods, from value-based to policy-gradient and hybrid approaches, while minimizing redesign effort and improving longevity of hardware investments. Equally important is the development of \ac{RL}-aware design automation and toolchains that bridge the gap between algorithm design and hardware realization. Future \ac{HLS} and RTL flows should natively support \ac{RL}-specific constructs, including replay buffers, stochastic sampling, gradient pipelines, and multi-agent synchronization. Automated design-space exploration—potentially driven by \ac{RL} itself—can further optimize architectural parameters under power, latency, and area constraints. Together with domain-specific languages and reusable templates for \ac{RL} hardware description, such toolchains can dramatically reduce development time, lower the entry barrier for non-hardware experts, and accelerate the translation of emerging \ac{RL} algorithms into efficient, deployable hardware systems.

\item \textbf{Secure, Reliable, and Trustworthy \ac{RL} Hardware on FPGA:}
As \ac{RL}-based systems are increasingly deployed in safety-critical, industrial, and cybersecurity-sensitive environments, security, robustness, and trustworthiness emerge as fundamental research directions. Recent studies in cybersecurity demonstrate the strong potential of \ac{RL} and \ac{DRL} for complex threat attribution and adaptive defense. For instance, \ac{DRL}-based malware attribution has been shown to significantly outperform conventional \ac{ML} techniques in identifying advanced persistent threats \cite{basnet2025advanced}, albeit at the cost of substantial computational resources and data dependencies. These characteristics strongly motivate the integration of \ac{FPGA} acceleration to enable low-latency, energy-efficient, and deployable \ac{RL} security solutions. In parallel, advances in attention-based and Transformer-inspired intrusion detection systems for industrial \ac{IoT} (IIoT) and internet of medical thing (IoMT) environments report near-perfect detection accuracy and ultra-low inference latency on \ac{CPU}/\ac{GPU} platforms \cite{gueriani2025se,gueriani2024adaptive,alemayehu2025systematic,gueriani2025robust}, highlighting the feasibility of real-time AI-driven defence. However, these works largely rely on static inference models and lack adaptive, online learning capabilities. Incorporating \ac{RL} into such frameworks—and deploying them on \ac{FPGA} platforms—would enable continuous adaptation to evolving attack patterns while preserving strict timing and reliability constraints required by critical infrastructures. Furthermore, blockchain-enabled security architectures have demonstrated strong potential for ensuring data integrity, trust management, and decentralized decision-making in smart grids and \ac{IoT} systems, such as in \cite{mohanta2025smart}. Future research may explore the co-design of blockchain mechanisms with \ac{RL}-based \ac{FPGA} accelerators, where smart contracts ensure secure policy updates, auditable decisions, and resilient coordination among distributed agents. Such integration could address ethical, transparency, and accountability concerns increasingly associated with autonomous \ac{RL} systems. In addition, \ac{GNN}-based information security and biometric \ac{DL} reveal strong potential for \ac{RL}-based \ac{FPGA} systems. Integrating \ac{RL} with GNN steganalysis and biometric authentication enables adaptive, low-latency, and energy-efficient intelligence, overcoming static inference limitations \cite{hemis2025hierarchical}.  FPGA-based \ac{RL} can uniquely combine adaptive intelligence, deterministic latency, and hardware-level security primitives, enabling next-generation intrusion detection systems, malware attribution engines, and resilient cyber--physical defence platforms suitable for IIoT and critical national infrastructures.

\end{itemize}

These future directions indicate a shift from isolated, algorithm-specific accelerators toward scalable, secure, and application-driven \ac{RL}-based \ac{FPGA} and \ac{IC} platforms, {where \ac{RL} is leveraged both as a workload to be efficiently executed and as an optimizer to enable adaptive, self-configuring, and intelligent system behavior,} capable of long-term autonomous operation in real-world environments. {
Recent advances in related domains further highlight promising directions for scalable and adaptive \ac{RL} systems. In particular, the transition from single-agent to multi-agent learning paradigms enables coordinated decision-making in complex environments \cite{zhang2025reinforcement}. In parallel, adaptive dynamic programming-based control frameworks incorporating event-triggered mechanisms have demonstrated the potential to reduce communication and computational overhead in multi-agent systems \cite{zhang2026dynamic}. Moreover, emerging human--machine cooperative control strategies based on dynamic authority allocation illustrate how learning-based approaches can be extended to interactive and resource-constrained settings \cite{zhang2026dynamic2}. These developments suggest that future \ac{RL}-based hardware systems could benefit from integrating multi-agent coordination, communication-efficient learning, and adaptive control mechanisms.
}

\section{Conclusion}
\label{sec7}

This review has provided a comprehensive and hardware-centric synthesis of \ac{RL} implementations on \ac{FPGA} and \ac{IC} platforms, spanning more than two decades of research and over one hundred representative studies. By systematically organizing the literature across \ac{RL} algorithmic families, hardware abstraction levels, architectural design strategies, and application domains, the survey clarifies how learning-based decision-making has progressively transitioned from software-centric experimentation to practical, resource-aware hardware realizations.

A central outcome of this review is the identification of recurring architectural patterns that underpin successful \ac{RL} deployment on \ac{FPGA} and \ac{IC} substrates. Across classical tabular methods, deep value-based approaches, and modern \ac{AC} and policy-gradient algorithms, effective designs consistently exploit fine-grained parallelism, pipelined datapaths, quantized arithmetic, and tightly coupled on-chip memory hierarchies. At the same time, the review highlights a clear divergence between \ac{FPGA}-oriented solutions—favoring flexibility, reconfigurability, and rapid prototyping—and IC-based realizations, which prioritize determinism, compactness, and ultra-low power operation. The unified abstraction framework presented in this work demonstrates that, despite these differences, coherent design flows can bridge \ac{FPGA} prototyping and \ac{IC} deployment when hardware–algorithm co-design principles are applied early. From an algorithmic perspective, the survey reveals that no single \ac{RL} paradigm dominates across all hardware and application contexts. Lightweight tabular methods such as \ac{QL} and \ac{SARSA} remain attractive for real-time embedded control due to their simplicity and predictable timing, whereas deep \ac{RL} approaches, including \ac{DQN}, \ac{DDPG}, \ac{PPO}, and \ac{AC} variants, enable scalability to high-dimensional sensing and control problems at the cost of increased architectural complexity. \ac{MARL} further extends these capabilities to distributed and large-scale systems but introduces new challenges related to coordination, non-stationarity, and interconnect overhead. These trade-offs emphasize that algorithm selection must be driven jointly by task requirements and hardware constraints rather than by learning performance alone.

The application-level analysis confirms that \ac{RL}-enabled \ac{FPGA}/\ac{IC} systems have demonstrated tangible impact across diverse domains, including networking, information security, robotics, autonomous systems, intelligent sensing and imaging, and embedded optimization. Reported gains in latency, energy efficiency, throughput, and adaptability illustrate the practical benefits of embedding learning directly into hardware. However, the review also underscores persistent limitations that hinder wider adoption, such as sensitivity to numerical precision, exploration–exploitation trade-offs under deterministic execution, limited standardization of evaluation metrics, and challenges in scalability, portability, and reproducibility.

Looking forward, the findings of this survey suggest several promising research directions. These include the development of standardized hardware-aware \ac{RL} benchmarks, greater emphasis on end-to-end system-level evaluation, tighter integration of \ac{RL} with design automation and digital twin frameworks, and the exploration of hybrid learning paradigms that combine model-based reasoning with data-driven adaptation. Advances in emerging hardware technologies, including processing-in-memory, heterogeneous \acp{SoC}, and domain-specific accelerators, further open opportunities to rethink how \ac{RL} algorithms are mapped onto silicon.

\bibliographystyle{elsarticle-num}
\bibliography{references}

\end{document}